\documentclass[11pt]{article}
\usepackage[margin=1in]{geometry}
\usepackage{lmodern}
\usepackage[T1]{fontenc}
\usepackage{courier}
\usepackage{microtype}
\usepackage{xcolor}
\usepackage{amsmath,amssymb,amsthm,mathtools,amsfonts}
\usepackage{booktabs}
\usepackage{tabularx}
\usepackage{graphicx}
\usepackage{tikz}
\usetikzlibrary{intersections,arrows.meta,fit}
\usepackage{float}
\floatstyle{ruled}
\usepackage{algorithm}
\usepackage{algpseudocode}
\usepackage{subcaption}
\usepackage{enumitem}
\usepackage{aliascnt}
\usepackage[colorlinks=true,
            linkcolor=blue!55!black,
            citecolor=blue!55!black,
            urlcolor=blue!60!black]{hyperref}
\usepackage[nameinlink,capitalize,noabbrev]{cleveref}
\AddToHook{cmd/appendix/after}{\crefalias{section}{appendix}}
\usepackage{braket}

\definecolor{revisionblue}{RGB}{16,65,163}
\definecolor{qsafe}{RGB}{0,158,115}
\definecolor{qcert}{RGB}{0,114,178}
\definecolor{qpert}{RGB}{230,159,0}
\definecolor{qplay}{RGB}{213,94,0}

\newcommand{\RR}{\mathbb{R}}
\newcommand{\EE}{\mathbb{E}}

\newcommand{\cA}{\mathcal{A}}
\newcommand{\cS}{\mathcal{S}}
\newcommand{\cF}{\mathcal{F}}
\newcommand{\one}{\mathbf{1}}
\newcommand{\norm}[1]{\left\lVert #1\right\rVert}
\newcommand{\ip}[2]{\left\langle #1,#2\right\rangle}
\newcommand{\Reg}{\mathrm{Reg}}
\newcommand{\polylog}{\mathrm{polylog}}

\DeclareMathOperator*{\argmax}{arg\,max}

\newtheorem{theorem}{Theorem}[section]
\newaliascnt{proposition}{theorem}\newtheorem{proposition}[proposition]{Proposition}\aliascntresetthe{proposition}
\newaliascnt{lemma}{theorem}\newtheorem{lemma}[lemma]{Lemma}\aliascntresetthe{lemma}
\newaliascnt{corollary}{theorem}\aliascntresetthe{corollary}
\newaliascnt{assumption}{theorem}\newtheorem{assumption}[assumption]{Assumption}\aliascntresetthe{assumption}
\newaliascnt{definition}{theorem}\newtheorem{definition}[definition]{Definition}\aliascntresetthe{definition}
\theoremstyle{remark}
\newaliascnt{remark}{theorem}\newtheorem{remark}[remark]{Remark}\aliascntresetthe{remark}
\newtheorem*{remark*}{Remark}

\crefname{assumption}{Assumption}{Assumptions}
\Crefname{assumption}{Assumption}{Assumptions}

\title{Safe Quantum Stochastic Linear Bandits\thanks{A Preliminary version of this paper appeared at the 40th Annual Conference on Neural Information Processing Systems (NeurIPS 2026).}}
\author{ 
  Ruizhe Zhang\thanks{\texttt{rzzhang@purdue.edu}. Purdue University}
  \and
  Junyi Wu\thanks{\texttt{wu2393@purdue.edu}. Purdue University}
  \and
  Guang Lin\thanks{\texttt{guanglin@purdue.edu}. Purdue University}
}

\date{}

\begin{document}
\maketitle

\begin{abstract}
We initiate the study of safe quantum stochastic linear bandits, in which a learner maximizes an unknown linear reward while satisfying unknown linear mean constraints at every interaction. Under coherent unitary access to the joint reward-and-constraint distribution, we develop QRS-COLTS, a staged quantum resampled-scaling analog of constrained linear Thompson sampling (COLTS). With high probability, QRS-COLTS plays only feasible actions and achieves regret that is polylogarithmic in the query horizon and scales only with the square root of the number of constraints. We also study a model in which the learner queries superpositions of actions while an unsafe-action flag, a ``bomb'' that terminates the interaction upon violation, is measured after every query. In this setting, the bomb provides exact membership access to the safe set, and query regret is evaluated over the action superposition. For this model, we propose BCO-QLinTS, which selects feasible actions via bomb-certified quantum convex optimization without estimating the constraint matrix. Its query-weighted regret remains polylogarithmic in the horizon, with the constraints entering the bound only through the geometry of the safe set.
\end{abstract}

\section{Introduction}\label{sec:intro}

Sequential decision-making systems are increasingly deployed in settings where exploration must respect hard operational constraints. In adaptive clinical trials, a treatment policy must avoid unsafe toxicity levels; in recommendation systems, decisions may be constrained by budget or fairness requirements; and in robotics or power-grid control, unsafe actions can drive the system outside a stability envelope. The same tension arises in quantum-native control and simulation. For example, consider tuning a microwave pulse to read out a superconducting qubit, where a stronger pulse can make the two logical states easier to distinguish in a short measurement, yet even a pulse that reports the correct bit can drive the transmon into a higher energy level, compromising subsequent operations~\cite{HDC+25}. Starting from a conservative baseline pulse with a certified leakage margin, the learner seeks to maximize readout accuracy while keeping the leakage probability below a prescribed threshold at every step. When such controlled quantum dynamics are simulated coherently on a quantum computer, the reward and constraint signals are generated by a reversible quantum circuit, and we ask how much coherent access helps a learner that must remain safe.

We study this question through the lens of safe stochastic linear bandits~\cite{amani2019safeconstraints,khezeli2020safelinear,moradipari2021safelts,gangrade2024polytopes,gangrade2025colts}. At each interaction, the learner chooses a classical action $a_t\in\cA\subseteq\RR^d$ and receives a random feedback vector
$$
    X_t=(R_t,C_t)\in[0,1]^{1+m},
$$
where $R_t$ is the reward and $C_t=(C_{t,1},\ldots,C_{t,m})$ consists of noisy measurements of $m$ safety-related quantities, each normalized to lie in $[0,1]$. While the feedback distribution may depend on the chosen action, its expectation is assumed to be linear:
$$
    \EE[R_t\mid a_t=a]=\ip{\theta^\star}{a},
    \qquad
    \EE[C_t\mid a_t=a]=\Phi^\star a,
$$
for an unknown reward vector $\theta^\star\in\RR^d$ and an unknown constraint matrix $\Phi^\star\in\RR^{m\times d}$.

Safety is defined in terms of these expected constraint values. Specifically, an action $a\in\cA$ is \emph{feasible} if $\Phi^\star a\le \alpha$, where the inequality holds coordinate-wise, i.e., $\ip{\phi_i^\star}{a}\le \alpha_i$ for every $i\in[m]$, with $\phi_i^\star$ denoting the $i$-th row of $\Phi^\star$. Although individual noisy realizations $C_t$ need not satisfy the thresholds, the \emph{round-wise mean-safety} requirement~\cite{amani2019safeconstraints,moradipari2021safelts,gangrade2025colts} requires that, with high probability, every selected action is feasible in expectation: $\Phi^\star a_t\le \alpha$ for all $t\le T$. This mean-constraint formulation captures settings where safety is governed by expected performance, risk, or resource metrics, including efficacy--toxicity tradeoffs in dose-finding trials~\cite{thall2004dosefinding}, expected resource consumption~\cite{badanidiyuru2018bwk}, fairness of exposure~\cite{wang2021fairness}, and stability under uncertain dynamics~\cite{brunke2022saferobotics}. It also encompasses binary failure indicators whose expectation bounds the probability of an unsafe event.

Performance is measured by cumulative regret relative to the best feasible action:
$$
    a^\star\in\argmax_{a\in\cA:\,\Phi^\star a\le\alpha}\ip{\theta^\star}{a},\qquad \Reg(T)=\sum_{t=1}^T \ip{\theta^\star}{a^\star-a_t} .
$$
Following standard practice in safe bandits with round-wise constraints, we assume access to a known strictly safe baseline action $a_{\mathrm{safe}}\in\cA$ with true safety slack
$$
    \Gamma:=\min_{i\in[m]}\{\alpha_i-\ip{\phi_i^\star}{a_{\mathrm{safe}}}\}>0.
$$
Importantly, in this setting the learner knows $a_{\mathrm{safe}}$ and the thresholds $\alpha$, but need not know the value of $\Gamma$ in advance; instead, a lower confidence bound on $\Gamma$ can be constructed directly from the observed constraint feedback. This baseline provides the learner with a certified fallback action while it explores and expands its estimate of the unknown feasible region.

Classical algorithms for safe linear bandits, including confidence-set methods, conservative and stage-wise exploration, Thompson-sampling variants, and local-feedback or polytope models~\cite{amani2019safeconstraints,khezeli2020safelinear,moradipari2020stagewise,moradipari2021safelts,liu2021pessimistic,varma2023local,hutchinson2024directional,gangrade2024polytopes,gangrade2025colts}, are fundamentally bottlenecked by classical Monte Carlo estimation. Even recent computationally efficient methods such as Constrained Linear Thompson Sampling (COLTS)~\cite{gangrade2025colts}, which replaces expensive robust subproblems with coupled perturbations and constrained optimization, observe only a single noisy sample of the reward and constraints per round; for instance, its zero-violation variant S-COLTS incurs a leading regret term of $\widetilde O(\sqrt{d^3T})$ for fixed baseline gap and slack~\cite{gangrade2025colts}. Indeed, even when all constraints are inactive (e.g., identically zero constraint feedback), classical linear-bandit lower bounds~\cite{dani2008stochastic,abbasi2011linearbandits,bubeck2012regret} imply an unavoidable $\Omega((d-1)\sqrt T)$ minimax expected regret under our $[0,1]$ feedback and unit-norm normalization for $d\ge 2$ and $T\ge(d-1)^2$ (\Cref{thm:classical-lower}).

In unconstrained settings, recent quantum bandit algorithms have shown that coherent oracle access can break this $\sqrt{T}$ barrier. Under quantum reward oracles, multi-armed and stochastic linear bandits admit polylogarithmic regret in the query horizon $T$~\cite{wan2023quantumbandits}, with related quantum speedups established for convex bandits, Bayesian optimization and kernelized bandits, heavy-tailed bandits, and Lipschitz bandits~\cite{li2022quantumconvexbandits,dai2023quantumbo,hikima2024quantumkernelized,wu2023heavytailed,yi2026quantumlipschitz}. However, these unconstrained methods do not address the core challenge of safe exploration: certifying round-wise feasibility under unknown constraints. Doing so requires simultaneously handling reward uncertainty optimistically and constraint uncertainty conservatively across $(1+m)$-dimensional vector feedback. This motivates the central question of our work:
\begin{center}
    \emph{Can quantum access break the classical $\sqrt{T}$ regret barrier for safe linear bandits while keeping every selected action safe?}
\end{center}

\subsection{Main results}
\label{sec:intro-main-results}

We make two main contributions. First, at the conceptual level, we initiate the study of safe quantum linear bandits and formulate three quantum feedback models with round-wise safety constraints: (i) a classical-action, coherent-feedback model; (ii) a classical-action, classical-feedback model, which serves as a statistical baseline and admits no regret speedup; and (iii) a coherent-action bomb model evaluated under query-weighted regret. Second, at the algorithmic and statistical levels, we establish positive results in the first and third models showing that coherent quantum access enables polylogarithmic regret in the horizon $T$ while preserving each model's safety guarantee, complemented by lower bounds characterizing the dimension and baseline-slack dependencies under coherent feedback.

\begin{figure}
    \centering
    \begin{subfigure}[c]{0.49\textwidth}
        \centering
        \includegraphics[scale=0.5]{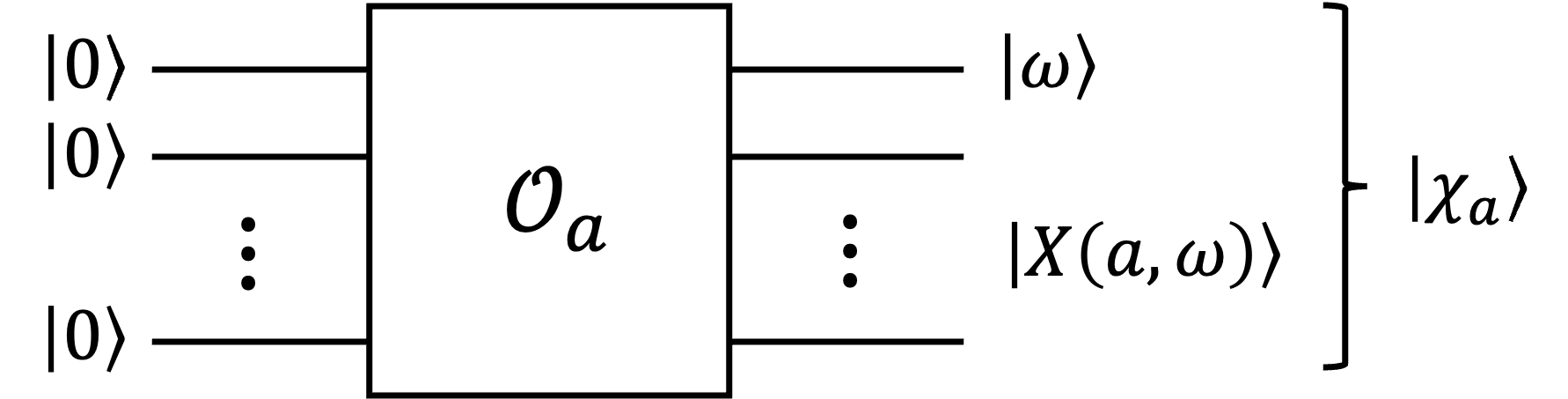}
        \caption{Classical-action, coherent-feedback model}
        \label{fig:model_1}
    \end{subfigure}
    \hfill
    \begin{subfigure}[c]{0.49\textwidth}
        \centering
        \includegraphics[scale=0.5]{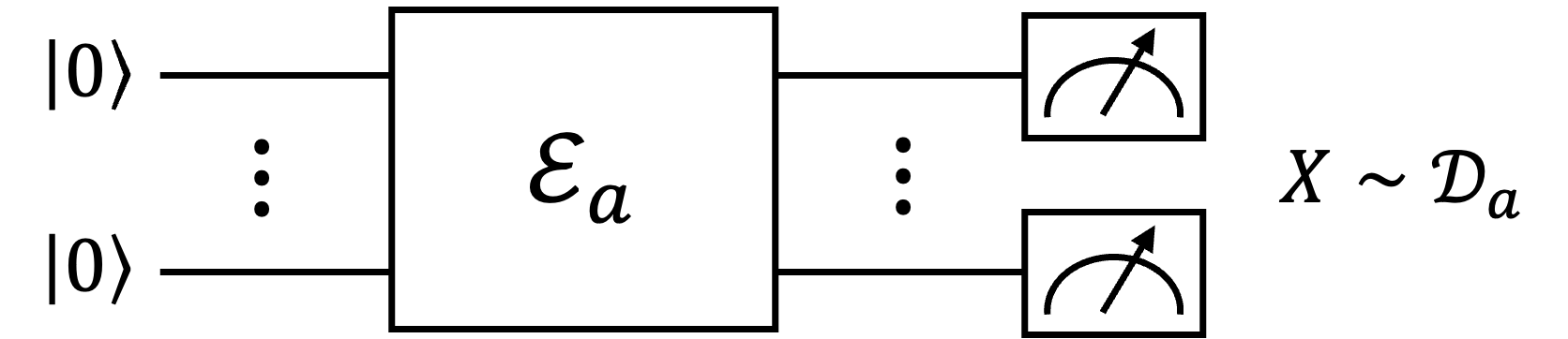}
        \caption{Classical-action, classical-feedback model}
        \label{fig:model_2}
    \end{subfigure}
    
    \begin{subfigure}[c]{\textwidth}
        \centering
        \vspace{1em}
        \includegraphics[scale=0.5]{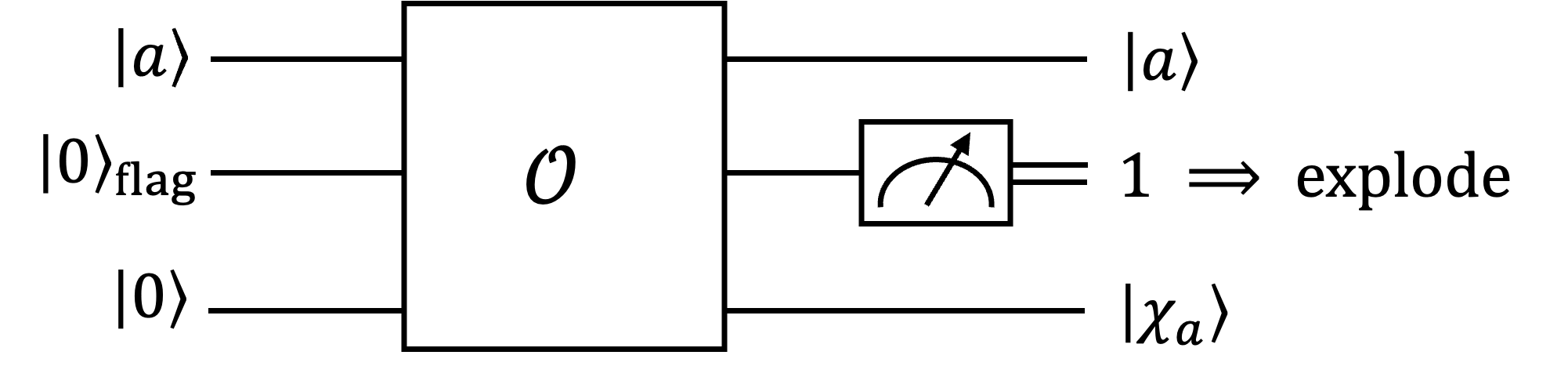}
        \caption{Coherent-action, coherent-feedback model with a safe-action bomb}
        \label{fig:model_3}
    \end{subfigure}
    \caption{Safe quantum bandit models. In~\Cref{fig:model_1}, each action $a$ has a corresponding unitary oracle $\mathcal{O}_a$ that prepares a coherent qsample of the stochastic feedback. In~\Cref{fig:model_2}, each query applies a quantum channel $\mathcal{E}_a$ followed by measurement, yielding classical feedback. In~\Cref{fig:model_3}, the learner queries actions in superposition (first register) while the oracle attaches a safety flag (second register) and prepares the reward qsample (third register); measuring $1$ on the flag register triggers a bomb explosion and terminates the algorithm.}
    \label{fig:models}
\end{figure}

\paragraph{Safe quantum bandit models.}
We consider three quantum feedback models in this paper (\Cref{fig:models}). Across all three models, the benchmark is the optimal feasible action satisfying the mean constraints $\Phi^\star a\le \alpha$, while the models differ in how the learner queries the feedback distribution and the safe set.

The first model extends the coherent reward oracle of Wan et al.~\cite{wan2023quantumbandits} to vector-valued reward and constraint feedback. Intuitively, after selecting a classical action $a$, the learner can call a unitary quantum oracle $\mathcal O_a$ that prepares a coherent quantum sample (``qsample'') of the joint feedback $X(a)$, as defined below.

\begin{definition}[Classical-action, coherent-feedback model]
\label{def:intro-classical-action-coherent}
To evaluate a classical action $a\in\cA$, the quantum learner receives coherent unitary access to the feedback sampler for that action via a quantum oracle $\mathcal O_a$ and its inverse $\mathcal O_a^\dagger$:
$$
    \mathcal O_a\ket{0}\ket{0}
    =
    \sum_{\omega}\sqrt{p_a(\omega)}\ket{\omega}\ket{X(a,\omega)},
$$
where the first register holds the coherent random seed $\omega$ and the second register encodes the feedback vector satisfying
\begin{align*}
    \sum_\omega p_a(\omega) X(a,\omega)=\EE[X_t\mid a_t=a]=(\ip{\theta^\star}{a},\Phi^\star a)\,.
\end{align*}
Each call to $\mathcal O_a$ or $\mathcal O_a^\dagger$ counts as one of the $T$ queries and incurs regret $\ip{\theta^\star}{a^\star-a}$. Safety requires every queried action to satisfy $\Phi^\star a\le\alpha$ with high probability.
\end{definition}

To instantiate \Cref{def:intro-classical-action-coherent} in the qubit readout setting, let $a\in\Delta^{d-1}$ be a probability vector over $d$ candidate readout circuits, and assume controlled access to $U_j$ and $U_j^\dagger$, where $U_j\ket{0}=\sum_y\sqrt{p_j(y)}\ket{y}$ for $j\in[d]$. Here, an outcome $y=(b,\hat b,\ell)$ records the prepared logical bit $b$, the reported bit $\hat b$, and the post-readout energy level $\ell$, with reward and leakage indicators $h_j(y)=(\one\{\hat b=b\},\,\one\{\ell\ge 2\})$. By preparing the superposition $\sum_{j=1}^d\sqrt{a_j}\ket{j}$, applying the $j$-th circuit conditioned on $\ket{j}$, and computing $h_j(y)$ into the feedback register, we obtain the oracle
$$
  \mathcal O_a\ket{0}\ket{0}=\sum_{j,y}\sqrt{a_jp_j(y)}\ket{j,y}\ket{h_j(y)},
$$
where each call to this joint circuit or its inverse counts as a single quantum query. Letting $v_j=\EE_{y\sim p_j}[h_j(y)]\in[0,1]^2$, the expected feedback under policy $a$ is $\sum_{j=1}^d a_j v_j$, which is linear in $a$, and any mixture of circuits with certified leakage margins provides a baseline action $a_{\mathrm{safe}}$. Finally, rescaling both the scores and the leakage threshold by $1/\sqrt{d}$ ensures that the unknown parameter vectors have at most unit $\ell_2$-norm, so regret in the original units is simply $\sqrt{d}$ times the normalized regret.

The second model provides only incoherent access to the feedback, as in the classical bandit setting, with quantum processing allowed only between queries.

\begin{definition}[Classical-action, classical-feedback model]
\label{def:intro-classical-action-classical}
At each round, the learner chooses a classical action $a\in\cA$ and observes a single classical feedback sample $X(a,\omega)$ drawn from $\omega\sim p_a$. Equivalently, the query applies a quantum channel $\mathcal{E}_a$ followed by an immediate standard-basis measurement:
\begin{align*}
    \rho_0\mapsto \mathcal{E}_a(\rho_0),\quad \rho_0={\ket{0}\!\bra{0}}, \qquad \text{followed by standard-basis measurement}\,.
\end{align*}
Because the measurement collapses the output state, the learner does not have reversible coherent access to the feedback randomness. This model subsumes the standard classical safe stochastic linear bandit setting.
\end{definition}

Because all information about $(\theta^\star,\Phi^\star)$ arrives through classical samples, any quantum learner in \Cref{def:intro-classical-action-classical} induces a classical randomized policy over actions and therefore inherits the $\Omega((d-1)\sqrt{T})$ expected-regret lower bound of \Cref{thm:classical-lower}. Related classical-order query lower bounds have also been established for best-arm identification under one-time stochastic channel oracles by Buchholz et al.~\cite{buchholz2025channel}.

The third model additionally allows coherent superposition queries over actions, analogous to the quantum stochastic convex bandit model of Li et al.~\cite{li2022quantumconvexbandits}. To enforce safety across action superpositions, we adopt the bomb-query mechanism of Lin and Lin~\cite{lin2014bomb}, inspired by the Elitzur--Vaidman bomb-testing problem~\cite{Elitzur:1993xh}. Both forward and inverse queries compute the exact unsafe-action bit into a fresh flag qubit, and measuring outcome $1$ on the flag terminates the algorithm. To survive, the learner must keep the query amplitude on unsafe actions sufficiently small throughout execution.

\begin{definition}[Coherent-action, coherent-feedback model with a safe-action bomb]
\label{def:intro-coherent-action-bomb}
The quantum learner may query a superposition over actions. For each action $a\in\cA$, let
$$
    \ket{\chi_a}_X
    :=
    \sum_{\omega\in\Omega_a}
    \sqrt{p_a(\omega)}\,
    \ket{\omega}\ket{R(a,\omega)}
$$
denote a coherent qsample of the reward distribution. On a forward query with blank output registers, the oracle coherently attaches an unsafe-action flag and the reward qsample:
$$
    \ket{a}\ket{0}_f\ket{0}_X
    \mapsto
    \ket{a}\ket{f(a)}_f\ket{\chi_a}_X,
    \qquad
    f(a)=\one\{\Phi^\star a\nleq \alpha\}.
$$
Immediately after each forward or inverse query, the flag register $\ket{\cdot}_f$ is measured; if the outcome is $1$, a bomb explosion occurs and the algorithm terminates. Following Lin and Lin~\cite{lin2014bomb}, we require the cumulative explosion probability over all $T$ queries to be at most $\varepsilon$. Because a query may place amplitude across multiple actions, its regret is evaluated with respect to the action distribution of the query state (\Cref{def:query-weighted-regret-main}).
\end{definition}

\paragraph{Classical actions with coherent feedback: QRS-COLTS.}
We establish positive results for the first and third models. In the classical-action, coherent-feedback model, we propose QRS-COLTS (Quantum Resampled Scaling COLTS), a staged quantum analog of S-COLTS~\cite{gangrade2025colts}. QRS-COLTS preserves the perturb-and-scale structure of COLTS, combining coupled Gaussian Thompson perturbations, linear constrained subproblems, and conservative scaling toward $a_{\mathrm{safe}}$, while replacing classical sample averaging within each stage by bounded multivariate quantum mean estimation and estimating a lower confidence bound on $\Gamma$ online.

\begin{theorem}[QRS-COLTS; Informal version of \Cref{thm:main}]
\label{thm:informal-main}
Consider the classical-action, coherent-feedback model over a compact convex action set $\cA\subseteq\RR^d$ under unit $\ell_2$-norm bounds on actions, $\theta^\star$, and the rows of $\Phi^\star$, given a known strictly safe baseline $a_{\mathrm{safe}}\in\cA$ (without knowing its slack $\Gamma>0$). With probability at least $1-3\delta$, QRS-COLTS satisfies $\Phi^\star a_t\le\alpha$ at every query $t\le T$ and achieves
$$
    \Reg(T)\le\widetilde O\!\left(\sqrt{m+1}\,d^2\log^{3/2}(2T/\delta)
        \left(1+\frac{\Delta(a_{\mathrm{safe}})}{\Gamma}\right)\right),
$$
where $\Delta(a_{\mathrm{safe}})=\ip{\theta^\star}{a^\star-a_{\mathrm{safe}}}$ is the baseline's reward gap to the best safe action and $\Gamma$ is its true safety slack.
\end{theorem}

The classical computational cost of QRS-COLTS is governed by its stage-wise subproblems. Across $S=\widetilde O(d\log T)$ stages, stage $s$ solves at most $K_s=O(\log(s/\delta))$ linear optimization subproblems over $\cA$ intersected with $m$ sampled halfspaces (which reduce to linear programs whenever $\cA$ is a polytope). Each stage also performs a rank-one update to the upper-triangular Cholesky factor $U_s$ of the $d\times d$ design matrix $V_s=U_s^\top U_s$ to generate the coupled perturbations, and recomputes the $m+1$ ridge regression estimates in $O((m+1)d^2)$ operations. If each subproblem takes $\mathcal C_{\mathrm{opt}}$ classical operations, the total classical optimization and regression cost is
$$
  \widetilde O\!\left(d\log T\,\left(\mathcal C_{\mathrm{opt}}+(m+1)d^2\right)\right).
$$
For general convex sets $\cA$ where exact maximization is unavailable, solving each sampled-feasible subproblem to certified additive objective error $T^{-2}$ (with exact infeasibility detection and conservative scaling) adds at most $1/T$ to the regret bound (\Cref{rem:subproblem_accuracy}), in which case $\mathcal C_{\mathrm{opt}}$ denotes the solver cost at accuracy $T^{-2}$. Consequently, quantum query savings translate into wall-clock speedups primarily when calls to the coherent quantum simulator dominate these classical optimization and regression updates. Achieving this statistical speedup requires more than plugging a quantum mean estimator into COLTS, as the stage-wise quantum estimates must be synchronized with resampled unsaturation and conservative scaling (\Cref{sec:technical-overview}).

Note that the $d^2$ factor in \Cref{thm:informal-main} includes a $\sqrt{d}$ contribution from the Euclidean norm of the $d$-dimensional Gaussian perturbations used to keep each stage subproblem linear. If computational tractability is set aside and each sampling stage instead globally maximizes an optimistic upper-confidence objective subject to optimistic constraints (a generally nonconvex problem), the dimension dependence improves to $d^{3/2}$ under the same oracle (\Cref{thm:global-optimism-upper}).

\paragraph{Coherent actions with a safe-action bomb: BCO-QLinTS.}
In the coherent-action bomb model, the bomb flag directly certifies whether an action is safe, albeit at the risk of terminating the algorithm. This allows the learner to treat the flag as an exact membership oracle and optimize directly over the true safe set $\cS$ without ever estimating the constraint matrix $\Phi^\star$. We realize this approach in BCO-QLinTS (Bomb-Certified Optimistic Quantum Linear Thompson Sampling), which in each stage maximizes $K_s$ perturbed linear reward objectives over $\cS$ via bomb-certified quantum convex optimization and estimates the reward at the most optimistic candidate action.

\begin{theorem}[BCO-QLinTS; Informal version of \Cref{thm:bco-qlints}]
\label{thm:informal-bomb-main}
Consider the coherent-action bomb model under unit $\ell_2$-norm bounds on actions and $\theta^\star$, and suppose the safe set $\cS$ satisfies $\mathbb{B}(a_{\mathrm{safe}},r)\subseteq\cS\subseteq\mathbb{B}(a_{\mathrm{safe}},R)$ around a known safe action $a_{\mathrm{safe}}$ with known radii $0<r\le R$. Then BCO-QLinTS has cumulative explosion probability at most $\varepsilon$ and, with probability at least $1-\delta$, its query-weighted regret up to termination satisfies
$$
    R_{\mathrm w}(T)
    \le \widetilde O\!\left(d^2\log^{3/2}(2T/\delta)
          +\frac{d^4\log^2(2T)}{\varepsilon}\right),
$$
where $\widetilde O(\cdot)$ suppresses additional polylogarithmic factors in $d$, $T$, $1/\delta$, and $R/r$.
\end{theorem}

In \Cref{thm:informal-bomb-main}, the first term bounds the regret incurred during reward estimation, while the second term captures the bomb-query overhead of optimizing over $\cS$. Specifically, BCO-QLinTS runs for at most $S=S_{\max}=\widetilde O(d\log T)$ stages and divides its per-stage explosion budget $\varepsilon/S$ equally across $K_s=O(\log(s/\delta))$ convex optimization calls, each requiring $Q=\widetilde O(d)$ coherent membership queries. Simulating these calls via the Lin--Lin bomb-query framework~\cite{lin2014bomb} consumes $\widetilde O(Q^2K_s^2S/\varepsilon)$ bomb queries in stage $s$, which sums to $\widetilde O(d^2S^2/\varepsilon)=\widetilde O(d^4\log^2T/\varepsilon)$ over all stages. Crucially, the number of constraints $m$ does not appear in the regret bound: because the bomb flag aggregates feasibility across all $m$ constraints into a single bit, the constraints enter only polylogarithmically through the geometric ratio $R/r$.

\paragraph{Lower bounds for classical actions.}
Finally, we establish two lower bounds in the classical-action, coherent-feedback model (\Cref{def:intro-classical-action-coherent}) showing that both the $d^{3/2}$ dimension dependence of \Cref{thm:global-optimism-upper} and an inverse-slack warm-up cost are unavoidable. First, for every $d\ge 2$, every $T\ge 1$, and every classical-action policy, there is an instance over a compact convex action set with inactive constraints on which the policy incurs $\Omega(d^{3/2}\log(1+T/d^{3/2}))$ expected regret (\Cref{thm:convex-quantum-lower}), obtained by simulating each convex-set query using one query to the spherical-cap construction of Liu, Li, and Lui~\cite{liu2026regretbounds}. Second, already for $d=2$ and $m=1$, enforcing round-wise mean safety with probability at least $3/4$ forces regret of order $\Omega(\Delta(a_{\mathrm{safe}})\min\{T,1/\Gamma\})$ with probability at least $1/2$ on some instance, even when the exact slack $\Gamma$ is revealed to the learner (\Cref{thm:slack-lower}).

\section{Related work}\label{sec:related}

\paragraph{Classical safe and constrained linear bandits.}
Classical approaches to safe linear bandits with unknown linear constraints broadly fall into two families: pessimistic-optimistic confidence-set methods~\cite{amani2019safeconstraints,khezeli2020safelinear,moradipari2020stagewise,liu2021pessimistic,varma2023local,hutchinson2024directional,gangrade2024polytopes} and randomized Thompson-sampling methods~\cite{moradipari2021safelts,gangrade2025colts}, with related approximate Thompson sampling techniques via Langevin Monte Carlo studied in~\cite{zheng2024accelerating,wang2025rethinking}. The closest predecessor, S-COLTS~\cite{gangrade2025colts}, uses coupled perturbations and constrained linear optimization, then scales toward $a_{\mathrm{safe}}$ to ensure round-wise safety. Our first algorithm, QRS-COLTS, keeps this perturb-and-scale architecture while replacing classical sample averaging within each stage by multivariate quantum mean estimation (\Cref{tab:linear-comparison}).

\paragraph{Quantum bandits and query models.}
Wan et al.~\cite{wan2023quantumbandits} initiated the study of regret minimization with quantum reward oracles, establishing polylogarithmic regret for multi-armed and stochastic linear bandits (QLinUCB; see \Cref{tab:linear-comparison}). Subsequent works by Dai et al.~\cite{dai2023quantumbo} and Hikima et al.~\cite{hikima2024quantumkernelized} similarly select actions classically while querying coherent reward unitaries and their inverses; in particular, specializing the kernelized analysis of Hikima et al.\ to the linear kernel yields $O(d^{3/2}\operatorname{polylog} T)$ regret for unconstrained quantum linear bandits~\cite[Appendix~C]{liu2026quantumlower}. Quantum speedups have also been established for best-arm identification~\cite{wang2021quantumexploration}, online exp-concave optimization~\cite{he2024expconcave}, stochastic convex bandits~\cite{li2022quantumconvexbandits}, heavy-tailed bandits~\cite{wu2023heavytailed}, Lipschitz bandits~\cite{yi2026quantumlipschitz}, and bandits with knapsacks~\cite{su2025quantumbwk}. Buchholz et al.~\cite{buchholz2025channel} give a classical-order lower bound for best-arm identification under their one-time stochastic channel oracle.

None of the above works address round-wise safety under unknown linear constraints. The closest constrained quantum formulation is the quantum bandits-with-knapsacks model of Su et al.~\cite{su2025quantumbwk}, which enforces cumulative resource budgets over the horizon rather than per-round feasibility constraints. Beyond the classical-action coherent-feedback model, we also introduce a coherent-action model equipped with a safe-action bomb oracle, inspired by the bomb-query framework of Lin and Lin~\cite{lin2014bomb}. Whereas QRS-COLTS queries individual classical actions and observes noisy numerical constraint feedback, the bomb oracle tests exact feasibility across action superpositions. Related coherent-action queries appear in the quantum convex bandit framework of Li and Zhang~\cite{li2022quantumconvexbandits}, where regret is charged to a committed classical action at every query round, with the learner playing the previous phase's action while making coherent queries in the current phase. In contrast, our bomb model evaluates performance directly on the queried superpositions via query-weighted regret, integrating the suboptimality gap over safe action mass and assigning a unit penalty to unsafe mass, while separately bounding the cumulative explosion probability. When every query is supported on a single safe action, query-weighted regret reduces to standard cumulative regret.

\begin{table}[tb]
\centering
\caption{Comparison of linear bandit algorithms that select one classical action per query. Here $\Delta=\Delta(a_{\mathrm{safe}})$ and $\Gamma=\min_i(\alpha_i-\langle\phi_i^\star,a_{\mathrm{safe}}\rangle)> 0$.}
\label{tab:linear-comparison}
\setlength{\tabcolsep}{3pt}
\renewcommand{\arraystretch}{1.4}
\renewcommand{\tabularxcolumn}[1]{m{#1}}
\begin{tabularx}{\linewidth}{@{}>{\raggedright\arraybackslash}m{1.9cm} c
  >{\centering\arraybackslash\hsize=0.82\hsize}X
  >{\centering\arraybackslash\hsize=1.18\hsize}X
  >{\centering\arraybackslash}m{3.5cm}@{}}
\toprule
 Algorithm & Unknowns & Feedback\linebreak per query & Action selection\linebreak and safety & Regret\\
\midrule
 QLinUCB\newline\cite{wan2023quantumbandits} & $\theta^\star$ & coherent reward sampler & optimistic maximization; no constraints & $\widetilde O(d^2\log^{3/2} T)$\\
 S-COLTS\newline\cite{gangrade2025colts} & $\theta^\star,\Phi^\star$ & one classical sample of $(R,C)$ & perturbed subproblem, then scaling toward $a_{\mathrm{safe}}$ & $\widetilde O((1+\Delta/\Gamma)\sqrt{d^3T})$ ${}+\widetilde O(d^2\Delta/\Gamma^2)$\\
 QRS-COLTS\newline(\Cref{thm:informal-main}) & $\theta^\star,\Phi^\star$ & coherent sampler of $(R,C)$ & $K_s$ perturbed subproblems, then scaling toward $a_{\mathrm{safe}}$ & $\widetilde O(\sqrt{m+1}\,d^2\log^{3/2}T)$ $\times(1+\Delta/\Gamma)$\\
\bottomrule
\end{tabularx}
\end{table}

\paragraph{Quantum mean estimation.}
Quantum Monte Carlo and amplitude estimation achieve the quadratic speedup $\widetilde O(1/\varepsilon)$ for scalar mean estimation~\cite{montanaro2015montecarlo,kothari2023mean}. For bounded vector feedback, we use the multivariate estimator of Cornelissen, Hamoudi, and Jerbi~\cite[Theorem~3.3]{cornelissen2022multivariate}. For $Z\in[0,1]^M$, it achieves $\ell_\infty$ error $\varepsilon$ with $\widetilde O(\sqrt M/\varepsilon)$ queries after rescaling (\Cref{thm:isotropifier}). Marsden et al.~\cite{marsden2025isotropic} study directional error guarantees in stochastic convex optimization.

\paragraph{Quantum convex optimization.}
Our coherent-action bomb algorithm BCO-QLinTS (\Cref{thm:informal-bomb-main}) relies on quantum convex optimization to maximize perturbed linear objectives over the unknown safe set via membership queries. Quantum convex optimization with membership and separation oracles was studied by van Apeldoorn et al.~\cite{vanapeldoorn2020convexopt} and Chakrabarti et al.~\cite{chakrabarti2020quantum}, who showed that optimizing over a well-rounded convex body in $\mathbb{R}^d$ requires only $\widetilde O(d)$ quantum membership queries, compared to the best known quadratic classical upper bound. We combine this quantum membership-query primitive with the bomb-query simulation of Lin and Lin~\cite{lin2014bomb} to obtain a bomb-certified safe optimization subroutine.

\section{Preliminaries}\label{sec:prelim}
\paragraph{Notation.}
For a positive integer $n$, $[n]=\{1,\dots,n\}$. Vectors are columns, $\ip{x}{y}=x^\top y$, and inequalities between vectors hold coordinate-wise. We write $\|x\|_2$ for the Euclidean norm, $\|x\|_A=\sqrt{x^\top Ax}$ for a positive definite matrix $A$, $\mathbb B(x,r)$ for the closed Euclidean ball of radius $r$ around $x$, $\one\{\cdot\}$ for the indicator function, and $\one_m$ for the all-ones vector in $\RR^m$. As in \Cref{sec:intro}, $a^\star$ is an optimal safe action, $\mu^\star=\ip{\theta^\star}{a^\star}$ is its mean reward, and $\Delta(a)=\mu^\star-\ip{\theta^\star}{a}$ is the gap of an action $a$. We use $\widetilde O(\cdot)$ to hide constant and polylogarithmic factors. A power of $\log T$ written inside $\widetilde O$ is the dependence that comes from the number of stages, and each theorem states which other logarithms it hides.

\subsection{Quantum computing basics}\label{sec:quantum-prelim}
We use only standard notions from the quantum circuit and query models; see, e.g., \cite{nielsen2010quantum} for a comprehensive reference.  A quantum register with basis indexed by a finite set $\Omega$ is a Hilbert space with computational basis $\{\ket{\omega}:\omega\in\Omega\}$.  A pure state is a unit vector $\ket{\psi}$, while a mixed state is represented by a density matrix $\rho$, i.e., a positive semidefinite operator with trace one.  Quantum gates are unitary maps acting on one or more registers.  More general noisy operations are quantum channels, i.e., completely positive trace-preserving maps $\mathcal E$ on density matrices.  A measurement is described by positive operators whose sum is the identity; measuring in the computational basis of $\sum_i c_i\ket{i}$ returns outcome $i$ with probability $|c_i|^2$ and collapses the state accordingly.

A quantum query algorithm interleaves fixed quantum gates with calls to an oracle.  In a coherent unitary oracle model, a query is a unitary transformation and may be used inside a larger coherent computation, typically along with its inverse.  This is the setting in which amplitude-estimation and quantum-mean-estimation algorithms obtain their speedups.  When only a measured feedback value is returned, the learner has no coherent access to the sampling randomness.

We use two kinds of action access.  In a \emph{classical-action} query, the learner first chooses a classical action $a$, and the oracle acts only on the feedback registers associated with that action.  In a \emph{coherent-action} query, the action itself is stored in a quantum register, for example $\sum_a \psi_a\ket{a}$, and the oracle acts coherently on all actions in superposition.  If such a query is later assigned regret, the natural action weight is the marginal probability $|\psi_a|^2$ (or the corresponding diagonal marginal for a mixed state).  This distinction is why the coherent-action model in \Cref{sec:coherent-action-bomb-model} uses query-weighted regret rather than ordinary played-action regret.

\subsection{Quantum mean-estimation primitives}
\label{subsec:mean-estimation-prelim}

The quantum algorithms in this paper use quantum subroutines to estimate expectations of random variables.  We state these subroutines as black boxes in the form needed for our bandit analysis.

A \emph{quantum sampling oracle} for a random variable $Y$ with distribution $\mathcal D$ is a unitary that prepares a coherent sample of $Y$ (so-called \emph{qsample}).  For example, if $Y=Y(\omega)$ and $\omega\sim p$, then the oracle prepares
$$
    \ket{0}\ket{0}
    \mapsto
    \sum_{\omega}\sqrt{p(\omega)}\,\ket{\omega}\ket{Y(\omega)}.
$$
Measuring the value register produces one ordinary classical sample from $\mathcal D$.  The quantum advantage comes from using the state coherently before measurement.

We first recall the scalar mean-estimation primitive (see, e.g., \cite{montanaro2015montecarlo,kothari2023mean}).

\begin{lemma}[Scalar quantum mean estimation]
\label{lem:qmc_scalar}
Let $Y\in[0,1]$ be a random variable with mean $\mu=\EE[Y]$. Given a quantum sampling oracle for $Y$ and its inverse, there is a quantum algorithm that outputs an estimate $\widehat\mu$ satisfying $\Pr\bigl[|\widehat\mu-\mu|>\varepsilon\bigr]\le\delta$ using at most $C_{\mathrm{qmc}}\varepsilon^{-1}\log(2/\delta)$ oracle queries for a universal constant $C_{\mathrm{qmc}}>0$.
\end{lemma}

Thus a scalar expectation can be estimated to accuracy $\varepsilon$ with
$\widetilde O(1/\varepsilon)$ coherent oracle queries, instead of the classical
$\Theta(1/\varepsilon^2)$ samples required by Monte Carlo.

In our safe-bandit problem, however, the feedback is not scalar: one query returns a vector
$$
    X(a)=(R(a),C_1(a),\ldots,C_m(a))\in[0,1]^{m+1}.
$$
A direct approach is to apply \Cref{lem:qmc_scalar} separately to all $m+1$ coordinates.  This gives coordinate-wise accuracy $\varepsilon$ at query cost $\widetilde O\!\left(\frac{m+1}{\varepsilon}\right)$. The query cost thus grows linearly with the number of constraints.

For vector feedback we use \textsc{QBounded}~\cite[Theorem~3.3]{cornelissen2022multivariate}. Rescaling the feedback vector to the unit Euclidean ball gives the following simultaneous coordinate-wise bound, which we prove in \Cref{sec:instantiations}.

\begin{theorem}[Bounded multivariate quantum mean estimation]
\label{thm:isotropifier}
Let $Z\in[0,1]^M$ be a vector-valued random variable with mean
$\mu=\EE[Z]$.  Given a quantum sampling oracle for $Z$ and its inverse, for any target accuracy
$0<\varepsilon\le1$ and failure probability $\delta\in(0,1)$, there is a quantum algorithm that outputs $\widehat\mu\in\RR^M$ such that $\Pr\bigl[\norm{\widehat\mu-\mu}_{\infty}>\varepsilon\bigr]\le \delta$ using $\widetilde{O}(\sqrt{M}/\varepsilon)$ oracle queries in the worst case, where $\widetilde O$ hides factors polylogarithmic in $M$, $1/\varepsilon$ and $1/\delta$.
\end{theorem}
For our feedback vector $X(a)$, the dimension is $M=m+1$.  For accuracy $1/n$ in every coordinate, scalar estimation therefore needs $\widetilde O((m+1)n)$ queries, while the multivariate estimator needs $\widetilde O(\sqrt{m+1}\,n)$ queries.  This gap gives the factor $\sqrt{m+1}$ in \Cref{thm:informal-main}.

\section{Technical overview}
\label{sec:technical-overview}

QRS-COLTS and BCO-QLinTS exploit coherent quantum access in fundamentally different ways, leading to distinct proof structures. QRS-COLTS operates closer to the classical safe-bandit paradigm: it commits to a single classical action before each query and must learn both the reward vector $\theta^\star$ and the constraint matrix $\Phi^\star$ from noisy feedback. In contrast, BCO-QLinTS operates in the coherent-action model, using the safe-action bomb as an exact membership oracle for the true feasible set $\cS$ so that only the reward vector $\theta^\star$ needs to be learned.

\subsection{QRS-COLTS}
\label{sec:qrs-overview}

In this subsection, we first reduce the regret guarantee of QRS-COLTS (\Cref{alg:qrs-colts}) to three per-stage properties, and then develop the algorithm in six steps that motivate each component and parameter choice. Throughout, let $a_{\mathrm{safe}}$ denote the known safe baseline and let $\Gamma=\min_{i\in[m]}\{\alpha_i-\ip{\phi_i^\star}{a_{\mathrm{safe}}}\}>0$ denote its true safety slack, which is unknown to the learner. \Cref{tab:qrs-parameters} in \Cref{app:qrs-colts-alg} summarizes all algorithm parameters.

\paragraph{The design in one calculation.}
Because quantum mean estimation produces a single estimate only after repeatedly querying the oracle at a fixed action, QRS-COLTS proceeds in stages $s=1,2,\dots$. Stage $s$ commits to one action $a_s\in\cA$ for $Q_s$ queries and estimates the mean feedback vector $\mu(a_s)=(\ip{\theta^\star}{a_s},\Phi^\star a_s)\in \mathbb{R}^{m+1}$ to coordinate-wise accuracy $1/n_s$. Since every call to $\mathcal O_{a_s}$ or $\mathcal O_{a_s}^\dagger$ incurs regret $\Delta(a_s)$, cumulative regret decomposes across stages as
\begin{equation}\label{eq:qrs-regret-accounting}
 \Reg(T)=\sum_{s}Q_s\,\Delta(a_s).
\end{equation}
QRS-COLTS is designed so that, with probability at least $1-3\delta$, its stages satisfy the following three key properties. Below, $\|a\|_{V_s^{-1}}$ measures the learner's directional uncertainty about $\mu(a)$ at the start of stage $s$, $\beta_s=1+\sqrt{s-1}$ is the stage confidence radius, $B=\widetilde O(\sqrt d)$ bounds the Euclidean norms of the Gaussian perturbations, and $c_{\mathrm{len}}>0$ is a fixed stage-length constant (all defined formally in Steps~1--6):
\begin{enumerate}[label=(P\arabic*)]
\item\label{item:P1} \emph{Round-wise safety.} Every stage plays a feasible action: $\Phi^\star a_s\le\alpha$.
\item\label{item:P2} \emph{Gap bounded by uncertainty.} Every non-warm-up stage satisfies $\Delta(a_s)\le C_\Delta\,B\,\beta_s\,\|a_s\|_{V_s^{-1}}$, where $C_\Delta=O(1+\Delta(a_{\mathrm{safe}})/\Gamma)$.
\item\label{item:P3} \emph{Stage length matched to uncertainty.} Every stage with $a_s\ne 0$ satisfies $Q_s\le\widetilde O(\sqrt{m+1}\,n_s)$ and $c_{\mathrm{len}}\le n_s\|a_s\|_{V_s^{-1}}\le c_{\mathrm{len}}+1$.
\end{enumerate}
Combining~\ref{item:P2} with the upper bounds in~\ref{item:P3} cancels the directional uncertainty $\|a_s\|_{V_s^{-1}}$ from the per-stage regret:
\[
 Q_s\,\Delta(a_s)
 \;\le\;\widetilde O(\sqrt{m+1}\,n_s)\cdot C_\Delta B\beta_s\|a_s\|_{V_s^{-1}}
 \;\le\;\widetilde O\bigl(\sqrt{m+1}\,C_\Delta\,B\,\beta_s\bigr).
\]
Intuitively, a stage makes many queries only when its chosen action already has small uncertainty and is therefore nearly optimal. Meanwhile, the lower bound in~\ref{item:P3} ensures that every completed stage multiplies $\det (V_s)$ by at least $1+c_{\mathrm{len}}^2$. Since $\det (V_s)$ grows at most polynomially in $T$, at most $S=\widetilde O(d\log T)$ stages can occur. Summing the per-stage bound over $s\le S$ gives
\[
 \Reg(T)\;\lesssim\;\sqrt{m+1}\,C_\Delta\,B\sum_{s\le S}\beta_s
 \;\approx\;\sqrt{m+1}\,\Bigl(1+\frac{\Delta(a_{\mathrm{safe}})}{\Gamma}\Bigr)\sqrt d\,S^{3/2},
\]
recovering \Cref{thm:informal-main} up to lower-order warm-up terms. In what follows, Steps~1--2 establish the confidence geometry underlying $\|a\|_{V_s^{-1}}$, Steps~3--4 construct an optimistic candidate action whose gap is bounded by its uncertainty, Step~5 contracts this candidate toward $a_{\mathrm{safe}}$ to enforce~\ref{item:P1} while preserving~\ref{item:P2}, and Step~6 sets the stage length to satisfy~\ref{item:P3}. %

\begin{algorithm}[t]
\caption{QRS-COLTS}
\label{alg:qrs-colts}
\begin{algorithmic}[1]
\State Initialize $V_1=\lambda I_d$, $\widehat\theta_1=0$, $\widehat\Phi_1=0$, query counter $t=0$, stage $s=1$ (with $\beta_s=1+\sqrt{s-1}$ and $B=B(\delta)$).
\While{$t<T$}
    \State Set $h_s(a_{\mathrm{safe}})\gets\beta_s\|a_{\mathrm{safe}}\|_{V_s^{-1}}$, $M_s(a_{\mathrm{safe}})\gets Bh_s(a_{\mathrm{safe}})$, and $\underline\Gamma_s\gets\min_{i\in[m]}(\alpha_i-\langle\widehat\phi_{s,i},a_{\mathrm{safe}}\rangle)-h_s(a_{\mathrm{safe}})$.
    \If{$\underline\Gamma_s<3M_s(a_{\mathrm{safe}})$}
        \State $a_s\gets a_{\mathrm{safe}}$.
    \Else
        \State Draw $K_s$ coupled perturbations, solve their perturbed subproblems, and keep the feasible candidate $b_s$ with the highest sampled value.
        \State Scale $b_s$ toward $a_{\mathrm{safe}}$ via \eqref{eq:ucb_safe_constraint}; if no subproblem is feasible or no step passes, set $a_s\gets a_{\mathrm{safe}}$.
    \EndIf
    \State If $a_s=0$, play it for the remaining queries and stop. Otherwise set $n_s\gets\lceil c_{\mathrm{len}}/\|a_s\|_{V_s^{-1}}\rceil$ and compute the QME budget $q_s$ for accuracy $1/n_s$ and failure probability $\delta_s=6\delta/(\pi^2s^2)$.
    \State If the stage query cost $q_s$ exceeds $T-t$, spend the remaining $T-t$ queries at $a_s$ and stop.
    \State Otherwise run QME at $a_s$ using $q_s$ queries and update $V_{s+1}$, $\widehat\theta_{s+1}$, $\widehat\Phi_{s+1}$, $t\gets t+q_s$, and $s\gets s+1$.
\EndWhile
\end{algorithmic}
\end{algorithm}

\paragraph{Step 1: staged estimation and confidence widths.}
In each completed stage $j<s$, QME returns an estimate $\widehat\mu_j=(\widehat r_j,\widehat c_{j,1},\dots,\widehat c_{j,m})$ of $\mu(a_j)=(\ip{\theta^\star}{a_j},\Phi^\star a_j)$ with coordinate-wise error at most $1/n_j$, using $\widetilde O(\sqrt{m+1}\,n_j)$ coherent queries. For the reward entry, writing $\widehat r_j=\ip{\theta^\star}{a_j}+e_j$ with $|e_j|\le 1/n_j$ and multiplying by $n_j$ yields
\[
 n_j\widehat r_j=\ip{\theta^\star}{n_ja_j}+\xi_j,\qquad|\xi_j|\le 1,
\]
which is a linear observation at feature vector $n_ja_j$ with unit-bounded error. The exact same rescaling applies to each constraint estimate $n_j\widehat c_{j,i}=\ip{\phi_i^\star}{n_ja_j}+\xi_{j,i}$ with $|\xi_{j,i}|\le 1$. Performing ridge regression on these rescaled observations with regularization $\lambda>0$ gives
\begin{align*}
 V_s &= \lambda I_d+\sum_{j<s}n_j^2a_ja_j^\top,&
 \widehat\theta_s &= V_s^{-1}\sum_{j<s}n_j^2a_j\widehat r_j,&
 \widehat\phi_{s,i} &= V_s^{-1}\sum_{j<s}n_j^2a_j\widehat c_{j,i},
\end{align*}
where $\widehat\phi_{s,i}^\top$ forms the $i$-th row of $\widehat\Phi_s\in\RR^{m\times d}$. Effectively, a completed stage of accuracy $1/n_j$ contributes the statistical weight of $n_j^2$ unit-error classical observations at $a_j$ while consuming only $\widetilde O(\sqrt{m+1}\,n_j)$ quantum queries. Conditioned on all completed stage estimates meeting their $1/n_j$ accuracy targets, \Cref{lem:Con_explicit} yields simultaneous confidence bounds for all $s\ge 1$, $a\in\RR^d$, and $i\in[m]$:
\begin{equation}\label{eq:qrs-conf}
\begin{gathered}
 \bigl|\ip{\widehat\theta_s-\theta^\star}{a}\bigr|\le h_s(a),
 \qquad
 \bigl|\ip{\widehat\phi_{s,i}-\phi_i^\star}{a}\bigr|\le h_s(a),\qquad h_s(a):=\beta_s\|a\|_{V_s^{-1}},
\end{gathered}
\end{equation}
where $\beta_s:=\sqrt\lambda+\sqrt{s-1}$.
We refer to $h_s(a)$ as the \emph{confidence width} at $a$. Here, $\sqrt\lambda$ bounds the regularization bias under $\|\theta^\star\|_2,\|\phi_i^\star\|_2\le 1$, while $\sqrt{s-1}$ bounds the Euclidean norm of the $s-1$ rescaled stage errors: because QME errors need not be independent or zero-mean across stages, their worst-case norm grows as $\sqrt{s-1}$. Nevertheless, since $s\le S=\widetilde O(d\log T)$, we still have $\beta_s=\widetilde O(\sqrt{d\log T})$, matching the confidence radius order of classical linear bandits~\cite{abbasi2011linearbandits}. Throughout, we fix $\lambda=1$ so that the regularization bias is at most one normalized stage error and $\|a\|_{V_s^{-1}}\le\|a\|_2\le 1$ for all $a\in\cA$.

\paragraph{Step 2: vector feedback.}
Because all $m+1$ regression estimates share the design matrix $V_s$, stage $s$ requires accuracy $1/n_s$ across every coordinate of $\mu(a_s)$. Since a single invocation of $\mathcal O_{a_s}$ prepares the entire $(m+1)$-dimensional feedback vector coherently, the bounded multivariate estimator of \Cref{thm:isotropifier} achieves the simultaneous $\ell_\infty$-accuracy using $q_s=\widetilde O(\sqrt{m+1}\,n_s)$ queries (\Cref{lem:stage_estimator}). Setting the stage-$s$ QME failure probability to
\begin{equation}\label{eq:delta_s_def_main}
\delta_s=\frac{6\delta}{\pi^2s^2}
\end{equation}
ensures $\sum_{s\ge 1}\delta_s=\delta$, so by a union bound~\eqref{eq:qrs-conf} holds across all stages simultaneously with probability at least $1-\delta$.

\paragraph{Step 3: coupled linear subproblems.}
A natural deterministic UCB strategy would maximize $\langle\widehat\theta_s, a\rangle+h_s(a)$ over the optimistic constraint set $\{a\in\cA:\langle \widehat\phi_{s,i}, a\rangle-h_s(a)\le\alpha_i\ \forall i\in[m]\}$. However, subtracting the norm $h_s(a)$ makes these constraints nonconvex even when $\cA$ is a polytope, so we treat global UCB optimization solely as an intractable statistical benchmark (\Cref{sec:global-optimism-upper}). Following COLTS~\cite{gangrade2025colts}, QRS-COLTS instead linearizes the confidence bonus $h_s(a)=\beta_s\|V_s^{-1/2}a\|_2$ via a coupled Gaussian perturbation $\zeta\sim\mathcal N(0,I_d)$, adding $\beta_s\langle \zeta, V_s^{-1/2}a\rangle$ to the estimated reward and subtracting it from every estimated constraint:
\begin{equation}\label{eq:qrs-perturbation}
\widetilde\theta_s=\widehat\theta_s+\beta_s V_s^{-1/2}\zeta,
\qquad
\widetilde\Phi_s=\widehat\Phi_s-\beta_s\one_m\zeta^\top V_s^{-1/2},
\qquad\qquad
\max_{a\in\cA}\ \ip{\widetilde\theta_s}{a}
\ \ \text{s.t.}\ \
\widetilde\Phi_s a\le\alpha.
\end{equation}
The $i$-th row of $\widetilde\Phi_s$ is $\widetilde\phi_{s,i}^\top$ with $\widetilde\phi_{s,i}=\widehat\phi_{s,i}-\beta_sV_s^{-1/2}\zeta$. Whenever $\cA$ is a polytope, this perturbed subproblem is a standard linear program.

We say that a perturbation draw is \emph{optimistic} if $a^\star$ is feasible for its subproblem ($\widetilde\Phi_s a^\star\le\alpha$) and achieves the perturbed objective at least the optimal safe value ($\langle\widetilde\theta_s, a^\star\rangle\ge\mu^\star$). Sharing a single vector $\zeta$ across the objective and all $m$ constraints ensures that each draw is optimistic with constant probability independent of $m$: along $a^\star$, the perturbation shift equals $h_s(a^\star)\,Z$ with $Z\sim\mathcal N(0,1)$, so the single event $Z\ge 1$ simultaneously dominates the confidence errors~\eqref{eq:qrs-conf} in the reward and all $m$ constraints (\Cref{lem:local_opt}). Thus, optimism holds with probability at least $\pi_{\mathrm{opt}}:=\Pr[Z\ge 1]\approx 0.159$. In contrast, if we draw independent perturbations for each constraint, then the success probability will decay exponentially in $m$.

\paragraph{Step 4: resampling and the candidate gap.}
Because stage $s$ commits $q_s$ queries to one action, constant-probability optimism per draw is insufficient on its own. QRS-COLTS therefore draws $K_s$ independent perturbations $\zeta_s^{(1)},\dots,\zeta_s^{(K_s)}$, forms $(\widetilde\theta_s^{(k)},\widetilde\Phi_s^{(k)})$, and solves the $K_s$ subproblems
\begin{equation}\label{eq:qrs-subproblems}
b_s^{(k)}\in\argmax_{a\in\cA}\ip{\widetilde\theta_s^{(k)}}{a}
\quad\text{s.t.}\quad
\widetilde\Phi_s^{(k)}a\le\alpha
\qquad \forall k\in[K_s].
\end{equation}
Discarding infeasible subproblems, the learner selects the feasible solution $b_s$ maximizing $\langle\widetilde\theta_s^{(k)}, b_s^{(k)}\rangle$, and we denote its associated draw by $(\zeta_s,\widetilde\theta_s,\widetilde\Phi_s)$. Whenever at least one draw $k\in[K_s]$ is optimistic, the selection rule, optimality of $b_s^{(k)}$ for subproblem $k$, and optimism at $a^\star$ yield $\langle \widetilde\theta_s, b_s\rangle\ge\langle\widetilde\theta_s^{(k)},b_s^{(k)}\rangle\ge\langle\widetilde\theta_s^{(k)},a^\star\rangle\ge\mu^\star$. Choosing
\begin{equation}\label{eq:Ks_def_main}
K_s=\left\lceil \frac{1}{\pi_{\mathrm{opt}}}\log\frac{\pi^2s^2}{\delta}\right\rceil
\end{equation}
reduces the per-stage probability of missing optimism to $(1-\pi_{\mathrm{opt}})^{K_s}\le e^{-\pi_{\mathrm{opt}}K_s}\le\delta/(\pi^2s^2)$, which sums to $\delta/6<\delta$ over all stages (\Cref{lem:global_opt}).

Global optimism $\langle\widetilde\theta_s,b_s\rangle\ge\mu^\star$ immediately relates the suboptimality gap of $b_s$ to its confidence width:
\[
\Delta(b_s)=\mu^\star-\ip{\theta^\star}{b_s}
\ \le\ \ip{\widetilde\theta_s-\widehat\theta_s}{b_s}+\ip{\widehat\theta_s-\theta^\star}{b_s}
\ \le\ \bigl(\|\zeta_s\|_2+1\bigr)\,h_s(b_s),
\]
where we applied Cauchy--Schwarz to the perturbation term and~\eqref{eq:qrs-conf} to the estimation error. Given a deterministic upper bound $N_{\mathrm{Gauss}}$ on the total number of Gaussian draws across the run (\Cref{app:qrs-colts-alg}), Gaussian concentration and a union bound guarantee that with probability at least $1-\delta/2$, every draw satisfies $\|\zeta_s^{(k)}\|_2\le B$, where
\begin{equation}\label{eq:B_def_main}
B=B(\delta):=\sqrt d+\sqrt{2\log(2N_{\mathrm{Gauss}}/\delta)}
\end{equation}
(\Cref{lem:gauss_norm}). Defining the \emph{uncertainty proxy} $M_s(a):=B\,h_s(a)=B\beta_s\|a\|_{V_s^{-1}}$, which bounds the maximum shift induced at $a$ by any perturbation of norm at most $B$, we obtain $\Delta(b_s)\le(1+1/B)\,M_s(b_s)$ (\Cref{lem:candidate_gap}). Notice that the factor $B=\widetilde O(\sqrt d)$ is the statistical price of linearizing the confidence bonus: a vector $\zeta_s$ that achieves projection $\Omega(1)$ along $a^\star$ has Euclidean norm $\Theta(\sqrt d)$ in $d$ dimensions and can inflate the bonus along $b_s$ by $\sqrt d$, accounting for the gap between $d^2$ in \Cref{thm:informal-main} and $d^{3/2}$ in \Cref{thm:global-optimism-upper}.

\begin{figure}[t]
\centering
\begin{tikzpicture}[x=0.95cm,y=0.95cm,font=\small,
  lab/.style={fill=white,fill opacity=0.88,text opacity=1,inner sep=1.3pt,rounded corners=1pt},
  pt/.style={circle,draw=white,line width=0.7pt,inner sep=0pt,minimum size=6.4pt}]
  \def\safeset{(0,0.2)--(4.6,0)--(6.0,1.4)--(5.0,3.2)--(1.4,3.5)--(-0.4,2.0)--cycle}
  \def\pertset{(0.8,0.55)--(5.6,0.6)--(7.4,2.3)--(5.6,3.9)--(1.2,3.1)--cycle}
  \fill[qsafe!12] \safeset;
  \begin{scope}
    \clip \pertset;
    \fill[qpert!30,even odd rule] (-1,-1) rectangle (9,5) \safeset;
  \end{scope}
  \path[name path=cert,fill=qcert!22,draw=qcert!85!black,line width=1pt]
    plot[smooth cycle,tension=0.75] coordinates
      {(0.45,1.6) (1.0,0.8) (2.4,0.55) (3.7,0.95) (4.25,1.85) (3.55,2.65) (2.0,2.8) (0.8,2.45)};
  \draw[qsafe!75!black,line width=1.1pt,line join=round] \safeset;
  \draw[qpert!85!black,line width=1.1pt,dash pattern=on 4pt off 2.5pt,line join=round] \pertset;
  \draw[gray!65,line width=0.8pt,densely dotted] (8.06,0.41)--(6.74,4.19);
  \draw[-{Stealth[length=2.8mm]},line width=1pt,gray!55!black] (7.22,2.87)--++(0.9,0.315)
    node[right] {$\widetilde\theta_s$};
  \coordinate (u) at (1.6,1.7);
  \coordinate (b) at (7.4,2.3);
  \coordinate (astar) at (6.0,1.4);
  \path[name path=seg] (u)--(b);
  \path[name intersections={of=seg and cert,by=as}];
  \draw[gray!60!black,line width=0.9pt,dash pattern=on 2pt off 1.6pt] (as)--(b);
  \draw[qplay,line width=1.7pt] (u)--(as);
  \node[pt,fill=qcert!85!black] at (u) {};
  \node[pt,fill=qpert!90!black] at (b) {};
  \node[pt,fill=qsafe!70!black] at (astar) {};
  \node[pt,fill=qplay] at (as) {};
  \node[lab,below=3pt] at (u) {$a_{\mathrm{safe}}$};
  \node[lab,right=3pt] at (b) {$b_s$};
  \node[lab,below right=1pt] at (astar) {$a^\star$};
  \node[lab,above=4pt] at (as) {$a_s$};
  \node[text=qsafe!70!black] at (0.35,0.55) {$\cS$};
  \begin{scope}[shift={(9.55,3.55)},font=\footnotesize]
    \fill[qsafe!12] (0,0) rectangle (0.6,0.34);
    \draw[qsafe!75!black,line width=1.1pt] (0,0) rectangle (0.6,0.34);
    \node[right] at (0.75,0.17) {true safe set $\cS$};
    \fill[qcert!22] (0,-0.62) rectangle (0.6,-0.28);
    \draw[qcert!85!black,line width=1pt] (0,-0.62) rectangle (0.6,-0.28);
    \node[right] at (0.75,-0.45) {passes the conservative test~\eqref{eq:ucb_safe_constraint}};
    \draw[qpert!85!black,line width=1.1pt,dash pattern=on 4pt off 2.5pt] (0,-1.24) rectangle (0.6,-0.9);
    \node[right] at (0.75,-1.07) {feasible for the perturbed subproblem};
    \fill[qpert!30] (0,-1.86) rectangle (0.6,-1.52);
    \draw[qpert!85!black,line width=1.1pt,dash pattern=on 4pt off 2.5pt] (0,-1.86) rectangle (0.6,-1.52);
    \node[right] at (0.75,-1.69) {\dots\ and outside $\cS$, hence unsafe};
    \draw[gray!65,line width=0.8pt,densely dotted] (0,-2.31)--(0.6,-2.31);
    \node[right] at (0.75,-2.31) {level line of $\langle\widetilde\theta_s,\cdot\rangle$ through $b_s$};
    \draw[qplay,line width=1.7pt] (0,-2.93)--(0.6,-2.93);
    \node[right] at (0.75,-2.93) {actions $a_s(\rho)$ that pass the test};
  \end{scope}
\end{tikzpicture}
\caption{Geometry of a sampling stage (Steps~4 and~5). On the confidence event, the actions that pass the conservative test form a convex subset of the true safe set $\cS$ that contains $a_{\mathrm{safe}}$. The perturbed subproblem is solved over the dashed region, which contains $a^\star$ when the draw is optimistic and may extend beyond $\cS$, so its maximizer $b_s$ may be unsafe. The played action $a_s$ is the last point on the segment from $a_{\mathrm{safe}}$ to $b_s$ that passes the test.}
\label{fig:qrs-geometry}
\end{figure}
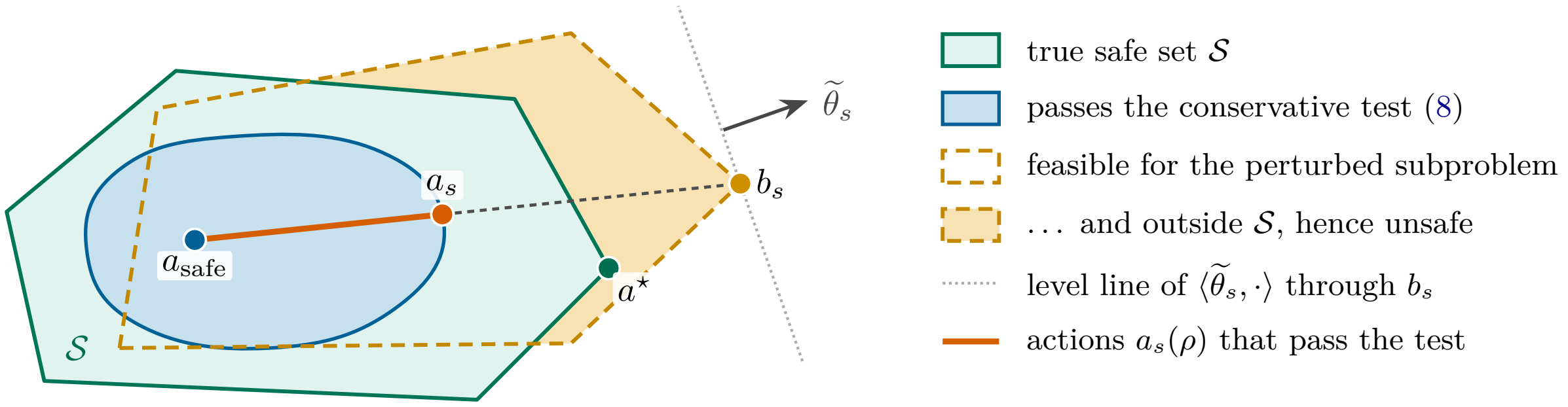

\paragraph{Step 5: conservative scaling and warm-up.}
Because $b_s$ only satisfies the optimistically perturbed constraints $\widetilde\Phi_sb_s\le\alpha$, playing $b_s$ directly could violate the true constraints by up to $(1+1/B)M_s(b_s)$. Instead, the learner enforces feasibility through the \emph{conservative test}
\begin{equation}\label{eq:ucb_safe_constraint}
\widehat\Phi_s a+\beta_s\|a\|_{V_s^{-1}}\one_m\le\alpha .
\end{equation}
By~\eqref{eq:qrs-conf}, $\langle \phi_i^\star, a\rangle\le\langle\widehat\phi_{s,i}, a\rangle+h_s(a)$ for every $i\in[m]$, so any action passing~\eqref{eq:ucb_safe_constraint} is guaranteed to lie in $\cS$ (\Cref{lem:safety}). QRS-COLTS therefore moves from the safe baseline $a_{\rm safe}$ toward $b_s$ as far as~\eqref{eq:ucb_safe_constraint} permits (\Cref{fig:qrs-geometry}):
\begin{equation}\label{eq:qrs-scaling}
a_s(\rho)=(1-\rho)\,a_{\rm safe}+\rho\,b_s,
\qquad
\rho_s=\max\bigl\{\rho\in[0,1]:a_s(\rho)\text{ passes~\eqref{eq:ucb_safe_constraint}}\bigr\},
\qquad
a_s=a_s(\rho_s).
\end{equation}
Since the left-hand side of~\eqref{eq:ucb_safe_constraint} is convex in $a$, the admissible values of $\rho$ form a closed interval and $\rho_s$ can be found efficiently by bisection (taking $a_s=a_{\rm safe}$ if all subproblems are infeasible or no $\rho\in[0,1]$ passes, neither of which occurs on the good event). In every case, $a_s$ is feasible, establishing~\ref{item:P1}.

To keep the reward loss from scaling small, the baseline $a_{\rm safe}$ must first have a certified safety margin large enough to absorb the excess of $b_s$. Specifically, our scaling analysis requires $3M_s(a_{\rm safe})\le\Gamma$. Because $\Gamma$ is unknown, the learner tracks it via the observable lower confidence bound
\begin{equation}\label{eq:baseline-slack-lcb}
 \underline{\Gamma}_s=\min_{i\in[m]}\left\{\alpha_i-\ip{\widehat\phi_{s,i}}{a_{\rm safe}}\right\}-h_s(a_{\rm safe}),
\end{equation}
which satisfies $\Gamma-2h_s(a_{\rm safe})\le\underline\Gamma_s\le\Gamma$ on the confidence event~\eqref{eq:qrs-conf}. 

Whenever $\underline\Gamma_s<3M_s(a_{\rm safe})$, the algorithm executes a \emph{warm-up stage}: it plays $a_s=a_{\rm safe}$ directly without solving any subproblems, which is safe by assumption and directly shrinks $\|a_{\rm safe}\|_{V_s^{-1}}$. 

Once $\underline\Gamma_s\ge 3M_s(a_{\rm safe})$ (a \emph{sampling stage}), we are guaranteed $3M_s(a_{\rm safe})\le\underline\Gamma_s\le\Gamma$, so $a_{\rm safe}$ has certified pessimistic slack of order $\Gamma$ in~\eqref{eq:ucb_safe_constraint}. Convexity along the segment from $a_{\rm safe}$ to $b_s$ then yields a lever rule: the weight $1-\rho_s$ placed on $a_{\rm safe}$ is at most the candidate's pessimistic excess, of order $M_s(b_s)$, divided by the baseline's certified slack, of order $\Gamma$, giving $(1-\rho_s)\Gamma\lesssim\rho_s M_s(b_s)$. Applying the triangle inequality to $\rho_s b_s=a_s-(1-\rho_s)a_{\rm safe}$ together with $3M_s(a_{\rm safe})\le\Gamma$ transfers $\rho_s M_s(b_s)$ onto the played action's uncertainty $M_s(a_s)$, establishing~\ref{item:P2} in the explicit form (\Cref{lem:scaling_generic,lem:stage_gap}): 
\begin{align*}
    \Delta(a_s)=(1-\rho_s)\Delta(a_{\rm safe})+\rho_s\Delta(b_s)\le(6+9\Delta(a_{\rm safe})/\Gamma)\,M_s(a_s).
\end{align*}

Moreover, the total query cost of warm-up stages is small. Every warm-up stage satisfies $\Gamma<(3+2/B)M_s(a_{\rm safe})$, which can hold only while $\|a_{\rm safe}\|_{V_s^{-1}}>\Gamma/(4B\beta_s)$. Each completed warm-up stage shrinks $\|a_{\rm safe}\|_{V_s^{-1}}^2$ by at least a factor of $1+c_{\mathrm{len}}^2$, so the inverse accuracies $n_s$ across warm-up stages grow geometrically and sum to the order of the final warm-up stage, consuming at most $\widetilde O(\sqrt{m+1}B\beta_S/\Gamma)$ queries in total (\Cref{lem:warmup}), including if $\beta_s$'s slow growth triggers additional warm-up stages later in the run.

\paragraph{Step 6: stage length, update, and stopping.}
Whenever $a_s\ne 0$, stage $s$ sets its target inverse accuracy to
\begin{equation}\label{eq:ns_def}
n_s=\left\lceil \frac{c_{\mathrm{len}}}{\norm{a_s}_{V_s^{-1}}}\right\rceil ,
\end{equation}
matching the estimation accuracy $1/n_s$ to the current directional uncertainty of $a_s$. This rule immediately yields $c_{\mathrm{len}}\le n_s\|a_s\|_{V_s^{-1}}\le c_{\mathrm{len}}+1$, establishing~\ref{item:P3}: the upper bound prevents the stage from spending more queries at $a_s$ than the gap bound~\ref{item:P2} can amortize, while the lower bound guarantees a constant factor increase in the determinant of the design matrix upon completion:
\[
V_{s+1}=V_s+n_s^2a_sa_s^\top,
\quad\text{so that}\quad
\det (V_{s+1})=\det (V_s)\,\bigl(1+n_s^2\|a_s\|_{V_s^{-1}}^2\bigr)\ \ge\ (1+c_{\mathrm{len}}^2)\,\det (V_s) .
\]
Because $\det (V_s)$ is bounded polynomially in $\sum_{j<s}n_j\le T/\sqrt{m+1}$, at most $S=\widetilde O(d\log T)$ stages can complete over the horizon (\Cref{lem:stage_count}). Any fixed constant $c_{\mathrm{len}}>0$ suffices, with larger values trading fewer stages against longer stage lengths; numerically, $c_{\mathrm{len}}\approx 5.48$ minimizes the leading prefactor (\Cref{cor:clen_opt}).

Finally, two boundary rules handle early termination. First, if $a_s=0$, then~\eqref{eq:ns_def} is undefined; however, $0$ cannot be played in a warm-up stage (since $a_{\rm safe}=0$ gives $M_s(a_{\rm safe})=0\le\underline\Gamma_s/3$), so~\ref{item:P2} certifies $\Delta(0)\le 0$, allowing the learner to play the optimal safe action $0$ for all remaining queries at zero regret. Second, if the stage QME budget $q_s$ exceeds the remaining horizon $T-t$, the learner spends the remaining $T-t$ queries at $a_s$ and terminates without recording a truncated estimate; otherwise, it executes all $q_s$ queries, updates $(V_{s+1},\widehat\theta_{s+1},\widehat\Phi_{s+1})$, and advances to stage $s+1$.

\paragraph{Where each factor comes from.}
Substituting $S=\widetilde O(d\log T)$ and $B=\widetilde O(\sqrt d)$ into 
$$
\sqrt{m+1}\,C_\Delta B\sum_{s\le S}\beta_s
$$ 
yields the leading rate 
$$
\widetilde O\left(\sqrt{m+1}\,(1+\Delta(a_{\rm safe})/\Gamma)\,d^2\log^{3/2}T\right),
$$
where each factor traces back to a single mechanism: $\sqrt{m+1}$ arises from multivariate QME over $(m+1)$-dimensional feedback (Step~2), $(d\log T)^{3/2}=S^{3/2}$ comes from summing the worst-case bounded-error confidence radii $\beta_s=\Theta(\sqrt s)$ (Step~1) across $S$ stages (Step~6), $\sqrt d$ comes from the Gaussian norm bound $B$ used to linearize the subproblems (Step~4), and $1+\Delta(a_{\rm safe})/\Gamma$ reflects conservative scaling toward $a_{\rm safe}$ (Step~5), which places weight $O(M_s(a_s)/\Gamma)$ on $a_{\rm safe}$ at per-unit cost $\Delta(a_{\rm safe})$, dominating the lower-order $\widetilde O(\sqrt{m+1}\,B\sqrt S\,\Delta(a_{\rm safe})/\Gamma)$ warm-up term.

This decomposition also clarifies the exact relationship to classical safe linear bandits. Replacing coherent QME by classical sample averaging increases the query cost of accuracy $1/n_s$ from $\widetilde O(n_s)$ to $\Theta(n_s^2)$, multiplying the stage-$s$ regret by $n_s$ to give $O(C_\Delta B\beta_s\,n_s)$. Under the classical budget constraint $\sum_s n_s^2\le T$, Cauchy--Schwarz gives $\sum_{s\le S}n_s\le\sqrt{ST}$, yielding total regret 
\begin{align*}
    O\left(C_\Delta B\beta_S\sqrt{ST}\right)=\widetilde O\left((1+\Delta(a_{\rm safe})/\Gamma)\,d^{3/2}\sqrt T\right),
\end{align*}
which matches the leading rate of classical S-COLTS~\cite{gangrade2025colts}. Coherent quantum estimation reduces the per-stage sample complexity from $n_s^2$ to $n_s$, exponentially improving the horizon dependence from $\sqrt T$ to $\operatorname{polylog}T$ while trading $d^{3/2}$ for $d^2$ to keep every stage subproblem linear.

\subsection{BCO-QLinTS}
\label{sec:bco-overview}

In the coherent-action bomb model (\Cref{def:intro-coherent-action-bomb}), the oracle computes the exact unsafe-action bit $f(a)=\one\{a\notin\cS\}$ into a measured flag qubit rather than returning noisy constraint samples. BCO-QLinTS (\Cref{alg:bco-qlints} in \Cref{app:bomb-projection}) exploits this flag as a membership oracle to optimize directly over the true safe set $\cS$, learning only the reward parameter $\theta^\star$. Throughout, $a_{\mathrm{safe}}\in\cS$ denotes the known safe baseline satisfying the well-roundedness condition $\mathbb B(a_{\mathrm{safe}},r)\subseteq\cS\subseteq\mathbb B(a_{\mathrm{safe}},R)$ (\Cref{ass:safe-body}).

\paragraph{Decoupling optimization and reward estimation.}
Recall from \Cref{def:query-weighted-regret-main} that a query with pre-query action marginal $\pi_t$ explodes with conditional probability $\pi_t(\cA\setminus\cS)$ and incurs query-weighted loss
\begin{align*}
    L(\pi_t)=\int_{\cS}\Delta_+(a)\,\pi_t(da)+\pi_t(\cA\setminus\cS)\in[0,1].
\end{align*}
BCO-QLinTS runs in at most $S_{\max}=\widetilde O(d\log T)$ stages and separates each stage $s$ into two phases with distinct roles in $R_{\mathrm w}(T)$:
\begin{enumerate}[label=(B\arabic*)]
\item\label{item:B1} \emph{Bomb-certified safe optimization.} The learner draws $K_s$ perturbed reward directions $\widetilde\theta_s^{(k)}$ and maximizes each over $\cS$ up to additive error $\gamma_s=T^{-2}$ using superposition bomb queries, keeping the total stage explosion probability at most $\eta_s=\varepsilon/S_{\max}$ and certifying that every returned candidate $a_s^{(k)}$ lies strictly in $\cS$. Because $L(\pi_t)\le 1$ on every query, this phase contributes at most its query count to $R_{\mathrm w}(T)$.
\item\label{item:B2} \emph{Classical-action reward estimation.} Selecting the pair $(\widetilde\theta_s,a_s)$ with the largest perturbed value $\langle\widetilde\theta_s^{(k)},a_s^{(k)}\rangle$, the learner queries the fixed safe action $a_s\in\cS$ for $q_s=L_sn_s$ rounds, with $n_s=\lceil c_{\mathrm{len}}/\|a_s\|_{V_s^{-1}}\rceil$ and $L_s=\Theta(\log(s/\delta))$, to estimate $\ip{\theta^\star}{a_s}$ to accuracy $1/n_s$. Because $\pi_t$ is then the point mass at $a_s\in\cS$, this phase has \emph{zero} explosion probability and incurs loss $L(\pi_t)=\Delta_+(a_s)$ per query.
\end{enumerate}
Summing the explosion budgets over $s\le S_{\max}$ bounds the cumulative explosion probability by $\sum_{s\le S_{\max}}\eta_s=\varepsilon$. Below, we explain how BCO-QLinTS implements~\ref{item:B1} without ever returning an unsafe action, and how~\ref{item:B1} and~\ref{item:B2} combine to give the two terms of \Cref{thm:informal-bomb-main}.

\paragraph{Exact-feasibility convex optimization via inward contraction.}
Suppose first that the learner had ideal, explosion-free coherent membership access $\ket{x}\ket{b}\mapsto\ket{x}\ket{b\oplus\one\{x\notin\cS\}}$. To maximize a unit vector $v=c/\|c\|_2$ over $\cS$, the quantum convex optimization algorithm of van Apeldoorn et al.~\cite{vanapeldoorn2020convexopt} uses $Q=\widetilde O(d\operatorname{polylog}(R/\tau,1/\delta))$ membership queries, but solves only \emph{weak} optimization: its output $y_0$ can lie up to distance $\tau/2$ outside $\cS$, which is inadmissible when feasibility must hold exactly.

We convert weak feasibility into exact membership in $\cS$ by contracting inward toward the interior ball $\mathbb B(u,r)\subseteq\cS$ (\Cref{lem:qco}). Rounding $y_0$ to a grid point $y$ within distance $\tau/4$ gives $\operatorname{dist}(y,\cS)\le\tau$. Pulling $y$ slightly toward $u$ via $p=\sigma y+(1-\sigma)u$ with $\sigma=r/(r+2\tau)$ mixes the nearest point $z\in\cS$ to $y$ with $\mathbb B(u,r)$, placing an entire ball $\mathbb B(p,\sigma\tau)$ inside $\cS$ by convexity. Rounding $p$ to a grid label $\widehat a$ within $\tau/4<\sigma\tau$ then guarantees $\widehat a\in\cS$, while losing at most $O(\tau(1+R/r)\|c\|_2)$ in objective value. Choosing $\tau\asymp\gamma/((1+R/r)\|c\|_2)$ achieves additive error $\gamma$ using $Q=\widetilde O(d\operatorname{polylog}(R/r,\|c\|_2/\gamma,1/\delta))$ ideal membership queries.

\paragraph{Controlled bomb simulation and Zeno safety certification.}
Replacing ideal membership queries by the safe-action bomb oracle (\Cref{def:safe-action-bomb-oracle}) raises two obstacles: the oracle has no external control wire and entangles the action with a reward qsample $\ket{\chi_a}_X$, and a simulated run of the optimizer can still output an unsafe point.

First, after factoring out out-of-domain grid labels reversibly via $\one\{x\notin\cS\}=\one\{x\notin\cA\}\oplus\one\{x\in\cA\setminus\cS\}$, two calls to \Cref{def:safe-action-bomb-oracle} implement one \emph{controlled} bomb query to $h(x)=\one\{x\in\cA\setminus\cS\}$ on any basis state $\ket{x}\ket{e}_{\mathrm{ctrl}}$: we reversibly write $a_{\mathrm{aux}}=x$ into an ancilla register when $(e=1\text{ and }x\in\cA)$ and $a_{\mathrm{aux}}=u$ otherwise, apply one forward bomb query on $a_{\mathrm{aux}}$, and measure the flag. Branches with $e=0$ or $x\notin\cA$ query $u\in\cS$ and survive with certainty, so an explosion occurs if and only if $e=1$ and $x\in\cA\setminus\cS$. Crucially, on the surviving state every branch satisfies $a_{\mathrm{aux}}\in\cS$, so one inverse bomb call on $a_{\mathrm{aux}}$ uncomputes $\ket{\chi_{a_{\mathrm{aux}}}}_X$ back to $\ket{0}_X$ with zero explosion risk.

Second, given a per-call explosion budget $\eta_{\mathrm{call}}=\eta_s/K_s$, the bomb-query simulation of Lin and Lin~\cite{lin2014bomb} (\Cref{thm:linlin-simulation}) simulates the $Q$-query optimizer across $J=\lceil\log_2(4/\delta_{\mathrm{call}})\rceil$ trials using $O(Q^2 J/\eta_{\mathrm{call}})$ controlled bomb queries per trial. A surviving trial can still output an infeasible action $a\in\cA\setminus\cS$: the optimizer is run with a constant failure probability, and the surviving simulated state differs from the ideal output state by at most $\eta_{\mathrm{call}}/(5J)$ in Euclidean norm. Passing such an action to~\ref{item:B2} would trigger an immediate explosion. We eliminate false positives via an Elitzur--Vaidman repeated-rotation test~\cite{Elitzur:1993xh,lin2014bomb} (\Cref{lem:bomb-safe-opt}): starting a control qubit in $\ket{0}$, each of $N=\lceil\pi^2 J/(2\eta_{\mathrm{call}})\rceil$ rounds rotates the qubit by $\vartheta=\pi/(2N)$ and performs a controlled bomb query at $a$. If $a\in\cS$, all $N$ rounds survive and rotate $\ket{0}\mapsto\ket{1}$, accepting $a$ with certainty; if $a\notin\cS$, each surviving flag measurement projects the control qubit back to $\ket{0}$ (the quantum Zeno effect), so $a$ is accepted with probability \emph{zero} while exploding with probability at most $N\sin^2\vartheta\le N\vartheta^2\le\eta_{\mathrm{call}}/(2J)$. Taking the best accepted candidate (with fallback $u$) guarantees exact safety $\widehat a\in\cS$ on every return.

\paragraph{Reward learning and where each factor comes from.}
In phase~\ref{item:B2}, BCO-QLinTS estimates $\ip{\theta^\star}{a_s}$ to accuracy $1/n_s$ via scalar QMC (\Cref{lem:qmc_scalar}), using $q_s=L_sn_s$ reward queries at $a_s\in\cS$; the factor $L_s=\Theta(\log(s/\delta))$ absorbs the QMC failure-probability logarithm, so the estimation fits deterministically inside $q_s$ queries. As in QRS-COLTS, rescaling observation $j$ by $n_j$ bounds its error by one, yielding the reward confidence radius $\beta_s=1+\sqrt{s-1}$, and under $K_s$ resamples of $\widetilde\theta_s^{(k)}=\widehat\theta_s+\beta_s V_s^{-1/2}\zeta_s^{(k)}$, the stage gap bound $\Delta_+(a_s)\le(B+1)\beta_s\|a_s\|_{V_s^{-1}}+\gamma_s$ (\Cref{prop:bco-resampled-optimism,lem:bco-stage-gap}).

Just as in Section~\ref{sec:qrs-overview}, $n_s\|a_s\|_{V_s^{-1}}\le c_{\mathrm{len}}+1$ cancels $\|a_s\|_{V_s^{-1}}$ in the reward phase and limits the run to $S_{\max}=\widetilde O(d\log T)$ stages (\Cref{thm:bco-qlints}). Summing the reward-estimation regret across stages gives 
\begin{align*}
    (c_{\mathrm{len}}+1)(B+1)\sum_{s\le S_{\max}}L_s\beta_s+T\gamma_s=\widetilde O\left(d^2\log^{3/2}(2T/\delta)\right).
\end{align*}
Meanwhile, in stage $s$, each of the $K_s$ bomb-certified optimizer calls uses $Q=\widetilde O(d)$ ideal queries at explosion budget $\eta_{\mathrm{call}}=\varepsilon/(S_{\max}K_s)$, costing $\widetilde O(Q^2 S_{\max}K_s/\varepsilon)$ bomb queries per call; summing over all $K_s$ calls and all $S_{\max}$ stages yields
\begin{align*}
    \widetilde O\left(d^2 S_{\max}^2 K_{S_{\max}}^2/\varepsilon\right)=\widetilde O\left((d^4\log^2(2T)/\varepsilon)\operatorname{polylog}(2+R/r,2/\delta)\right),
\end{align*}
where the $m$ constraints enter solely through the geometric ratio $R/r$.

\section{Discussion and open questions}
\label{sec:discussion}

We view this work as a first step toward quantum algorithms for safe online learning. With classical actions and coherent feedback, QRS-COLTS preserves round-wise mean safety while improving the horizon dependence from $\sqrt{T}$ to $\operatorname{polylog} T$. With coherent actions and a bomb oracle, BCO-QLinTS bypasses learning the $m$ constraint rows altogether, trading constraint estimation for a bomb-certified optimization overhead.

In the classical-action, coherent-feedback model, \Cref{thm:global-optimism-upper,thm:convex-quantum-lower} pin down the minimax polynomial dependence on $d$ at $\widetilde\Theta(d^{3/2})$ for unrestricted classical optimization when $m$ and the positive baseline slack $\Gamma$ are fixed and $T\ge d^{3/2}$ (setting $\delta=1/(2T)$ in \Cref{thm:global-optimism-upper} and using $\Reg(T)\le T$ on the failure event to compare expected regret). Two natural questions remain in this regime: whether a computationally efficient algorithm based on linear subproblems can match this $d^{3/2}$ dependence, and whether the $\log^{3/2} T$ horizon factor can be sharpened to $\log T$.

Regarding the constraint parameters, \Cref{thm:slack-lower} shows that already for a single constraint ($m=1$), round-wise safety forces a regret term of order $\Delta(a_{\mathrm{safe}})/\Gamma$ once $T\gtrsim 1/\Gamma$. However, it remains open whether this inverse-slack cost must multiply the dimension and horizon factors as in our scaling analysis, and whether the $\sqrt{m+1}$ overhead from multivariate mean estimation is necessary for safe bandit regret.

In the coherent-action bomb model, analogous open questions include tightening the dimension dependence of the reward-learning term, establishing lower bounds under query-weighted regret, and determining whether direct bomb-safe optimization can improve upon the $\widetilde O(d^4\log^2 T/\varepsilon)$ query overhead in \Cref{thm:informal-bomb-main}.

Beyond tightening these rates, natural directions include studying hybrid quantum feedback models (such as noisy safety flags or coherent rewards paired with classical constraint observations) and extending safe quantum exploration to contextual bandits, knapsack constraints, Bayesian optimization, and reinforcement learning.

\section*{Acknowledgements}
Guang Lin gratefully acknowledges the support from DARPA MATHBAC program under grant number: HR0011-26-3-E152, the National Science Foundation (DMS-2533878, DMS-2053746, DMS-2134209, ECCS-2328241, CBET-2347401 and OAC-2311848), and U.S.~Department of Energy (DOE) Office of Science Advanced Scientific Computing Research program under the ``Uncertainty Quantification for Multifidelity Operator Learning (MOLUcQ)'' project (Project No. 81739), DE-SC0023161, the SciDAC LEADS Institute, and DOE--Fusion Energy Science, under grant number: DE-SC0024583.

This version was prepared with the help of the following AI tools: GPT-6 Astra Pro, Claude Fable 5.2, Claude Opus 5.5, and Gemini 4 Carbon. GPT-6 Astra Pro, under the authors' guided prompting, proposed the new results in~\cref{sec:global-optimism-upper,sec:classical-action-lower-bounds} and the removal of the assumption of a known lower bound on the baseline slack $\Gamma\geq \Gamma_0$ (see~\cref{thm:main}). The authors also used these AI tools to assist with plotting the figures, improving the presentation of the proofs, checking calculations, and ensuring consistent notation. All mathematical statements, proofs, and references were independently verified by the authors, who take full responsibility for the content of this manuscript.
\bibliographystyle{alpha}
\bibliography{ref}

\newpage
\appendix

\section*{Appendix}
\paragraph{Roadmap.}
\Cref{sec:models} defines the three models of the main text: the classical-action coherent-feedback model, the classical-action classical-feedback model, and the coherent-action bomb model with query-weighted regret. \Cref{app:qrs-colts-details} states the stage estimator and the parameters of QRS-COLTS, whose steps are described in \Cref{sec:qrs-overview}, and proves its safety and regret guarantees; it then proves the upper bound with exact global optimization (\Cref{sec:global-optimism-upper}) and the lower bounds for classical actions (\Cref{sec:classical-action-lower-bounds}). \Cref{app:bomb-projection} gives the bomb-certified optimization routine, the BCO-QLinTS algorithm, and the proofs of its explosion-probability and regret guarantees.

\section{Quantum models for safe stochastic linear bandits}\label{sec:models}

In this section, we provide formal details for the quantum action and feedback models studied in this paper.

\subsection{Classical-action coherent-feedback model}\label{sec:classical-action-coherent}

Our first positive result uses the coherent quantum reward-oracle model of Wan et al.~\cite{wan2023quantumbandits}, extended to vector-valued reward-and-constraint feedback. This model preserves standard online bandit semantics: before every query, the learner commits to a single classical action, and only the access to the feedback distribution of that action is quantum.

For each action $a\in\cA$, let $\Omega_a$ be a latent outcome space and let $p_a$ be a probability distribution over $\Omega_a$. A classical draw $\omega\sim p_a$ produces the feedback vector
$$
X(a,\omega)=(R(a,\omega),C_1(a,\omega),\ldots,C_m(a,\omega))\in[0,1]^{m+1},
$$
with expectation
$$
\mu(a)=\EE_{\omega\sim p_a}[X(a,\omega)]=(\ip{\theta^\star}{a},\Phi^\star a).
$$
We assume that feedback values are given in a known binary numerical encoding on which the learner can perform reversible arithmetic (if real-valued outputs in $[0,1]^{m+1}$ are instead discretized in $[0,1]^{m+1}$ to a requested precision, combining a coordinate-wise rounding error of at most $1/(2n_s)$ with a QME target error of at most $1/(2n_s)$ increases the stage query count by only a constant factor and achieves total stage accuracy $1/n_s$ for the original mean in \Cref{app:qrs-colts-details}). The learner knows neither $p_a$ nor the map $\omega\mapsto X(a,\omega)$, accessing them solely through the unitary oracle below. We write $\mathcal D_a$ for the law of $X(a,\omega)$ under $\omega\sim p_a$.

\begin{definition}[Classical-action coherent-feedback oracle]\label{def:q_oracle}
For each $a\in\cA$, the learner has coherent access to a unitary $\mathcal O_a$ and its inverse $\mathcal O_a^\dagger$, where
$$
\mathcal O_a\ket{0}\ket{0}
=
\sum_{\omega\in\Omega_a}\sqrt{p_a(\omega)}\,\ket{\omega}\ket{X(a,\omega)}.
$$
Each invocation of $\mathcal O_a$ or $\mathcal O_a^\dagger$ counts as one query, and the action $a$ is chosen classically prior to the query.
\end{definition}

Measuring the feedback register of $\mathcal O_a\ket{0}\ket{0}$ in the computational basis yields a single classical sample from $\mathcal D_a$. The advantage of this model is that the learner may apply $\mathcal O_a$ and $\mathcal O_a^\dagger$ coherently within quantum mean-estimation subroutines prior to measurement. Cumulative regret remains the standard played-action regret $\Reg(T)=\sum_{t=1}^T\Delta(a_t)$, and round-wise mean safety requires $\Phi^\star a_t\le\alpha$ for every queried action $a_t$. Thus, this model accelerates statistical estimation while preserving the standard safe-bandit decision protocol.

\subsection{Classical-action classical-feedback model}\label{sec:classical-action-noncoherent}

In the second model, actions stay classical and the learner has no coherent access to the feedback randomness.

\begin{definition}[Classical-action classical-feedback model]\label{def:noncoherent-feedback}
At each query, the learner chooses a classical action $a\in\cA$ and observes a single classical sample $X(a,\omega)$ with $\omega\sim p_a$ (equivalently, the learner queries a quantum channel whose output is immediately measured in the standard basis, precluding reversible unitary access to the sampling randomness). Regret and safety follow the standard played-action definitions in \Cref{sec:intro}.
\end{definition}

This model subsumes the classical stochastic safe linear bandit problem. Without reversible access to the feedback randomness, quantum algorithms cannot achieve the $\widetilde O(1/\varepsilon)$ mean-estimation rate~\cite{buchholz2025channel}, making this model our classical statistical baseline (see \Cref{thm:classical-lower}).

\subsection{Coherent-action safe-action bomb model and query-weighted regret}\label{sec:coherent-action-bomb-model}

The third model allows the learner to query superpositions of actions and provides exact membership access to the true safe set, at the risk of terminating the algorithm if an unsafe action is queried. Because a superposition query places amplitude across multiple actions simultaneously, there is no single played action per round, motivating a query-weighted notion of regret.

Let
$$
\cS:=\{a\in\cA:\Phi^\star a\le\alpha\}
$$
denote the true safe set, and define the unsafe-action indicator
$$
f(a):=\one\{a\notin\cS\}=\one\{\Phi^\star a\nleq\alpha\}.
$$
Rather than returning noisy constraint samples, the oracle reports feasibility directly through a binary flag encoding $f(a)$, which is measured immediately after every query.

For each action $a\in\cA$, let $W_a$ be a unitary preparing the coherent reward qsample
$$
W_a\ket{0}_X = \ket{\chi_a}_X:=\sum_{\omega\in\Omega_a}\sqrt{p_a(\omega)}\ket{\omega}\ket{R(a,\omega)},
$$
so that $\EE_{\omega\sim p_a}[R(a,\omega)]=\ip{\theta^\star}{a}$, and let $U_{\mathrm{rew}}:=\sum_{a}\ket{a}\!\bra{a}\otimes W_a$ be its action-controlled counterpart.

\begin{definition}[Safe-action bomb oracle]\label{def:safe-action-bomb-oracle}
On blank output registers, a forward call to the safe-action bomb oracle maps
$$
    \ket{a}\ket{0}_f\ket{0}_X
    \mapsto
    \ket{a}\ket{f(a)}_f\ket{\chi_a}_X .
$$
More generally, a forward (resp.\ inverse) query applies $U_{\mathrm{rew}}$ (resp.\ $U_{\mathrm{rew}}^\dagger$) to the action and reward registers, XORs $f(a)$ into a fresh flag qubit initialized to $\ket{0}_f$, and immediately measures the flag qubit in the computational basis. Measurement outcome $1$ triggers a bomb explosion and terminates the algorithm immediately, whereas outcome $0$ projects the action register onto the safe subspace $\mathrm{span}\{\ket{a}:a\in\cS\}$ after the reward operation.
\end{definition}

To define regret, let $\rho_t$ be the joint pre-query density operator at query $t$ across the action register and any internal work registers. Its action marginal is the probability measure $\pi_t$ on $\cA$ given by
$$
\pi_t(E)=\mathrm{Tr}\!\left[(\Pi_E\otimes I)\rho_t\right]
$$
for measurable $E\subseteq\cA$, where $\Pi_E$ projects the action register onto $E$. On a finite action set, a pure state of the form $\sum_{a\in\cA}\psi_{t,a}\ket a\ket{\mathrm{work}_a}$ with normalized work states has action marginal $\pi_t(\{a\})=|\psi_{t,a}|^2$.

\begin{definition}[Query-weighted regret and explosion probability]\label{def:query-weighted-regret-main}
For an action $a\in\cA$, let $\Delta_+(a):=(\mu^\star-\ip{\theta^\star}{a})_+$ denote the clipped gap, where $\mu^\star=\ip{\theta^\star}{a^\star}$. The per-query weighted loss in the coherent-action bomb model is
$$
L(\pi_t):=\int_{\cS}\Delta_+(a)\,\pi_t(da)+\pi_t(\cA\setminus\cS),
$$
and the cumulative query-weighted regret is
$$
R_{\mathrm w}(T):=\sum_{t=1}^T L(\pi_t).
$$
Here $\pi_t$ is conditioned on the history prior to query $t$ and on surviving up to query $t$, so $\pi_t(\cA\setminus\cS)$ is the conditional explosion probability at query $t$. The sum includes the terminating query (if an explosion occurs), with $L(\pi_t)=0$ for all subsequent steps.
\end{definition}

Assigning unit loss to unsafe probability mass matches the $[0,1]$ scale of the rewards. Whenever $\pi_t$ is a point mass on a safe action $a_t\in\cS$, we have $L(\pi_t)=\Delta_+(a_t)=\Delta(a_t)$, recovering standard played-action regret. In general, however, this model is incomparable to the classical-action model, as it grants coherent superposition access over $\cA$ and exact feasibility flags, which BCO-QLinTS (\Cref{sec:bomb-main}) exploits to optimize directly over $\cS$.

\section{Algorithms and lower bounds for classical actions}
\label{app:qrs-colts-details}

This appendix provides the full algorithmic specification and analysis of QRS-COLTS, establishing the classical-action coherent-feedback guarantee stated informally in \Cref{thm:informal-main}, together with an optimistic global-optimization upper bound and matching lower bounds for classical action selection. The algorithm is described step by step in \Cref{sec:qrs-overview}. Here we state its stage estimator, its guarantee, and its parameters (\Cref{app:qrs-colts-alg}), give the structure of the proof (\Cref{app:qrs-colts-proof-overview}), and present the detailed safety and regret proofs (\Cref{sec:proofs}). We then analyze the global-optimization variant achieving $d^{3/2}$ dimension dependence (\Cref{sec:global-optimism-upper}) and prove lower bounds for both coherent and classical feedback (\Cref{sec:classical-action-lower-bounds}).

\subsection{Algorithm and main guarantee}\label{app:qrs-colts-alg}

Throughout this section, $\cA\subseteq\RR^d$ is compact and convex with $\|a\|_2\le 1$ for all $a\in\cA$, $\mu(a)=(\ip{\theta^\star}{a},\Phi^\star a)$ is the mean feedback of action $a$, $u=a_{\mathrm{safe}}$, and $\Gamma=\min_{i\in[m]}(\alpha_i-\ip{\phi_i^\star}{u})>0$. \Cref{sec:qrs-overview} describes QRS-COLTS (\Cref{alg:qrs-colts}) step by step and gives the reason for each parameter; ``Step~$k$'' and the properties~\ref{item:P1}--\ref{item:P3} below refer to that description. This subsection states the properties of the stage estimator and the guarantee, fixes the constants that must be set before the run, and collects all parameters in \Cref{tab:qrs-parameters}.

\paragraph{Stages and histories.}
Stage $s$ is a \emph{warm-up stage} if $\underline\Gamma_s<3M_s(u)$ and a \emph{sampling stage} otherwise (Step~5). A stage is \emph{completed} if all $q_s$ of its queries are executed. We write $S$ for the number of stages begun by query $T$; stages $1,\dots,S-1$ are completed, and $S\le T$ because every stage uses at least one query. The history $\cF_{s-1}$ consists of the actions, estimates and Gaussian vectors of stages $1,\dots,s-1$; it determines $V_s$, $\widehat\theta_s$, $\widehat\Phi_s$ and whether stage $s$ is a warm-up stage. As in Step~1, $\lambda=1$ throughout, so $V_s\succeq I$; a formula keeps $\lambda$ where this shows the role of the regularizer.

\paragraph{The stage estimator.}
In stage $s$, the learner runs the multivariate estimator of \Cref{thm:isotropifier} at $a_s$ with accuracy $1/n_s$ and failure probability $\delta_s$, and obtains an estimate $\widehat\mu_s=(\widehat r_s,\widehat c_{s,1},\dots,\widehat c_{s,m})$ of $\mu(a_s)$. The analysis uses only the following two properties of this estimator.

\begin{lemma}[Stage estimator]\label{lem:stage_estimator}
Conditioned on the history up to the choice of $a_s$, the estimate satisfies
$$
\norm{\widehat\mu_s-\mu(a_s)}_\infty\le \frac{1}{n_s}
$$
with probability at least $1-\delta_s$, where $n_s$ and $\delta_s$ are given by~\eqref{eq:ns_def} and~\eqref{eq:delta_s_def_main}. The stage uses $q_s$ queries, where $q_s$ is a function of $m$, $n_s$, $s$ and $\delta$ with
$$
\sqrt{m+1}\,n_s\ \le\ q_s\ \le\ \sqrt{m+1}\,n_s\,L_s,\qquad L_s:=L(n_s,s),
$$
for a function $L(n,s)\ge 1$ that is polylogarithmic in $m$, $n$, $s$ and $1/\delta$.
\end{lemma}

\begin{proof}
By \Cref{thm:isotropifier} with $Z=X(a_s)$, $M=m+1$, $\varepsilon=1/n_s$ and failure probability $\delta_s=6\delta/(\pi^2s^2)$, the estimator has the stated accuracy and uses at most $\sqrt{m+1}\,n_s\,P(n_s,s)$ queries in the worst case, where $P(n,s)\ge 1$ is polylogarithmic in $m$, $n$, $s$ and $1/\delta$. Let $q_s=\lceil\sqrt{m+1}\,n_s\,P(n_s,s)\rceil$ and spend any unused calls at $a_s$. Since $\sqrt{m+1}\,n_s\ge 1$, both bounds hold with $L(n,s)=P(n,s)+1$.
\end{proof}

The upper bound on $q_s$ limits the regret of a stage, and the lower bound gives $\sum_sn_s\le T/\sqrt{m+1}$ over the completed stages. Since $q_s$ is known once $a_s$ is chosen, the learner can check before the first query of a stage whether the stage fits in the remaining horizon. We write $\bar L:=\max\{L(n,s):n,s\le T\}$, which is polylogarithmic in $m$, $T$ and $1/\delta$. %

\begin{theorem}[Round-wise mean safety and polylogarithmic query regret]\label{thm:main_preview}
Fix $\delta\in(0,1)$ and run QRS-COLTS for $T$ coherent oracle queries. Suppose $\norm{\theta^\star}_2\le 1$, $\norm{\phi_i^\star}_2\le 1$ for all $i\in[m]$, and there is a known strictly safe action $a_{\mathrm{safe}}\in\cA$ with true baseline slack $\Gamma:=\min_{i\in[m]}(\alpha_i-\langle\phi_i^\star,a_{\mathrm{safe}}\rangle)>0$. Then, with probability at least $1-3\delta$, QRS-COLTS is safe at every query and satisfies
$$
\Reg(T)
=
\widetilde O\!\left(
\sqrt{m+1}\,d^2\,\log^{3/2}(2T/\delta)
\left(1+\frac{\Delta(a_{\mathrm{safe}})}{\Gamma}\right)
\right),
$$
where $\widetilde O(\cdot)$ hides constants depending on the fixed $c_{\mathrm{len}}$ and additional logarithmic factors in $d,m,T$, and $1/\delta$. This bound is the last display of \Cref{thm:main}, which keeps the stage-count, perturbation-norm and warm-up terms separate.
\end{theorem}

\paragraph{Constants fixed before the run.}
The warm-up test uses the value of $B$, so the number of Gaussian draws must be bounded before the first stage. Let $S_{\max}$ be a deterministic bound on the number of stages. Since $K_s\le 1+\pi_{\mathrm{opt}}^{-1}\log(\pi^2s^2/\delta)$, the run makes at most
\begin{equation}\label{eq:NGauss_def_main}
 \sum_{s\le S_{\max}}K_s
 \ \le\ S_{\max}+\frac{S_{\max}}{\pi_{\mathrm{opt}}}\log\frac{\pi^2S_{\max}^2}{\delta}
 \ \le\ \frac{3S_{\max}}{\pi_{\mathrm{opt}}}\log\frac{\pi S_{\max}}{\delta}=:N_{\mathrm{Gauss}}
\end{equation}
draws, where the last step uses $\delta<1$ and $\pi_{\mathrm{opt}}<1<\log(\pi S_{\max}/\delta)$. This value in~\eqref{eq:B_def_main} gives $B=\sqrt d+O(\sqrt{\log(2S_{\max}/\delta)})$. We take
\begin{equation}\label{eq:qrs-stage-budget}
 S_{\max}=\min\!\left\{T,\;1+\left\lceil
 \frac{d\log(1+T^2/d)}{\log(1+c_{\mathrm{len}}^2)}
 \right\rceil\right\}.
\end{equation}
Indeed, every stage uses at least one query, so there are at most $T$ stages. Moreover $q_s\ge n_s$ by \Cref{lem:stage_estimator}, so the completed stages satisfy $\sum_sn_s\le T$, \Cref{lem:stage_count} bounds their number by the ceiling in~\eqref{eq:qrs-stage-budget} (with $\lambda=1$, using $d_{\cA}\le d$ and that $x\mapsto x\log(1+T^2/x)$ is increasing), and at most one further stage can begin before the horizon is reached.

\begin{table}[t]
\centering
\small
\renewcommand{\tabularxcolumn}[1]{p{#1}}
\renewcommand{\arraystretch}{1.08}
\caption{Parameters of QRS-COLTS and the reason for each choice.}
\label{tab:qrs-parameters}
\begin{tabularx}{\textwidth}{@{}ll>{\raggedright\arraybackslash}X@{}}
\toprule
Quantity & Value & Reason for the choice\\
\midrule
$\lambda$ & $1$ & ridge regularization; for parameters of norm at most one, the bias is at most one stage error, and $V_s\succeq I$\\ \addlinespace[3pt]
$\beta_s$ & $\sqrt\lambda+\sqrt{s-1}$ & confidence radius: regularization bias plus the norm of $s-1$ rescaled errors of size at most one\\ \addlinespace[3pt]
$h_s(a)$ & $\beta_s\|a\|_{V_s^{-1}}$ & confidence width of $a$ in~\eqref{eq:qrs-conf}\\ \addlinespace[3pt]
$\delta_s$ & $6\delta/(\pi^2s^2)$ & failure probability of the stage-$s$ estimate; $\sum_s\delta_s=\delta$\\ \addlinespace[3pt]
$\pi_{\mathrm{opt}}$ & $\Pr[Z\ge 1]\approx 0.159$ & lower bound on the probability that one coupled draw is optimistic\\ \addlinespace[3pt]
$K_s$ & $\lceil\pi_{\mathrm{opt}}^{-1}\log(\pi^2s^2/\delta)\rceil$ & number of draws; none is optimistic with probability at most $\delta/(\pi^2s^2)$\\ \addlinespace[3pt]
$B$ & $\sqrt d+\sqrt{2\log(2N_{\mathrm{Gauss}}/\delta)}$ & bound on $\|\zeta\|_2$ for all draws of the run, with probability $1-\delta/2$\\ \addlinespace[3pt]
$M_s(a)$ & $B\,h_s(a)$ & largest change that a perturbation with $\|\zeta\|_2\le B$ causes at $a$\\ \addlinespace[3pt]
$\underline\Gamma_s$ & see~\eqref{eq:baseline-slack-lcb} & computable lower bound on the unknown slack $\Gamma$\\ \addlinespace[3pt]
warm-up test & $\underline\Gamma_s<3M_s(u)$ & a sampling stage ($\underline\Gamma_s\ge 3M_s(u)$) has $3M_s(u)\le\Gamma$, as the scaling argument requires\\ \addlinespace[3pt]
$n_s$ & $\lceil c_{\mathrm{len}}/\|a_s\|_{V_s^{-1}}\rceil$ & accuracy matched to uncertainty: bounded regret per stage and constant growth of $\det (V_s)$\\ \addlinespace[3pt]
$c_{\mathrm{len}}$ & any constant $>0$ & larger values give fewer and longer stages; $c_{\mathrm{len}}\approx 5.48$ minimizes the leading constant\\ \addlinespace[3pt]
$S_{\max}$, $N_{\mathrm{Gauss}}$ & see~\eqref{eq:qrs-stage-budget},~\eqref{eq:NGauss_def_main} & deterministic bounds on the numbers of stages and draws, needed to fix $B$ before the run\\
\bottomrule
\end{tabularx}
\end{table}

\subsection{Proof overview}\label{app:qrs-colts-proof-overview}

\begin{figure}[!t]
\centering
\begin{tikzpicture}[x=1cm,y=1cm,font=\small,
  box/.style={draw,rounded corners=3pt,align=center,inner sep=5pt,line width=0.8pt,minimum height=1.05cm},
  prop/.style={box,text width=3.7cm},
  wide/.style={box,text width=12.6cm},
  arr/.style={-{Stealth[length=2.2mm]},line width=0.8pt,gray!60!black}]
  \node[wide,draw=gray!60!black,fill=gray!8] (g) at (0,0)
    {\textbf{Good event} $\mathcal G=$ \ref{item:E1}${}\cap{}$\ref{item:E2}${}\cap{}$\ref{item:E3}, probability at least $1-5\delta/3$\\
     \Cref{lem:Con_explicit,lem:gauss_norm,lem:local_opt,lem:global_opt}};
  \node[prop,draw=qsafe!80!black,fill=qsafe!9] (p1) at (-4.45,-2.2)
    {\textbf{\ref{item:P1} Safety}\\ \Cref{lem:safety}};
  \node[prop,draw=qplay!85!black,fill=qplay!7] (p2) at (0,-2.2)
    {\textbf{\ref{item:P2} Stage gap}\\ \Cref{lem:candidate_gap,lem:scaling_generic,lem:scaling_loss,lem:stage_gap}};
  \node[prop,draw=qcert!85!black,fill=qcert!8] (p3) at (4.45,-2.2)
    {\textbf{\ref{item:P3} Stage length}\\ stage rule~\eqref{eq:ns_def}};
  \node[wide,draw=black!75,fill=black!4,line width=1.1pt] (t) at (0,-4.4)
    {\textbf{\Cref{thm:main}}: sum over the stages, using the stage count (\Cref{lem:stage_count}) and the warm-up bound (\Cref{lem:warmup})};
  \draw[arr] (g.south -| p1.north) -- (p1.north);
  \draw[arr] (g.south -| p2.north) -- (p2.north);
  \draw[arr] (p1.south) -- (t.north -| p1.south);
  \draw[arr] (p2.south) -- (t.north -| p2.south);
  \draw[arr] (p3.south) -- (t.north -| p3.south);
\end{tikzpicture}
\caption{Structure of the proof of \Cref{thm:main}. Everything below the top box holds deterministically on $\mathcal G$; property~\ref{item:P3} holds at every stage.}
\label{fig:qrs-proof-structure}
\end{figure}
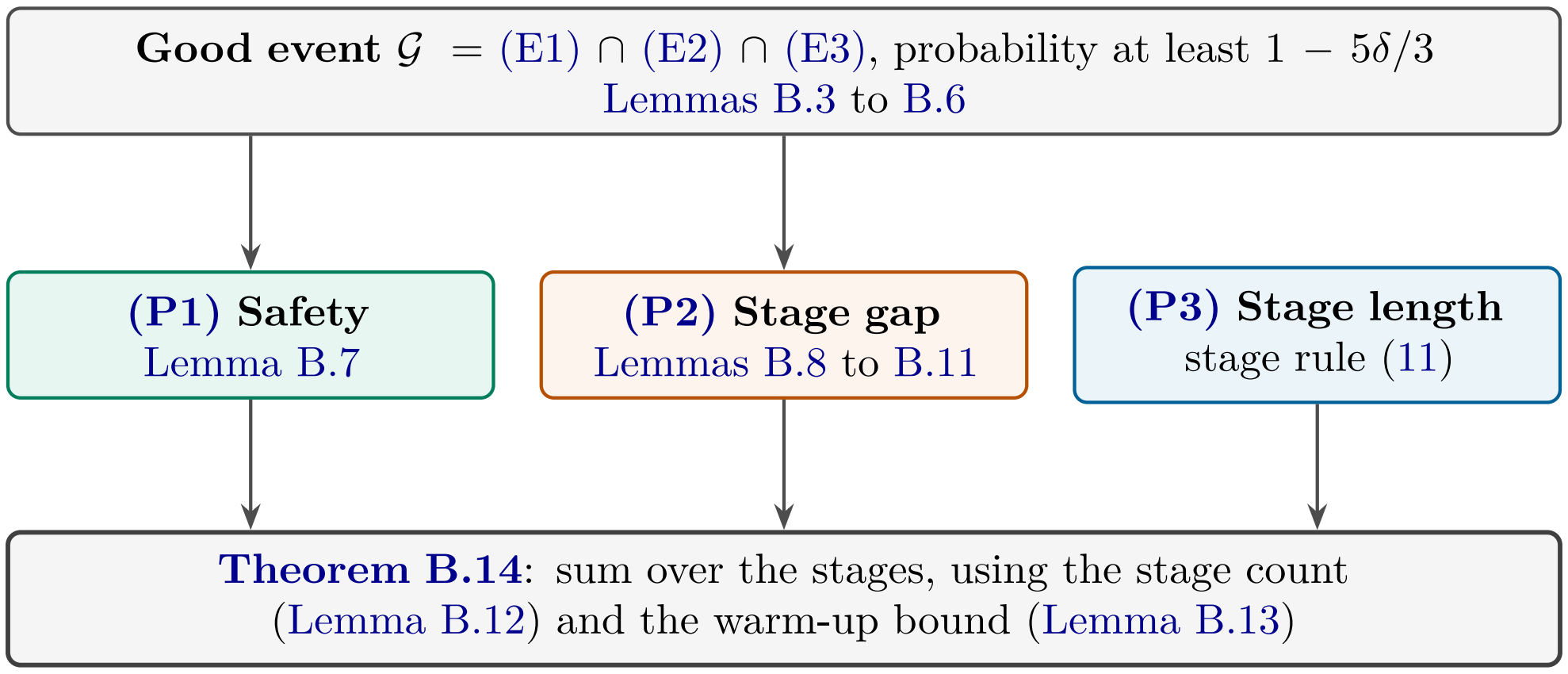

The proof of \Cref{thm:main} has three parts (\Cref{fig:qrs-proof-structure}). A good event $\mathcal G$ collects all the randomness. On $\mathcal G$, every sampling stage has the properties~\ref{item:P1}--\ref{item:P3} of \Cref{sec:qrs-overview}. Summing the stage regrets as in~\eqref{eq:qrs-regret-accounting} then gives the bound.

\paragraph{The good event.} The event $\mathcal G$ is the intersection of three events:
\begin{enumerate}[label=(E\arabic*),leftmargin=3.2em,itemsep=1pt,topsep=3pt]
\item\label{item:E1} every completed stage returns $\widehat\mu_s$ with $\|\widehat\mu_s-\mu(a_s)\|_\infty\le 1/n_s$; this is the event $\mathrm{QMEgood}(\delta)$ of \Cref{lem:Con_explicit}, and it implies the confidence event $\mathrm{Con}(\delta)$, on which~\eqref{eq:qrs-conf} holds at every stage (probability at least $1-\delta$);
\item\label{item:E2} every Gaussian vector drawn during the run has norm at most $B$ (\Cref{lem:gauss_norm}; probability at least $1-\delta/2$);
\item\label{item:E3} at every sampling stage at which~\eqref{eq:qrs-conf} holds, at least one of the $K_s$ draws is optimistic in the sense of Step~3 (\Cref{lem:global_opt}; probability at least $1-\delta/6$).
\end{enumerate}
Hence $\Pr[\mathcal G]\ge 1-5\delta/3$. On $\mathcal G$, \ref{item:E1} gives~\eqref{eq:qrs-conf} at every stage, so by~\ref{item:E3} every sampling stage has an optimistic draw.

\paragraph{The stage gap.} Properties~\ref{item:P1} and~\ref{item:P3} are direct: the conservative test certifies feasibility (\Cref{lem:safety}), and the stage length follows from the rule~\eqref{eq:ns_def}. Property~\ref{item:P2} needs one argument. The candidate $b_s$ has a small gap, $\Delta(b_s)\le\frac32M_s(b_s)$ (\Cref{lem:candidate_gap}), but it may be infeasible, so the algorithm plays $a_s=(1-\rho_s)u+\rho_sb_s$. The scaling lemma (\Cref{lem:scaling_generic,lem:scaling_loss}) shows that this costs only a constant factor:
\[
 (1-\rho_s)\,\Gamma\le 9\,M_s(a_s),\qquad \rho_s\,M_s(b_s)\le 4\,M_s(a_s).
\]
The warm-up test enters in the second bound. It ensures $3M_s(u)\le\Gamma$, which lets the proof move the uncertainty of $b_s$, an action that is not played, onto $a_s$. Since $\Delta$ is affine, the two bounds give~\ref{item:P2} with $C_\Delta=6+9\Delta(u)/\Gamma$ (\Cref{lem:stage_gap}).

\paragraph{Summing.} By~\ref{item:P2} and~\ref{item:P3}, a sampling stage adds $\widetilde O(\sqrt{m+1}\,C_\Delta B\beta_s)$ to the regret. There are $S=\widetilde O(d\log T)$ stages (\Cref{lem:stage_count}), and the warm-up stages use $T_{\mathrm{warm}}=\widetilde O(\sqrt{m+1}\,B\beta_S/\Gamma)$ queries, each with regret $\Delta(u)$ (\Cref{lem:warmup}). Since $\sum_{s\le S}\beta_s=O(S^{3/2})$,
\[
 \Reg(T)=\widetilde O\left(\sqrt{m+1}\,C_\Delta\,B\,S^{3/2}\right)+T_{\mathrm{warm}}\,\Delta(u),
\]
which is the first bound of \Cref{thm:main}. The warm-up term is of lower order, because $\Delta(u)/\Gamma\le C_\Delta$ and $\beta_S=O(\sqrt S)$.

\subsection{Detailed proofs}\label{sec:proofs}

Throughout this subsection, $\norm{\theta^\star}_2\le 1$ and $\norm{\phi_i^\star}_2\le 1$ for all $i\in[m]$, as in \Cref{thm:main_preview}. The value of $B$ in~\eqref{eq:B_def_main} satisfies $B\ge 1+\sqrt{2\log 2}>2$, since $d\ge 1$, $N_{\mathrm{Gauss}}\ge 1$ and $\delta<1$.

\subsubsection{Confidence event for staged QME}

\begin{lemma}[Confidence bounds from bounded stage errors]\label{lem:Con_explicit}
Let $\mathrm{QMEgood}(\delta)$ be the event that every completed stage returns $\widehat\mu_s$ with $\norm{\widehat\mu_s-\mu(a_s)}_\infty\le 1/n_s$. Then $\Pr[\mathrm{QMEgood}(\delta)]\ge 1-\delta$. On $\mathrm{QMEgood}(\delta)$, every stage $s$ that the run begins satisfies
\begin{equation}\label{eq:con_norm}
 \|\widehat\theta_s-\theta^\star\|_{V_s}\le\beta_s
 \qquad\text{and}\qquad
 \|\widehat\phi_{s,i}-\phi_i^\star\|_{V_s}\le\beta_s\quad\text{for all }i\in[m],
\end{equation}
and therefore, for all $a\in\RR^d$ and all $i\in[m]$,
\begin{align}
\big|\ip{\widehat\theta_s-\theta^\star}{a}\big|
&\le \beta_s\,\norm{a}_{V_s^{-1}},\label{eq:con_reward}\\
\big|\ip{\widehat\phi_{s,i}-\phi_i^\star}{a}\big|
&\le \beta_s\,\norm{a}_{V_s^{-1}}.\label{eq:con_constraints}
\end{align}
We write $\mathrm{Con}(\delta)$ for the event that~\eqref{eq:con_reward} and~\eqref{eq:con_constraints} hold for every stage $s$ that the run begins, all $a$ and all $i$.
\end{lemma}

\begin{proof}
\emph{Probability.} A completed stage $s$ fails with conditional probability at most $\delta_s$ by \Cref{lem:stage_estimator}, and a stage truncated by the horizon records no estimate. A union bound over the stages gives $\Pr[\mathrm{QMEgood}(\delta)^c]\le\sum_{s\ge 1}\delta_s=\delta$.

\emph{Confidence bounds.} Fix one of the $m+1$ coordinates of the feedback, with true parameter $w^\star$, stage estimates $\widehat y_j$, and ridge estimate $\widehat w_s=V_s^{-1}\sum_{j<s}n_j^2a_j\widehat y_j$. Let $X$ be the matrix with rows $n_ja_j^\top$ for $j<s$, and let $\xi$ be the vector with entries $\xi_j=n_j(\widehat y_j-\ip{w^\star}{a_j})$. On $\mathrm{QMEgood}(\delta)$, $|\xi_j|\le 1$ for every $j<s$. Then
\begin{align*}
 \|\widehat w_s-w^\star\|_{V_s}
 &=\bigl\|-\lambda V_s^{-1}w^\star+V_s^{-1}X^\top\xi\bigr\|_{V_s}\\
 &\le\lambda\,\|w^\star\|_{V_s^{-1}}+\bigl(\xi^\top XV_s^{-1}X^\top\xi\bigr)^{1/2}\\
 &\le\sqrt\lambda\,\|w^\star\|_2+\|\xi\|_2
 \le\sqrt\lambda+\sqrt{s-1}=\beta_s,
\end{align*}
where the equality uses the definition of $\xi$ and $X^\top X=V_s-\lambda I$, the first inequality is the triangle inequality, the second uses $V_s\succeq\lambda I$ and $XV_s^{-1}X^\top\preceq I$ (every eigenvalue of $XV_s^{-1}X^\top$ has the form $t^2/(t^2+\lambda)$ for a singular value $t$ of $X$), and the last uses $\|w^\star\|_2\le 1$ and $|\xi_j|\le 1$. For $s=1$ the matrix $X$ is empty and the same bound holds. This is~\eqref{eq:con_norm}. By the Cauchy--Schwarz inequality, for every $a\in\RR^d$,
\[
 \big|\ip{\widehat w_s-w^\star}{a}\big|\le\|\widehat w_s-w^\star\|_{V_s}\,\|a\|_{V_s^{-1}}\le\beta_s\,\|a\|_{V_s^{-1}}. \qedhere
\]
\end{proof}

\subsubsection{Bounding perturbations}

\begin{lemma}[Gaussian norm bound]\label{lem:gauss_norm}
Let $N\ge 1$ and $\delta'\in(0,1)$. If a run draws at most $N$ standard Gaussian vectors in total, then with probability at least $1-\delta'/2$ every draw satisfies $\|\zeta\|_2\le\sqrt d+\sqrt{2\log(2N/\delta')}$. In particular, QRS-COLTS draws at most $N_{\mathrm{Gauss}}$ vectors by~\eqref{eq:NGauss_def_main} and~\eqref{eq:qrs-stage-budget}, so with probability at least $1-\delta/2$ every draw satisfies $\|\zeta\|_2\le B(\delta)=\sqrt d+\sqrt{2\log(2N_{\mathrm{Gauss}}/\delta)}$, the value of $B$ in~\eqref{eq:B_def_main}.
\end{lemma}

\begin{proof}
Let $\zeta\sim\mathcal N(0,I_d)$. Jensen's inequality gives $\EE[\|\zeta\|_2]\le\sqrt d$, and the Gaussian concentration inequality~\cite[Theorem~5.6]{boucheron2013concentration} for the $1$-Lipschitz function $x\mapsto\|x\|_2$ gives
\[
 \Pr\left[\|\zeta\|_2\ge\sqrt d+t\right]\le e^{-t^2/2}\qquad\forall t\ge 0.
\]
An adaptive run reads a prefix of a sequence of $N$ independent standard Gaussian vectors. A union bound over this sequence with $t=\sqrt{2\log(2N/\delta')}$ bounds the failure probability by $Ne^{-t^2/2}=\delta'/2$.
\end{proof}

As in Step~4 of \Cref{sec:qrs-overview}, the stage-wise uncertainty proxy is
\begin{equation}\label{eq:M_def}
M_s(a) := B\,\beta_s\,\norm{a}_{V_s^{-1}}.
\end{equation}

\subsubsection{Local optimism and adaptive resampling}
We next show that on any stage history satisfying the confidence bounds, a single coupled perturbation draw is optimistic at $a^\star$ with constant probability at least $\pi_{\mathrm{opt}}$.

\begin{lemma}[Local optimism probability]\label{lem:local_opt}
Fix a stage $s$ and condition on a history $\cF_{s-1}$ satisfying the stage-$s$ confidence bounds~\eqref{eq:con_reward}--\eqref{eq:con_constraints}. Let $a^\star$ be an optimal feasible action. For a single coupled draw $(\widetilde\theta_s,\widetilde\Phi_s)$ of~\eqref{eq:qrs-perturbation},
$$
\Pr\Big[
\ip{\widetilde\theta_s}{a^\star} \ge \ip{\theta^\star}{a^\star}
\;\text{ and }\;
\widetilde\Phi_s a^\star \le \alpha
\ \Big|\ \cF_{s-1}
\Big]
\ \ge\
\pi_{\rm opt},
$$
where $\pi_{\rm opt}:=\Pr[Z\ge 1]\approx 0.1587$ for $Z\sim\mathcal{N}(0,1)$.
\end{lemma}

\begin{proof}
Fix a history $\cF_{s-1}$ on which~\eqref{eq:con_reward} and~\eqref{eq:con_constraints} hold at stage $s$. If $a^\star=0$, the perturbation vanishes at $a^\star$ and both conditions hold with probability $1$. Otherwise, set $w:=\|a^\star\|_{V_s^{-1}}>0$ and define the normalized projection
\begin{align*}
 Z=\frac{\langle\zeta,V_s^{-1/2}a^\star\rangle}{w}.
\end{align*}
Conditioned on $\cF_{s-1}$, $V_s^{-1/2}a^\star/w$ is a deterministic unit vector, so $Z\sim\mathcal N(0,1)$. We first lower-bound the perturbed reward at $a^\star$:
\begin{align*}
 \langle\widetilde\theta_s,a^\star\rangle
 = \langle\widehat\theta_s,a^\star\rangle+\beta_s w Z
 \ge \langle\theta^\star,a^\star\rangle+\beta_s w(Z-1),
\end{align*}
where the equality follows from the definition of $\widetilde\theta_s$ in~\eqref{eq:qrs-perturbation} and the definition of $Z$, and the inequality follows from the reward confidence bound~\eqref{eq:con_reward}. Similarly, for each constraint row $i\in[m]$, we upper-bound the perturbed constraint value at $a^\star$:
\begin{align*}
 \langle\widetilde\phi_{s,i},a^\star\rangle
 = \langle\widehat\phi_{s,i},a^\star\rangle-\beta_s wZ
 \le \langle\phi_i^\star,a^\star\rangle+\beta_s w(1-Z)
 \le \alpha_i+\beta_s w(1-Z),
\end{align*}
where the equality uses the row $\widetilde\phi_{s,i}=\widehat\phi_{s,i}-\beta_sV_s^{-1/2}\zeta$ of $\widetilde\Phi_s$ in~\eqref{eq:qrs-perturbation}, the first inequality follows from the constraint confidence bound~\eqref{eq:con_constraints}, and the second inequality uses the true feasibility $\langle\phi_i^\star,a^\star\rangle\le\alpha_i$.

Consequently, the single event $Z\ge 1$ implies both $\langle\widetilde\theta_s,a^\star\rangle\ge\langle\theta^\star,a^\star\rangle$ and $\widetilde\Phi_s a^\star\le\alpha$, and its conditional probability is $\pi_{\mathrm{opt}}$.
\end{proof}

Recall the resampling budget $K_s$ from~\eqref{eq:Ks_def_main} and write $\nu_s:=\delta/(\pi^2s^2)$, so that $K_s=\lceil\pi_{\rm opt}^{-1}\log(1/\nu_s)\rceil$.

\begin{lemma}[Global optimism via adaptive resampling]\label{lem:global_opt}
Let $K_s$ be as in~\eqref{eq:Ks_def_main}. With probability at least $1-\delta/6$, every sampling stage $s$ at which~\eqref{eq:con_reward}--\eqref{eq:con_constraints} hold has at least one optimistic draw, that is, a draw $k\in[K_s]$ with $\widetilde\Phi_s^{(k)}a^\star\le\alpha$ and $\langle\widetilde\theta_s^{(k)}, a^\star\rangle\ge \ip{\theta^\star}{a^\star}$. At such a stage, the pair $(\widetilde\theta_s,b_s)$ selected in Step~4 satisfies the \emph{global optimism} inequality
\begin{equation}\label{eq:global_opt}
\ip{\widetilde\theta_s}{b_s}\ge \ip{\theta^\star}{a^\star}.
\end{equation}
\end{lemma}

\begin{proof}
Fix any history $\cF_{s-1}$ at which stage $s$ is a sampling stage and satisfies~\eqref{eq:con_reward}--\eqref{eq:con_constraints}. We first bound the conditional probability that none of the $K_s$ draws succeeds in the sense of \Cref{lem:local_opt}:
\begin{align*}
 \Pr[\text{no successful draw at stage }s\mid\cF_{s-1}]
 &\le (1-\pi_{\mathrm{opt}})^{K_s}
 \le e^{-\pi_{\mathrm{opt}}K_s}\le\nu_s,
\end{align*}
where the first inequality uses conditional independence of the $K_s$ draws and \Cref{lem:local_opt}, the second inequality uses $1-x\le e^{-x}$, and the last inequality substitutes $K_s\ge\pi_{\mathrm{opt}}^{-1}\log(1/\nu_s)$.

Averaging over these histories and summing over stages, the probability that some sampling stage has valid confidence bounds yet has no successful draw is at most $\sum_{s\ge 1}\nu_s=\delta/6$.

Whenever draw $k\in[K_s]$ succeeds, we lower-bound the perturbed objective at the selected candidate $(\widetilde\theta_s,b_s)$:
\begin{align*}
 \langle\widetilde\theta_s,b_s\rangle
 \ge \langle\widetilde\theta_s^{(k)},b_s^{(k)}\rangle
 \ge \langle\widetilde\theta_s^{(k)},a^\star\rangle
 \ge \langle\theta^\star,a^\star\rangle,
\end{align*}
where the first step follows from the rule selecting $b_s$ among the solutions of all feasible subproblems~\eqref{eq:qrs-subproblems}, the second step uses the optimality of $b_s^{(k)}$ in the $k$-th subproblem (since $a^\star$ is feasible for draw $k$), and the last step follows from the optimism of draw $k$. This proves~\eqref{eq:global_opt}.
\end{proof}

\subsubsection{Safety, scaling, and the stage gap}
We show that the conservative test certifies true feasibility and bound the gap of the played action. Recall that $\Delta(a)=\langle\theta^\star,a^\star-a\rangle$ is the signed affine gap from \Cref{sec:prelim}; the candidate $b_s$ may lie outside $\cS$ and have $\Delta(b_s)<0$.

\begin{lemma}[Round-wise mean safety on the confidence event]\label{lem:safety}
On $\mathrm{Con}(\delta)$, every $a\in\cA$ with $\widehat\Phi_s a + \beta_s\norm{a}_{V_s^{-1}}\one_m \le \alpha$ satisfies $\Phi^\star a \le \alpha$. In particular, QRS-COLTS plays only feasible actions on $\mathrm{Con}(\delta)$.
\end{lemma}

\begin{proof}
By~\eqref{eq:con_constraints}, $\ip{\phi_i^\star}{a}\le\langle \widehat\phi_{s,i}, a\rangle+\beta_s\|a\|_{V_s^{-1}}\le\alpha_i$ for every $i\in[m]$. QRS-COLTS plays either an action that passes~\eqref{eq:ucb_safe_constraint} or $a_{\mathrm{safe}}$, which is feasible.
\end{proof}

\begin{lemma}[Candidate gap under global optimism]\label{lem:candidate_gap}
On $\mathrm{Con}(\delta)$ and the Gaussian norm event of \Cref{lem:gauss_norm}, if stage $s$ satisfies the global optimism inequality~\eqref{eq:global_opt}, then
\begin{equation}\label{eq:candidate_gap_bound}
\Delta(b_s) \le \Bigl(1+\frac{1}{B}\Bigr) M_s(b_s).
\end{equation}
\end{lemma}

\begin{proof}
Let $\widetilde\theta_s=\widehat\theta_s+\beta_sV_s^{-1/2}\zeta_s$ be the perturbed reward vector paired with $b_s$. Then
\begin{align*}
 \Delta(b_s)
 &\le\ip{\widetilde\theta_s-\theta^\star}{b_s}
 =\beta_s\ip{\zeta_s}{V_s^{-1/2}b_s}+\ip{\widehat\theta_s-\theta^\star}{b_s}\\
 &\le B\,\beta_s\|b_s\|_{V_s^{-1}}+\beta_s\|b_s\|_{V_s^{-1}}
 =\Bigl(1+\frac1B\Bigr)M_s(b_s),
\end{align*}
where the first inequality is~\eqref{eq:global_opt}, the second uses the Cauchy--Schwarz inequality with $\|\zeta_s\|_2\le B$ and~\eqref{eq:con_reward}, and the last equality is~\eqref{eq:M_def}.
\end{proof}

The scaling step is the same in QRS-COLTS and in the variant of \Cref{sec:global-optimism-upper}, so we state it for general convex test functions. Its notation follows the applications, where $u=a_{\mathrm{safe}}$, $\Gamma$ is the baseline slack, $M$ is a multiple of the confidence width $h_s$ and $\Delta$ is the gap.

\begin{lemma}[Scaling toward the baseline]\label{lem:scaling_generic}
Let $u,b\in\RR^d$, let $M$ be a norm on $\RR^d$, and let $F_1,\dots,F_m:\RR^d\to\RR$ be convex. Suppose that, for constants $\Gamma,\kappa>0$, $c\ge 0$ and $\tau\ge 0$ with $c\tau<\kappa$,
\begin{equation}\label{eq:scaling_hyp}
 F_i(u)\le\alpha_i-\kappa\Gamma\quad\text{and}\quad
 F_i(b)\le\alpha_i+c\,M(b)\quad\text{for all }i\in[m],\qquad
 M(u)\le\tau\,\Gamma .
\end{equation}
Let $\rho$ be the largest $\rho'\in[0,1]$ such that $F_i((1-\rho')u+\rho'b)\le\alpha_i$ for all $i\in[m]$, and let $a=(1-\rho)u+\rho b$. Then
\begin{equation}\label{eq:scaling_generic}
 (1-\rho)\,\Gamma\le\frac{c}{\kappa-c\tau}\,M(a)
 \qquad\text{and}\qquad
 \rho\,M(b)\le\frac{\kappa}{\kappa-c\tau}\,M(a).
\end{equation}
If moreover $\Delta$ is affine with $\Delta(u)\ge 0$ and $\Delta(b)\le c\,M(b)$, then
\begin{equation}\label{eq:scaling_generic_gap}
 \Delta(a)\le\frac{c}{\kappa-c\tau}\Bigl(\kappa+\frac{\Delta(u)}{\Gamma}\Bigr)M(a).
\end{equation}
\end{lemma}

\begin{proof}
The largest $\rho$ exists because the set of admissible $\rho'$ is closed and contains $0$.

\emph{Lever rule.} For $\rho'\in[0,1]$ and every $i\in[m]$,
\begin{align*}
 F_i\bigl((1-\rho')u+\rho'b\bigr)
 \le(1-\rho')F_i(u)+\rho'F_i(b)
 \le\alpha_i-(1-\rho')\,\kappa\Gamma+\rho'c\,M(b),
\end{align*}
where the first inequality is convexity of $F_i$ and the second is~\eqref{eq:scaling_hyp}. The right-hand side is at most $\alpha_i$ for $\rho'=\bar\rho:=\kappa\Gamma/(\kappa\Gamma+cM(b))$, so $\rho\ge\bar\rho$ and
\begin{equation}\label{eq:scaling_lever}
 (1-\rho)\,\Gamma\le(1-\bar\rho)\,\Gamma=\frac{c}{\kappa}\,\bar\rho\,M(b)\le\frac{c}{\kappa}\,\rho\,M(b).
\end{equation}

\emph{Moving the uncertainty to $a$.} Since $\rho b=a-(1-\rho)u$,
\begin{align*}
 \rho\,M(b)
 \le M(a)+(1-\rho)\,M(u)
 \le M(a)+\tau(1-\rho)\,\Gamma
 \le M(a)+\frac{c\tau}{\kappa}\,\rho\,M(b),
\end{align*}
where the first inequality is the triangle inequality, the second uses $M(u)\le\tau\Gamma$, and the third is~\eqref{eq:scaling_lever}. Since $c\tau<\kappa$, rearranging gives the second bound in~\eqref{eq:scaling_generic}, and~\eqref{eq:scaling_lever} then gives the first.

\emph{Gap.} Since $\Delta$ is affine,
\begin{align*}
 \Delta(a)=(1-\rho)\,\Delta(u)+\rho\,\Delta(b)
 \le\frac{\Delta(u)}{\Gamma}\,(1-\rho)\,\Gamma+c\,\rho\,M(b),
\end{align*}
where the inequality uses $\Delta(b)\le cM(b)$. Substituting~\eqref{eq:scaling_generic} and using $\Delta(u)\ge 0$ gives~\eqref{eq:scaling_generic_gap}.
\end{proof}

\begin{lemma}[Scaling loss inequalities]\label{lem:scaling_loss}
On $\mathrm{Con}(\delta)$ and the Gaussian norm event of \Cref{lem:gauss_norm}, consider a stage with $3M_s(u)\le\Gamma$ whose candidate satisfies $\widetilde\Phi_sb_s\le\alpha$, and let $a_s=(1-\rho_s)u+\rho_sb_s$ be the action of~\eqref{eq:qrs-scaling}. Then
\begin{align}
(1-\rho_s)\,\Gamma&\le\frac{3(B+1)}{2B-3}\,M_s(a_s)\le 9\,M_s(a_s),\label{eq:scaling_3}\\
\rho_s\,M_s(b_s)&\le\frac{3B-2}{2B-3}\,M_s(a_s)\le 4\,M_s(a_s).\label{eq:scaling_2}
\end{align}
Both middle constants tend to $\frac32$ as $B\to\infty$.
\end{lemma}

\begin{proof}
We apply \Cref{lem:scaling_generic} with $b=b_s$, $M=M_s$, $F_i(x)=\ip{\widehat\phi_{s,i}}{x}+h_s(x)$, $\kappa=1-\frac{2}{3B}$, $c=1+\frac1B$ and $\tau=\frac13$. The condition $F_i(x)\le\alpha_i$ for all $i\in[m]$ is the conservative test~\eqref{eq:ucb_safe_constraint}, so the maximal step of the lemma is the $\rho_s$ of~\eqref{eq:qrs-scaling}, and $M_s(u)\le\tau\Gamma$ by assumption. For every $i\in[m]$,
\begin{align}
 F_i(u)&\le\ip{\phi_i^\star}{u}+2h_s(u)\le\alpha_i-\Gamma+\frac{2}{3B}\,\Gamma=\alpha_i-\kappa\Gamma,\label{eq:baseline_margin_kappa}\\
 F_i(b_s)&=\ip{\widetilde\phi_{s,i}}{b_s}+\beta_s\ip{\zeta_s}{V_s^{-1/2}b_s}+h_s(b_s)\notag\\
 &\le\alpha_i+M_s(b_s)+h_s(b_s)=\alpha_i+c\,M_s(b_s),\notag
\end{align}
where the first line uses~\eqref{eq:con_constraints}, the definition of $\Gamma$, and $2h_s(u)=\frac2BM_s(u)\le\frac{2}{3B}\Gamma$. The second line uses the rows $\widehat\phi_{s,i}=\widetilde\phi_{s,i}+\beta_sV_s^{-1/2}\zeta_s$ of~\eqref{eq:qrs-perturbation}, and the third uses the sampled feasibility $\widetilde\Phi_sb_s\le\alpha$ and the Cauchy--Schwarz inequality with $\|\zeta_s\|_2\le B$. Thus~\eqref{eq:scaling_hyp} holds. Since $\kappa-c\tau=\frac{2B-3}{3B}$, \eqref{eq:scaling_generic} gives the middle expressions in~\eqref{eq:scaling_3} and~\eqref{eq:scaling_2}. Both decrease in $B$ and equal $9$ and $4$ at $B=2$.
\end{proof}

\begin{lemma}[Stage gap bound]\label{lem:stage_gap}
Under the assumptions of \Cref{lem:scaling_loss}, if stage $s$ also satisfies~\eqref{eq:global_opt}, then
\begin{equation}\label{eq:stage_gap_bound}
\Delta(a_s)\le C_\Delta(B)\,M_s(a_s)\le\Bigl(6+9\,\frac{\Delta(u)}{\Gamma}\Bigr)M_s(a_s),
\end{equation}
where $C_\Delta(B):=\frac{(B+1)(3B-2)}{B(2B-3)}+\frac{3(B+1)}{2B-3}\cdot\frac{\Delta(u)}{\Gamma}$, which tends to $\frac32(1+\Delta(u)/\Gamma)$ as $B\to\infty$.
\end{lemma}

\begin{proof}
By \Cref{lem:candidate_gap}, $\Delta(b_s)\le cM_s(b_s)$ with $c=1+1/B$, and $\Delta(u)\ge 0$ because $u$ is feasible. With the constants of the previous proof, the second part of \Cref{lem:scaling_generic} gives
\[
 \Delta(a_s)\le\frac{3(B+1)}{2B-3}\Bigl(1-\frac{2}{3B}+\frac{\Delta(u)}{\Gamma}\Bigr)M_s(a_s)=C_\Delta(B)\,M_s(a_s).
\]
Both coefficients of $C_\Delta(B)$ decrease in $B$ and equal $6$ and $9$ at $B=2$.
\end{proof}

\subsubsection{Stage count}
Let $d_{\cA}:=\dim\mathrm{span}(\cA)\le d$. If $d_{\cA}=0$, regret vanishes identically; assume $d_{\cA}>0$ below.

\begin{lemma}[Stage count via log-determinant growth]\label{lem:stage_count}
Under the stage rule $n_j=\lceil c_{\mathrm{len}}/\norm{a_j}_{V_j^{-1}}\rceil$ and the update $V_{j+1}=V_j+n_j^2 a_j a_j^\top$, if stages $1,\dots,s$ are completed, then
$$
s
\ \le\
\frac{d_{\cA}}{\log(1+c_{\mathrm{len}}^2)}\;\log\!\Big(1+\frac{1}{d_{\cA}\lambda}\sum_{j=1}^s n_j^2\Big)
\ \le\
\frac{d_{\cA}}{\log(1+c_{\mathrm{len}}^2)}\;\log\!\Big(1+\frac{1}{d_{\cA}\lambda}\Big(\sum_{j=1}^s n_j\Big)^2\Big).
$$
\end{lemma}

\begin{proof}
We first lower-bound the determinant ratio of each completed stage $j\in[s]$:
\begin{align*}
\frac{\det (V_{j+1})}{\det (V_j)}
 = 1+n_j^2\norm{a_j}_{V_j^{-1}}^2
 \ge 1+c_{\mathrm{len}}^2,
\end{align*}
where the equality follows from the matrix determinant lemma and the inequality uses $n_j\|a_j\|_{V_j^{-1}}\ge c_{\mathrm{len}}$ from~\eqref{eq:ns_def}. Telescoping this product over $j=1,\dots,s$ gives $\det (V_{s+1})/\det (V_1)\ge(1+c_{\mathrm{len}}^2)^s$.

Next, let $D=V_{s+1}-\lambda I = \sum_{j=1}^s n_j^2 a_j a_j^\top$, which vanishes on $\mathrm{span}(\cA)^\perp$, and let $\gamma_1,\ldots,\gamma_{d_{\cA}}\ge 0$ denote its eigenvalues on $\mathrm{span}(\cA)$. We upper-bound $\det (V_{s+1})/\det (V_1)$:
\begin{align*}
\frac{\det (V_{s+1})}{\det (V_1)}
 = \prod_{i=1}^{d_{\cA}}(1+\gamma_i/\lambda)
 \le \left(1+\frac{\mathrm{Tr}[D]}{d_{\cA}\lambda}\right)^{d_{\cA}}
 \le \left(1+\frac{1}{d_{\cA}\lambda}\sum_{j=1}^s n_j^2\right)^{d_{\cA}},
\end{align*}
where the equality uses $V_1=\lambda I$ and diagonalizes $D$ on $\mathrm{span}(\cA)$, the first inequality applies the arithmetic--geometric mean inequality with $\sum_{i=1}^{d_{\cA}}\gamma_i=\mathrm{Tr}[D]$, and the second inequality uses $\mathrm{Tr}[D]=\sum_{j=1}^s n_j^2\|a_j\|_2^2\le\sum_{j=1}^s n_j^2$ since $\|a_j\|_2\le 1$. Taking logarithms of the lower and upper bounds on $\det (V_{s+1})/\det (V_1)$ and dividing by $\log(1+c_{\mathrm{len}}^2)$ proves the first inequality in the lemma, and $\sum_{j=1}^s n_j^2\le(\sum_{j=1}^s n_j)^2$ proves the second.
\end{proof}

\subsubsection{Warm-up bound}
Put $u=a_{\mathrm{safe}}$, $B=B(\delta)$, $h_s(u)=\beta_s\|u\|_{V_s^{-1}}$, and $\Gamma=\min_{i\in[m]}(\alpha_i-\langle\phi_i^\star,u\rangle)>0$. On $\mathrm{Con}(\delta)$, each estimated constraint $\langle\widehat\phi_{s,i},u\rangle$ lies within $h_s(u)$ of $\langle\phi_i^\star,u\rangle$; taking the minimum over $i\in[m]$ and subtracting $h_s(u)$ yields the two-sided bound
\begin{equation}\label{eq:baseline-slack-sandwich}
 \Gamma-2h_s(u)\le\underline{\Gamma}_s\le\Gamma.
\end{equation}
Consequently, every sampling stage ($\underline{\Gamma}_s\ge 3M_s(u)$) satisfies $3M_s(u)\le\underline{\Gamma}_s\le\Gamma$, whereas every warm-up stage ($\underline{\Gamma}_s<3M_s(u)$) satisfies
\begin{equation}\label{eq:warm-true-slack}
 \Gamma\le \underline{\Gamma}_s+2h_s(u)<3M_s(u)+2h_s(u)=(3+2/B)M_s(u).
\end{equation}
Notice that~\eqref{eq:baseline-slack-sandwich} and~\eqref{eq:warm-true-slack} hold at every stage $s$ on $\mathrm{Con}(\delta)$, including if the algorithm re-enters warm-up after a sampling stage.

\begin{lemma}[Warm-up query cost]\label{lem:warmup}
Let $B'\ge 2$ be a number fixed before the run, and suppose that stage $s$ is a warm-up stage exactly when $\underline{\Gamma}_s<3B'h_s(a_{\mathrm{safe}})$, with the stage estimator, stage rule~\eqref{eq:ns_def}, design update and horizon rule of \Cref{alg:qrs-colts}; QRS-COLTS is the case $B'=B$. Let $T_{\mathrm{warm}}$ be the total number of oracle queries spent in warm-up stages. Then on $\mathrm{Con}(\delta)$,
$$
T_{\mathrm{warm}}
\le\frac{4(c_{\mathrm{len}}+1)}{1-\kappa_{\mathrm w}}\,\sqrt{m+1}\,\bar L\,\frac{B'\beta_S}{\Gamma}
= \widetilde O\!\left(\sqrt{m+1}\,\frac{B'}{\Gamma}\,\beta_S\right),
$$
where $\kappa_{\mathrm w}:=(1+c_{\mathrm{len}}^2)^{-1/2}<1$, and $\widetilde O$ hides the constant depending on $c_{\mathrm{len}}$ and the factor $\bar L$ defined after \Cref{lem:stage_estimator}.
\end{lemma}

\begin{proof}
Write $u=a_{\mathrm{safe}}$ and $z_s=u^\top V_s^{-1}u$, so that $h_s(u)=\beta_s\sqrt{z_s}$. By~\eqref{eq:baseline-slack-sandwich}, every warm-up stage $s$ satisfies
\begin{equation}\label{eq:warm-general}
 \Gamma\le\underline{\Gamma}_s+2h_s(u)<(3B'+2)\,h_s(u)\le 4B'\beta_s\sqrt{z_s},
\end{equation}
where the last step uses $B'\ge 2$; for $B'=B$, the first two steps are~\eqref{eq:warm-true-slack}. Let $s_1<\cdots<s_\ell\le S$ index the warm-up stages begun by the horizon. If $\ell=0$, the bound holds trivially. Otherwise~\eqref{eq:warm-general} at stage $s_1$ implies $u\ne 0$, and $V_s\succeq I$ with $\|u\|_2\le 1$ gives $0<z_s\le 1$.

For any completed warm-up stage $s$, we bound the contraction of $z_s$ after the rank-one update $V_{s+1}=V_s+n_s^2 uu^\top$:
\begin{align*}
z_{s+1}
 = z_s-\frac{n_s^2z_s^2}{1+n_s^2z_s}
 = \frac{z_s}{1+n_s^2z_s}
 \le \frac{z_s}{1+c_{\mathrm{len}}^2},
\end{align*}
where the first equality applies the Sherman--Morrison formula, the second simplifies the fraction, and the inequality uses $n_s\sqrt{z_s}\ge c_{\mathrm{len}}$ from~\eqref{eq:ns_def}.

Updates at the other stages add positive semidefinite matrices to $V_s$, so they cannot increase $z_s$. Every warm-up stage $s_j$ with $j<\ell$ completed its budget before $s_\ell$ began. Iterating the contraction therefore gives, for all $1\le j\le\ell$,
\begin{align*}
z_{s_\ell}\le \kappa_{\mathrm w}^{2(\ell-j)}z_{s_j},
\qquad\text{and hence}\qquad
\frac{1}{\sqrt{z_{s_j}}}\le\frac{\kappa_{\mathrm w}^{\ell-j}}{\sqrt{z_{s_\ell}}}.
\end{align*}
We next sum the target inverse accuracies $n_{s_j}$ over all warm-up stages:
\begin{align*}
\sum_{j=1}^\ell n_{s_j}
 \le (c_{\mathrm{len}}+1)\sum_{j=1}^\ell\frac{1}{\sqrt{z_{s_j}}}
 \le \frac{c_{\mathrm{len}}+1}{(1-\kappa_{\mathrm w})\sqrt{z_{s_\ell}}}
 < \frac{4(c_{\mathrm{len}}+1)}{1-\kappa_{\mathrm w}}\cdot\frac{B'\beta_S}{\Gamma},
\end{align*}
where the first inequality uses $n_s\le c_{\mathrm{len}}/\sqrt{z_s}+1\le(c_{\mathrm{len}}+1)/\sqrt{z_s}$ (since $z_s\le 1$), the second inequality sums the geometric series in $\kappa_{\mathrm w}^{\ell-j}$, and the last inequality applies~\eqref{eq:warm-general} at stage $s_\ell$ together with $\beta_{s_\ell}\le\beta_S$.

Finally, warm-up stage $s_j$ executes at most $\min\{q_{s_j},T\}$ queries, so
\begin{align*}
 T_{\mathrm{warm}}\le\sum_{j=1}^{\ell}\min\{q_{s_j},T\}\le\sqrt{m+1}\,\bar L\sum_{j=1}^{\ell}n_{s_j},
\end{align*}
where a stage with $n_{s_j}\le T$ executes at most $q_{s_j}\le\sqrt{m+1}\,n_{s_j}L(n_{s_j},s_j)\le\sqrt{m+1}\,n_{s_j}\bar L$ queries by \Cref{lem:stage_estimator} (recall $s_j\le S\le T$), and a stage with $n_{s_j}>T$ executes at most $T<\sqrt{m+1}\,n_{s_j}\le\sqrt{m+1}\,n_{s_j}\bar L$ queries. The last two displays prove the bound.
\end{proof}

\subsubsection{Main regret theorem}
We now assemble the safety, scaling, stage-count, and warm-up bounds into our main theorem.

\begin{theorem}[High-probability safety and regret]\label{thm:main}
Fix $\delta\in(0,1)$ and run QRS-COLTS for a total of $T$ oracle queries. Assume:
(i)~$\norm{\theta^\star}_2\le 1$ and $\norm{\phi_i^\star}_2\le 1$ for all $i\in[m]$, and
(ii)~the learner knows a strictly safe baseline $a_{\mathrm{safe}}\in\cA$ with true baseline slack $\Gamma:=\min_{i\in[m]}(\alpha_i-\langle\phi_i^\star,a_{\mathrm{safe}}\rangle)>0$.
Then, with probability at least $1-3\delta$, the algorithm is safe at every query and satisfies
$$
\Reg(T)
\;=\;
\widetilde O\!\left(
\sqrt{m+1}\,B\,(c_{\mathrm{len}}+1)\,C_\Delta(B)\,S^{3/2}
\right)
\;+\;
T_{\mathrm{warm}}\,\Delta(a_{\mathrm{safe}}),
$$
where $\widetilde O$ hides an absolute constant and the factor $\bar L$ defined after \Cref{lem:stage_estimator}; $B=B(\delta)$ and $C_\Delta(B)$ are as in \Cref{lem:gauss_norm,lem:stage_gap}, $T_{\mathrm{warm}}$ is bounded by \Cref{lem:warmup}, and the number $S$ of stages begun by query $T$ satisfies, with $d_{\cA}=\dim\mathrm{span}(\cA)\le d$,
$$
S-1
\le
\frac{d_{\cA}}{\log(1+c_{\mathrm{len}}^2)}\;
\log\!\Bigl(1+\frac{T^2}{(m+1)\,d_{\cA}\,\lambda}\Bigr).
$$
Consequently, absorbing logarithmic factors in $d$, $m$, $T$ and $1/\delta$,
\begin{align*}
\Reg(T)&=\widetilde O\!\left(\sqrt{m+1}\,d^2\log^{3/2}(2T/\delta)
 \left(1+\frac{\Delta(a_{\mathrm{safe}})}{\Gamma}\right)\right).
\end{align*}
\end{theorem}

\begin{proof}
We establish the joint high-probability event, decompose regret across warm-up and sampling stages, and bound the total stage count.

\emph{Joint event and safety.} Let $\mathcal G$ be the good event of \Cref{app:qrs-colts-proof-overview}, the intersection of~\ref{item:E1}, \ref{item:E2} and~\ref{item:E3}. By \Cref{lem:Con_explicit,lem:gauss_norm,lem:global_opt},
\[
 \Pr[\mathcal G^c]\le\delta+\frac{\delta}{2}+\frac{\delta}{6}=\frac{5\delta}{3}\le 3\delta .
\]
On $\mathcal G$, $\mathrm{Con}(\delta)$ holds, every Gaussian draw has norm at most $B$, and, because the confidence bounds hold at every stage, the global optimism inequality~\eqref{eq:global_opt} holds at every sampling stage. \Cref{lem:safety} then ensures that every queried action is feasible, including the queries of a truncated final stage. Now fix a sampling stage $s$. By~\eqref{eq:baseline-slack-sandwich}, $3M_s(a_{\mathrm{safe}})\le\underline{\Gamma}_s\le\Gamma$. Global optimism (\Cref{lem:global_opt}) provides a draw whose subproblem has $a^\star$ as a feasible point, so at least one subproblem is feasible. By~\eqref{eq:baseline_margin_kappa}, $\rho=0$ passes~\eqref{eq:ucb_safe_constraint}. Hence every sampling stage plays a scaled candidate $a_s=(1-\rho_s)a_{\mathrm{safe}}+\rho_s b_s$.

\emph{Regret on sampling stages.} Write $Q_s$ for the number of queries executed in stage $s$, so that $Q_s=q_s$ at completed stages and $Q_s\le\min\{q_s,T\}$ always, and let $\mathcal I\subseteq[S]$ index the sampling stages. Then
\begin{align*}
\Reg(T)=T_{\mathrm{warm}}\Delta(a_{\mathrm{safe}})
  +\sum_{s\in\mathcal I}Q_s\Delta(a_s).
\end{align*}
Fix $s\in\mathcal I$. If $a_s=0$, \Cref{lem:stage_gap} gives $\Delta(0)\le 0$, so playing $0$ for the rest of the horizon adds no regret. If $a_s\ne 0$,
\begin{align*}
Q_s\Delta(a_s)
 &\le C_\Delta(B)\,B\beta_s\,Q_s\norm{a_s}_{V_s^{-1}}
 \le C_\Delta(B)\,B\beta_s\,\frac{Q_s}{n_s}\bigl(c_{\mathrm{len}}+\norm{a_s}_{V_s^{-1}}\bigr)\\
 &\le C_\Delta(B)\,B\,(c_{\mathrm{len}}+1)\,\beta_s\,\frac{Q_s}{n_s},
\end{align*}
where the first inequality is \Cref{lem:stage_gap} with~\eqref{eq:M_def}, the second uses $n_s\le c_{\mathrm{len}}/\norm{a_s}_{V_s^{-1}}+1$ from~\eqref{eq:ns_def}, and the third uses $\norm{a_s}_{V_s^{-1}}\le\|a_s\|_2\le 1$ since $V_s\succeq I$. By \Cref{lem:stage_estimator}, $Q_s/n_s\le\sqrt{m+1}\,L_s\le\sqrt{m+1}\,\bar L$ when $n_s\le T$ (recall $s\le S\le T$); when $n_s>T$, $Q_s/n_s\le T/n_s<1$. Summing over $s\in\mathcal I\subseteq[S]$ and using
\[
 \sum_{s=1}^S\beta_s=S+\sum_{j=1}^{S-1}\sqrt j\le S+\frac23S^{3/2}\qquad(\lambda=1)
\]
gives the first regret bound.

\emph{Stage count and rate.} The first $S-1$ stages completed their budgets, so by \Cref{lem:stage_estimator}
\[
 \sum_{s<S}n_s\le\frac{1}{\sqrt{m+1}}\sum_{s<S}q_s\le\frac{T}{\sqrt{m+1}},
\]
and \Cref{lem:stage_count} gives the bound on $S-1$. For the rate, $d_{\cA}\le d$ gives $S=O(d\log T)$, because $x\mapsto x\log(1+N^2/x)$ is increasing for every $N>0$; \eqref{eq:qrs-stage-budget} gives $B=\sqrt d+O(\sqrt{\log(d\log(2T)/\delta)})$, and \Cref{lem:stage_gap} gives $C_\Delta(B)\le 6+9\Delta(a_{\mathrm{safe}})/\Gamma$. The first term is therefore $\widetilde O(\sqrt{m+1}\,\sqrt d\,(d\log T)^{3/2}(1+\Delta(a_{\mathrm{safe}})/\Gamma))$. The warm-up term $\widetilde O(\sqrt{m+1}\,B\beta_S\Delta(a_{\mathrm{safe}})/\Gamma)$ of \Cref{lem:warmup} is of lower order because $\beta_S\le 1+\sqrt S$.
\end{proof}

\begin{remark}[Accuracy of the sampled subproblems]\label{rem:subproblem_accuracy}
Suppose each feasible sampled subproblem returns a sampled-feasible candidate ($\widetilde\Phi_s^{(k)}b_s^{(k)}\le\alpha$) whose objective value is within $\gamma\ge 0$ of its subproblem optimum, while infeasible subproblems are identified exactly, under the same conservative test~\eqref{eq:ucb_safe_constraint} and maximal scaling step $\rho_s$.

Whenever draw $k$ makes $a^\star$ sampled-feasible and optimistic, the selected candidate satisfies $\ip{\widetilde\theta_s}{b_s}\ge\mu^\star-\gamma$, adding $\gamma$ to the right-hand side of \Cref{lem:candidate_gap}. Because $b_s$ remains sampled-feasible, \Cref{lem:scaling_loss} applies unchanged, so the additional gap at the played action $a_s$ is at most $\rho_s\gamma\le\gamma$ (which also bounds the per-query loss if $a_s=0$ is played for the remaining horizon). Summing over at most $T$ queries adds at most $T\gamma$ to $\Reg(T)$, including when the final stage is truncated; in particular, choosing $\gamma=T^{-2}$ adds at most $1/T$ to the regret bound.
\end{remark}

\begin{remark}[Choice of $c_{\mathrm{len}}$]\label{cor:clen_opt}
For fixed $c=c_{\mathrm{len}}$, each stage contributes the factor $c+1$ to the leading term of \Cref{thm:main}, and \Cref{lem:stage_count} bounds $S$ by a multiple of $1/\log(1+c^2)$. The leading term is therefore proportional to $F(c)=(c+1)/\log^{3/2}(1+c^2)$. Since $F(c)\to\infty$ as $c\to 0$ and as $c\to\infty$, it attains a minimum on $(0,\infty)$; a numerical search places it at $c\approx 5.48$, where $F(c)\approx 1.02$.
\end{remark}

\subsubsection{Bounded multivariate quantum mean estimation}\label{sec:instantiations}
We prove \Cref{thm:isotropifier}, which gives the stage estimator of \Cref{lem:stage_estimator}.

\begin{proof}[Proof of \Cref{thm:isotropifier}]
We reduce to \textsc{QBounded}~\cite[Theorem~3.3]{cornelissen2022multivariate}. Let $Z'$ be a random vector in $\RR^M$ with $\|Z'\|_2\le 1$. \textsc{QBounded} takes a state-preparation unitary for the sample space of $Z'$ (with its inverse and controlled state reflections), a reversible binary value oracle that writes $Z'$ into a fresh register, and a parameter $n\ge 1$. Using $\widetilde O(n)$ calls to these primitives, where $\widetilde O$ hides a factor polylogarithmic in $n$ and $M$, it returns $\widehat\mu_{Z'}$ with
\[
 \Pr\Bigl[\|\widehat\mu_{Z'}-\EE[Z']\|_\infty>\frac{\log(M/\delta)}{n}\Bigr]\le\delta .
\]
This is the statement of~\cite{cornelissen2022multivariate} with their bound on $\EE[\|Z'\|_2]$ taken equal to one.

\emph{Primitives from the oracle.} Let $U$ be the sampling oracle of $Z$ (\Cref{subsec:mean-estimation-prelim}), which prepares $\sum_\omega\sqrt{p(\omega)}\ket{\omega}\ket{Z(\omega)}$, and regard the basis pair $(\omega,Z(\omega))$ as the sample label. Then $U$ and $U^\dagger$ implement the state-preparation unitary and its inverse. A controlled reflection $U(I-2\ket{0}\!\bra{0})U^\dagger$ needs the control only on the middle reflection, so it uses uncontrolled calls to $U$ and $U^\dagger$. In the bandit model of \Cref{def:q_oracle}, $U=\mathcal O_a$ for $Z=X(a)$.

\emph{Rescaling.} Let $D=2^{\lceil\log_2\sqrt M\rceil}$, so that $\sqrt M\le D<2\sqrt M$, and let $Z'=Z/D$. Since $Z\in[0,1]^M$, $\|Z'\|_2\le\sqrt M/D\le 1$. The value $Z(\omega)$ is part of the sample label and is given in binary, and $D$ is a power of two, so a reversible copy and a bit shift implement the value oracle for $Z/D$, its inverse, and their controlled versions without sampler queries.

\emph{Accuracy and cost.} Take $n=\lceil D\log(2M/\delta)/\varepsilon\rceil$ and output $\widehat\mu=D\widehat\mu_{Z'}$. With probability at least $1-\delta$,
\[
 \|\widehat\mu-\EE[Z]\|_\infty=D\,\|\widehat\mu_{Z'}-\EE[Z']\|_\infty\le\frac{D\log(M/\delta)}{n}\le\varepsilon .
\]
The number of sampler queries is $\widetilde O(n)$. Since $D<2\sqrt M$ and $\varepsilon\le 1$, we have $n\le 4\sqrt M\log(2M/\delta)/\varepsilon$, so this number is $\sqrt M/\varepsilon$ times a factor polylogarithmic in $M$, $1/\varepsilon$ and $1/\delta$.
\end{proof}

\subsection{An additional regret bound with exact global optimization}\label{sec:global-optimism-upper}

We next analyze a variant of QRS-COLTS that replaces the resampled Gaussian perturbations with deterministic global optimization over the reward and constraint confidence bounds, under the same classical-action coherent-feedback oracle (\Cref{def:q_oracle}). Write $u=a_{\mathrm{safe}}$ and $h_s(x)=\beta_s\|x\|_{V_s^{-1}}$. The stage estimates, design update, and horizon rule are those of \Cref{alg:qrs-colts}. At stage $s$, compute
$$
  \underline\Gamma_s=\min_{i\in[m]}\left\{\alpha_i-\ip{\widehat\phi_{s,i}}{u}\right\}-h_s(u).
$$
If $\underline\Gamma_s<6h_s(u)$, stage $s$ is a warm-up stage and $a_s=u$. Otherwise it is a sampling stage: choose a global maximizer
\begin{equation}\label{eq:global-optimism-subproblem}
 b_s\in\argmax_{x\in\cA}
       \left\{\ip{\widehat\theta_s}{x}+h_s(x)\right\}
 \quad\text{subject to}\quad
       \ip{\widehat\phi_{s,i}}{x}-h_s(x)\le\alpha_i\quad\forall i\in[m],
\end{equation}
and choose the largest $\rho_s\in[0,1]$ for which $a_s=(1-\rho_s)u+\rho_s b_s$ satisfies the conservative check~\eqref{eq:ucb_safe_constraint} (defaulting to $a_s=u$ if the feasible set of~\eqref{eq:global-optimism-subproblem} or the set of certified steps is empty). Note that~\eqref{eq:global-optimism-subproblem} maximizes a convex function over a generally nonconvex set and can be computationally intractable even for polyhedral $\cA$; the theorem below counts oracle queries assuming exact global optimization in~\eqref{eq:global-optimism-subproblem}.

\begin{theorem}[Safe regret with global optimization]\label{thm:global-optimism-upper}
Suppose $\cA\subseteq\RR^d$ is compact and convex, all actions $a\in\cA$, the reward parameter $\theta^\star$, and every constraint row $\phi_i^\star$ have Euclidean norm at most $1$, and the learner knows a strictly safe baseline $u\in\cA$ with true slack $\Gamma=\min_{i\in[m]}(\alpha_i-\ip{\phi_i^\star}{u})>0$ (without knowing $\Gamma$). For fixed $c_{\mathrm{len}}>0$, the algorithm above satisfies, with probability at least $1-\delta$, round-wise mean safety and
$$
 \Reg(T)=\widetilde O\!\left(\sqrt{m+1}\,d^{3/2}\log^{3/2}(2T)
                 \left(1+\frac{\Delta(u)}{\Gamma}\right)\right),
$$
where $\widetilde O(\cdot)$ suppresses logarithmic factors from the confidence parameter $1/\delta$ and the QME subroutine.
\end{theorem}

\begin{proof}
\emph{Safety.} Under the stage failure schedule~\eqref{eq:delta_s_def_main}, \Cref{lem:Con_explicit} ensures that $\mathrm{Con}(\delta)$ holds with probability at least $1-\delta$; we work on this event. Every queried action is either $u$ or passes~\eqref{eq:ucb_safe_constraint}, so \Cref{lem:safety} gives round-wise mean safety.

\emph{Stage gap.} Fix a sampling stage $s$, that is, $\underline\Gamma_s\ge 6h_s(u)$, and abbreviate $h=h_s$, $b=b_s$, $a=a_s$ and $\rho=\rho_s$. By~\eqref{eq:baseline-slack-sandwich}, $6h(u)\le\underline\Gamma_s\le\Gamma$. On $\mathrm{Con}(\delta)$, $a^\star$ is feasible for~\eqref{eq:global-optimism-subproblem}, whose feasible set is compact, so a global maximizer $b$ exists. With $F_i(x):=\ip{\widehat\phi_{s,i}}{x}+h(x)$,
\begin{align*}
 \mu^\star&\le\ip{\widehat\theta_s}{a^\star}+h(a^\star)\le\ip{\widehat\theta_s}{b}+h(b)\le\ip{\theta^\star}{b}+2h(b),\\
 F_i(b)&=\ip{\widehat\phi_{s,i}}{b}-h(b)+2h(b)\le\alpha_i+2h(b),\\
 F_i(u)&\le\ip{\phi_i^\star}{u}+2h(u)\le\alpha_i-\Gamma+\frac{\Gamma}{3}=\alpha_i-\frac{2\Gamma}{3},
\end{align*}
where the first line uses~\eqref{eq:con_reward} at $a^\star$, the optimality of $b$, and~\eqref{eq:con_reward} at $b$; the second uses the constraints of~\eqref{eq:global-optimism-subproblem}; and the third uses~\eqref{eq:con_constraints} and $2h(u)\le\Gamma/3$. The first line says that $\Delta(b)\le 2h(b)$. Thus \Cref{lem:scaling_generic} applies with $M=h$, $\kappa=\frac23$, $c=2$ and $\tau=\frac16$. Since $\kappa-c\tau=\frac13$, it gives
\[
 \Delta(a)\le\Bigl(4+6\,\frac{\Delta(u)}{\Gamma}\Bigr)h(a).
\]
This also covers $a=0$, whose remaining queries incur zero regret. On any warm-up stage, \eqref{eq:baseline-slack-sandwich} gives $\Gamma\le\underline\Gamma_s+2h_s(u)<8h_s(u)$, which rules out $u=0$.

\emph{Counting.} Let $S$ be the number of stages begun up to horizon $T$. By \Cref{lem:stage_estimator}, $q_s\ge n_s$, so \Cref{lem:stage_count} applied to the $S-1$ completed stages gives $S=O(d\log(2T))$ for fixed $c_{\mathrm{len}}$. By \Cref{lem:warmup} with $B'=2$, the warm-up stages use $\widetilde O(\sqrt{m+1}\,\beta_S/\Gamma)$ queries in total and incur regret $\widetilde O(\sqrt{m+1}\,\beta_S\Delta(u)/\Gamma)$. Every other stage with $a_s\ne 0$ executes $Q_s$ queries and adds regret
\[
 Q_s\Delta(a_s)
 \le\Bigl(4+6\,\frac{\Delta(u)}{\Gamma}\Bigr)\beta_s\,\frac{Q_s}{n_s}\,n_s\|a_s\|_{V_s^{-1}}
 =\widetilde O\left(\sqrt{m+1}\,\beta_s\Bigl(1+\frac{\Delta(u)}{\Gamma}\Bigr)\right),
\]
where the last step uses $n_s\|a_s\|_{V_s^{-1}}\le c_{\mathrm{len}}+1$ and $Q_s/n_s=\widetilde O(\sqrt{m+1})$; the latter holds because $Q_s/n_s\le q_s/n_s=\widetilde O(\sqrt{m+1})$ when $n_s\le T$, and $Q_s/n_s\le T/n_s<1$ otherwise. Summing $\beta_s=1+\sqrt{s-1}$ over $s\le S$ gives $O(S^{3/2})=O(d^{3/2}\log^{3/2}(2T))$, which proves the theorem.
\end{proof}

\subsection{Lower bounds for classical action selection}\label{sec:classical-action-lower-bounds}

\begin{theorem}[Dimension dependence with coherent feedback]\label{thm:convex-quantum-lower}
For every $d\ge 2$,\footnote{The restriction $d\ge 2$ here and in \Cref{thm:classical-lower} is necessary under inactive constraints and $[0,1]$-bounded rewards. When $d=1$, the mean reward $\theta^\star a$ is nonnegative for all $a\in\cA$, so the known compact set $\cA\subseteq\RR$ determines an optimal action: $\max\cA$ if $\cA\subseteq[0,\infty)$, $\min\cA$ if $\cA\subseteq(-\infty,0]$, and any action if $\cA$ contains both positive and negative numbers, which forces $\theta^\star=0$.} there is a compact convex action set $\cA$ in the unit ball of $\RR^d$ such that, for each $T\ge 1$ and every policy in the classical-action coherent-feedback model (\Cref{def:q_oracle}), some instance satisfies
$$
    \EE[\Reg(T)]\ge c\, d^{3/2}\log\!\left(1+T/d^{3/2}\right)
$$
for a universal constant $c>0$. Moreover, the instance has identically zero constraint feedback, a known baseline action of slack $1/2$, and $\|\theta^\star\|_2\le 1$.
\end{theorem}

\begin{proof}
\emph{The cap instance.} Liu, Li, and Lui~\cite[Theorem~3.1]{liu2026regretbounds} prove an $\Omega(d^{3/2}\log(1+T/d^{3/2}))$ lower bound on the expected regret of unconstrained quantum linear bandits over the nonconvex spherical cap
\[
 C=\{y\in\RR^d:\|y\|_2=1,\ y_1\ge\sqrt{3}/2\}.
\]
Their model is that of \Cref{def:q_oracle} with a scalar reward and no constraint: a query at $y\in C$ applies a reward oracle or its inverse, and we call this unitary the \emph{cap oracle} at $y$. Their instances have Bernoulli rewards with means $\ip{\theta^\star}{y}\in[1/4,3/4]$ and parameter norm $\|\theta^\star\|_2\le 5/8$. We embed this family in the compact convex solid cap
\[
 \cA=\{x\in\RR^d:\|x\|_2\le 1,\ x_1\ge\sqrt{3}/2\},
\]
with constraint vector $\phi^\star=0$, threshold $\alpha=1/2$, and $a_{\mathrm{safe}}=e_1\in\cA$, which has slack $\Gamma=1/2$. Every $x\in\cA$ has $r:=\|x\|_2\in[\sqrt3/2,1]$, since $\|x\|_2\ge x_1\ge\sqrt3/2$. The point $y:=x/r$ then satisfies $\|y\|_2=1$ and $y_1=x_1/r\ge x_1\ge\sqrt{3}/2$, so $y\in C$.

\emph{Simulating a query.} Given a classical-action policy on $\cA$, we simulate a query at $x=ry\in\cA$ with one query at $y\in C$. On fresh registers:
\begin{enumerate}[label=(\arabic*)]
\item apply the cap oracle at $y$, which prepares an outcome bit $R_y\sim\mathrm{Bern}(\ip{\theta^\star}{y})$;
\item independently prepare $\sqrt{1-r}\ket{0}+\sqrt r\ket{1}$, whose basis bit is $J\sim\mathrm{Bern}(r)$;
\item reversibly write $J\cdot R_y$ into the reward output register and append $\ket{0}$ for the constraint coordinate, keeping the cap seed, $R_y$ and $J$ in the latent seed register.
\end{enumerate}
This unitary uses one forward cap query, and its inverse uses one inverse cap query. Its reward feedback is Bernoulli with mean $\EE[JR_y]=r\ip{\theta^\star}{y}=\ip{\theta^\star}{x}$.

\emph{Comparing gaps.} Since $C\subseteq\cA$ and $\ip{\theta^\star}{x}=r\ip{\theta^\star}{y}\le\ip{\theta^\star}{y}$ (as $r\le 1$ and $\ip{\theta^\star}{y}\ge 1/4>0$), both sets have the same optimal mean reward, $\max_{x\in\cA}\ip{\theta^\star}{x}=\max_{y\in C}\ip{\theta^\star}{y}$. At every coupled query,
\[
 \Delta(x)-\Delta(y)=(1-r)\,\ip{\theta^\star}{y}\ge 0 .
\]
Thus the induced policy on $C$ has cumulative regret at most that of the policy on $\cA$, and~\cite[Theorem~3.1]{liu2026regretbounds} gives the stated lower bound. Appending $m-1$ zero constraint coordinates extends the result to any $m\ge 1$.
\end{proof}

\begin{theorem}[Cost of strict safety]\label{thm:slack-lower}
Fix $T\ge 1$, $0<\Gamma\le 1/8$, and $0<b\le 1/2$. In the classical-action coherent-feedback model (\Cref{def:q_oracle}), there exist two instances with $d=2$, $m=1$, and a common known baseline $a_{\mathrm{safe}}$ of exact slack $\Gamma$ such that every policy satisfying round-wise mean safety with probability at least $3/4$ on both instances incurs, on the first instance,
$$
 \Reg(T)\ge \frac b2\min\!\left\{T,1+\left\lfloor\frac{1}{12\Gamma}\right\rfloor\right\}
 \quad\text{with probability at least }1/2 .
$$
Moreover, $b=\Delta(a_{\mathrm{safe}})$ on the first instance, so the expected regret there is $\Omega(\Delta(a_{\mathrm{safe}})\min\{T,1/\Gamma\})$, even if the learner knows $\Gamma$ exactly.
\end{theorem}

\begin{proof}
\emph{Instances.} Let $\cA=\{a(z):=(1,z)/\sqrt 2 : z\in[0,1]\}\subseteq\RR^2$, with $a_{\mathrm{safe}}=a(0)$, threshold $\alpha=1/2$, and common reward vector $\theta^\star=\sqrt 2(0,b)$, so that the mean reward is $\ip{\theta^\star}{a(z)}=bz$. The two instances differ only in their constraint vectors, whose expected constraint values $p_\iota(z)=\ip{\phi_\iota}{a(z)}$ are
\begin{align*}
 \phi_0&=\sqrt 2\bigl(\tfrac12-\Gamma,0\bigr), & p_0(z)&=\tfrac12-\Gamma,\\
 \phi_1&=\sqrt 2\bigl(\tfrac12-\Gamma,2\Gamma\bigr), & p_1(z)&=\tfrac12-\Gamma+2\Gamma z .
\end{align*}
Since $z\in[0,1]$, $b\le 1/2$ and $\Gamma\le 1/8$, all actions, $\theta^\star$, $\phi_0$ and $\phi_1$ have Euclidean norm at most $1$, and both instances have exact baseline slack $\alpha-p_\iota(0)=\Gamma$. Under instance~0, every $z\in[0,1]$ is feasible, $a(1)$ is optimal, and $\Delta(a_{\mathrm{safe}})=b$. Under instance~1, the feasible actions are those with $z\le 1/2$.

\emph{Per-query distance.} Both instances use the same Bernoulli reward unitaries. The constraint coordinate is produced by the single-qubit rotation $R(f(p_\iota(z)))$ with angle $f(p)=\arcsin\sqrt p$; the rotated bit is copied into the constraint output register, and the original bit stays in the seed register. Let $U_\iota(z)$ be the resulting single-query unitary of instance $\iota$ at $a(z)$. Then
\[
 \bigl\|U_1(z)-U_0(z)\bigr\|_{\mathrm{op}}
 \le\bigl|f(p_1(z))-f(p_0(z))\bigr|
 \le\frac{2}{\sqrt 3}\cdot 2\Gamma z
 \le\frac{4\Gamma}{\sqrt 3}\le 3\Gamma ,
\]
where the first inequality holds because the two unitaries differ only in the rotation angle, and the second uses $p_0(z),p_1(z)\in[3/8,5/8]$ and $|f'(p)|=1/(2\sqrt{p(1-p)})\le 2/\sqrt 3$ on $[1/4,3/4]$. The same bound holds for the adjoints.

\emph{Hybrid argument.} For $n\in[T]$, let $H_n$ be the event that one of the first $n$ queried actions has $z>1/2$. Round-wise safety on instance~1 requires $\Pr\nolimits_1[H_n]\le 1/4$, and $H_n$ is determined before the $n$-th oracle response.

For any input state, including one entangled with a reference register, one forward or inverse query changes the trace distance $\tfrac12\|\cdot\|_1$ by at most $3\Gamma$; mixed states follow by purification, uniformly over the classically chosen action. Replacing the first $n-1$ queries one at a time, and using that trace distance does not increase under the policy's operations, gives
\[
 \Pr\nolimits_0[H_n]\le \Pr\nolimits_1[H_n]+3(n-1)\Gamma .
\]
With $n=\min\{T,1+\lfloor 1/(12\Gamma)\rfloor\}$, we have $3(n-1)\Gamma\le 1/4$ and hence $\Pr\nolimits_0[H_n]\le 1/2$. On $H_n^c$, which has probability at least $1/2$ under instance~0, each of the first $n$ queries has $z_t\le 1/2$ and gap $b(1-z_t)\ge b/2$, and later queries have nonnegative gap. This proves both bounds.
\end{proof}

\begin{theorem}[Horizon dependence with classical feedback]\label{thm:classical-lower}
For every $d\ge 2$ and $T\ge(d-1)^2$, there is a compact convex action set $\cA$ in the unit ball of $\RR^d$ such that, for every policy in the classical-action classical-feedback model (\Cref{def:noncoherent-feedback}), some instance satisfies
$$
    \EE[\Reg(T)]\ge c\,(d-1)\sqrt T
$$
for a universal constant $c>0$. Moreover, the instance has a single identically zero constraint coordinate, a known baseline action of slack $1/2$, and $\|\theta^\star\|_2\le 1$.
\end{theorem}

\begin{proof}
With classical feedback, any quantum processing between queries can be simulated by a classical randomized policy that sees the same transcript, so it suffices to consider classical policies. Let $k=d-1$ and $c_0=1/16$. For $z\in[-1,1]^k$ and $v\in\{-1,1\}^k$, let
\begin{align*}
 a(z)&=\Bigl(\frac{1}{\sqrt2},\frac{z}{\sqrt{2k}}\Bigr), &
 \theta_v&=\Bigl(\frac{1}{\sqrt2},c_0\sqrt{\frac{2k}{T}}\,v\Bigr), &
 \ip{\theta_v}{a(z)}&=\frac12+\frac{c_0}{\sqrt T}\,v^\top z ,
\end{align*}
with Bernoulli rewards of these means, constraint vector $\phi^\star=0$, and threshold $\alpha=1/2$. The set $\cA=\{a(z):z\in[-1,1]^k\}$ is compact and convex with $\|a(z)\|_2\le 1$. Since $T\ge k^2$, $\|\theta_v\|_2^2=\frac12+2c_0^2k^2/T\le 1$ and every mean lies in $[1/4,3/4]$. Every action is feasible, with slack $1/2$ at $a_{\mathrm{safe}}=a(0)$.

Up to the constant $1/2$, which does not affect the regret, this is the hypercube instance of~\cite[Theorem~24.1]{lattimore2020bandit} on $[-1,1]^k$, with gap parameter $c_0/\sqrt T$ in place of $1/\sqrt T$; see also~\cite[Theorem~3]{dani2008stochastic}. That theorem assumes Gaussian noise. Its proof combines the Bretagnolle--Huber inequality with the divergence decomposition of the relative entropy between the transcript laws of two parameters that differ in one coordinate, and the noise enters only through the relative entropy of one observation, which is $(p-q)^2/2$ for unit-variance Gaussian rewards with means $p$ and $q$. For Bernoulli rewards with means $p,q\in[1/4,3/4]$,
\[
 \mathrm{KL}(p\|q)\le\frac{(p-q)^2}{q(1-q)}\le\frac{16}{3}\,(p-q)^2 ,
\]
a constant multiple of the Gaussian value. The same proof therefore gives some $v$ with $\EE[\Reg(T)]\ge c\,k\sqrt T=c\,(d-1)\sqrt T$ for a universal constant $c>0$.
\end{proof}

\section{Query-weighted safe-action bomb model: algorithm and proofs}
\label{app:bomb-projection}

This appendix proves the guarantee for the coherent-action bomb model stated informally in \Cref{thm:informal-bomb-main}. \Cref{sec:bco-overview} describes the algorithm.

\subsection{Overview}\label{sec:bomb-main}

We use the definitions of \Cref{sec:coherent-action-bomb-model}: the safe set $\cS$, the clipped gap $\Delta_+$, the safe-action bomb oracle (\Cref{def:safe-action-bomb-oracle}), and the query-weighted regret $R_{\mathrm w}(T)$ (\Cref{def:query-weighted-regret-main}). A query whose pre-query state has action marginal $\pi_t$ triggers an explosion with conditional probability $\pi_t(\cA\setminus\cS)$, and we bound the cumulative explosion probability separately from $R_{\mathrm w}(T)$.

In \Cref{thm:bco-qlints}, we prove that under unit $\ell_2$-norm bounds and a well-rounded safe set $\mathbb{B}(a_{\mathrm{safe}},r)\subseteq\cS\subseteq\mathbb{B}(a_{\mathrm{safe}},R)$ (\Cref{ass:safe-body}), BCO-QLinTS has cumulative explosion probability at most $\varepsilon$ and, with probability at least $1-\delta$, achieves query-weighted regret
$$
    R_{\mathrm w}(T)
    \le
    \widetilde O\!\left(d^2\log^{3/2}(2T/\delta)\right)
    +
    \widetilde O\!\left(\frac{d^4\log^2(2T)}{\varepsilon}\operatorname{polylog}(2+R/r,2/\delta)\right)
$$
up to termination. The first term is the cost of learning $\theta^\star$, and the second is the cost of the bomb-certified optimization, in which the $m$ constraints enter only through the ratio $R/r$. The loss differs from the played-action regret of QRS-COLTS, so the two bounds are not directly comparable.

\subsection{Bomb-certified safe optimization}

To optimize linear objectives over $\cS$ using membership queries, we require $\cS$ to be well-rounded around the known safe baseline $a_{\mathrm{safe}}$.

\begin{assumption}[Well-rounded safe set]
\label{ass:safe-body}
The safe set $\cS$ is a closed convex body, and the learner knows a safe point $a_{\mathrm{safe}}\in\cS$ and radii $0<r\le R$ such that
$$
    \mathbb{B}(a_{\mathrm{safe}},r)\subseteq \cS\subseteq \mathbb{B}(a_{\mathrm{safe}},R).
$$
Actions are stored in finite precision. For every spacing $\varrho>0$, the points of the grid $G_\varrho:=a_{\mathrm{safe}}+\varrho\,\mathbb Z^d$ have binary labels, the oracle of \Cref{def:safe-action-bomb-oracle} accepts the labels of $G_\varrho\cap\cA$ in its action register, and membership of a grid point in the known action domain $\cA$ can be computed reversibly without oracle queries. Every point of $\RR^d$ is within distance $\varrho\sqrt d/2$ of $G_\varrho$, and $a_{\mathrm{safe}}\in G_\varrho$.
\end{assumption}

We first state an exact-feasibility quantum convex optimization primitive using standard coherent membership queries, derived from the weak optimization algorithm of van Apeldoorn et al.~\cite{vanapeldoorn2020convexopt} via inward contraction toward $a_{\mathrm{safe}}$.

\begin{lemma}[Quantum convex optimization with membership queries]
\label{lem:qco}
Under \Cref{ass:safe-body}, suppose the learner has coherent membership access to $\cS$, i.e.,
$$
    \ket{a}\ket{b}\mapsto \ket{a}\ket{b\oplus \one\{a\notin\cS\}}.
$$
For any linear objective $c\in\RR^d$, accuracy $\gamma\in(0,1)$, and failure probability $\delta\in(0,1)$, there is a quantum algorithm that, with probability at least $1-\delta$, outputs a grid point $\widehat a\in\cS$ satisfying
$$
    c^\top \widehat a
    \ge
    \max_{a\in\cS}c^\top a-\gamma,
$$
using
$$
    M_{\mathrm{qco}}(d,R/r,\gamma,\delta)
    =
    \widetilde O\!\left(
        d\,\polylog\!\left(\frac{R}{r},\frac1\gamma,\frac1\delta\right)
    \right)
$$
coherent membership queries, where $\widetilde O(\cdot)$ hides logarithmic factors in $d$ and $2+R\|c\|_2/\gamma$. The count depends on $\|c\|_2$ only through these logarithms and is nondecreasing in $\|c\|_2$; we do not list this argument, and \Cref{thm:bco-qlints} evaluates $M_{\mathrm{qco}}$ at a bound on the objective norm.
\end{lemma}

\begin{proof}
Write $u=a_{\mathrm{safe}}$. If $c=0$, returning $\widehat a=u\in\cS$ achieves zero error with zero queries; henceforth assume $c\ne 0$ and set $v=c/\|c\|_2$.

\emph{Weak optimization and grid rounding.} Fix a tolerance $\tau\in(0,r/2]$, chosen below. Run the weak linear optimizer of van Apeldoorn et al.~\cite{vanapeldoorn2020convexopt}, with the guarantee of their Definition~5, to maximize $v^\top x$ over $\cS$ with tolerance $\tau/2$ and failure probability $\delta$. This optimizer combines their Theorem~23, which implements a separation oracle with $\widetilde O(1)$ queries to a membership oracle, with the classical reduction from optimization to $\widetilde O(d)$ separation calls recalled in their introduction. Their membership oracle (their Definition~3) is a weak one: for a precision $\varepsilon_{\mathrm{mem}}>0$ fixed by their Theorem~23, on input $w\in\RR^d$ it must correctly assert either $\operatorname{dist}(w,\cS)\le\varepsilon_{\mathrm{mem}}$ or $\mathbb B(w,\varepsilon_{\mathrm{mem}})\not\subseteq\cS$. The coherent membership map provides it. Round $w$ to a nearest point $w'$ of a grid $G_\varrho$ with $\varrho\sqrt d/2\le\varepsilon_{\mathrm{mem}}$ and return $\one\{w'\notin\cS\}$: if $w'\in\cS$ then $\operatorname{dist}(w,\cS)\le\varepsilon_{\mathrm{mem}}$, and otherwise $\mathbb B(w,\varepsilon_{\mathrm{mem}})\not\subseteq\cS$. The optimizer's comparison set $\{x:\mathbb B(x,\tau/2)\subseteq\cS\}$ contains $u$, since $\tau/2<r$, so it is nonempty. With probability at least $1-\delta$, the optimizer returns a point $y_0\in\RR^d$ with
\begin{align*}
 \operatorname{dist}(y_0,\cS)&\le\tau/2, &
 v^\top x-v^\top y_0&\le\tau/2\quad\text{whenever }\mathbb B(x,\tau/2)\subseteq\cS.
\end{align*}
The output is stored on the grid $G_\varrho$ with $\varrho=\tau/(2\sqrt d)$: round $y_0$ to a nearest grid point $y$, so that $\|y-y_0\|_2\le\tau/4$. Then
\begin{align*}
 \operatorname{dist}(y,\cS) &\le \tau/2+\tau/4\le\tau,\\
 v^\top x-v^\top y &\le \tau/2+\tau/4\le\tau
 \qquad\text{for all }x\text{ with }\mathbb B(x,\tau)\subseteq\cS,
\end{align*}
where the first line follows from the triangle inequality on $\operatorname{dist}(y_0,\cS)\le\tau/2$ and $\|y-y_0\|_2\le\tau/4$, and the second line uses $v^\top x-v^\top y_0\le\tau/2$ (since $\mathbb B(x,\tau)\subseteq\cS$ implies $\mathbb B(x,\tau/2)\subseteq\cS$) together with $|v^\top(y_0-y)|\le\|v\|_2\|y_0-y\|_2\le\tau/4$.

\emph{Inward contraction for exact feasibility.} The point $y$ may lie slightly outside $\cS$, so we pull it toward the interior center $u$ until a small ball around it lies inside $\cS$. Let $z\in\cS$ be a point with $\|y-z\|_2\le\tau$; it is used only in the analysis. Define
\[
 \sigma=\frac{r}{r+2\tau},\qquad p=\sigma y+(1-\sigma)u .
\]
For any point $p'\in\mathbb B(p,\sigma\tau)$, we write $p'$ as a convex combination and bound its second point:
\begin{gather*}
 p'=\sigma z+(1-\sigma)
 \left(u+\frac{\sigma(y-z)+p'-p}{1-\sigma}\right),\qquad
 \left\|\frac{\sigma(y-z)+p'-p}{1-\sigma}\right\|_2
 \le \frac{2\sigma\tau}{1-\sigma}=r,
\end{gather*}
where the identity rearranges $p=\sigma y+(1-\sigma)u$, the inequality uses the triangle inequality with $\|y-z\|_2\le\tau$ and $\|p'-p\|_2\le\sigma\tau$, and the final equality substitutes $\sigma=r/(r+2\tau)$. Since $z\in\cS$ and the point in parentheses lies in $\mathbb B(u,r)\subseteq\cS$, convexity of $\cS$ gives $\mathbb B(p,\sigma\tau)\subseteq\cS$. Since $\tau\le r/2$, we have $\sigma\ge 1/2$, so rounding $p$ to a nearest point $\widehat a$ of the same grid, within distance $\tau/4<\sigma\tau$, gives $\widehat a\in\mathbb B(p,\sigma\tau)\subseteq\cS$.

\emph{Objective comparison and query count.} Let $x^\star\in\argmax_{x\in\cS}v^\top x$ and $x'=(1-\tau/r)x^\star+(\tau/r)u$. Mixing $x^\star\in\cS$ with every point of $\mathbb B(u,r)\subseteq\cS$, with weights $1-\tau/r$ and $\tau/r$, shows by convexity that $\mathbb B(x',\tau)\subseteq\cS$. Then
\begin{align*}
 v^\top(x^\star-\widehat a)
 &\le \|x^\star-x'\|_2+v^\top(x'-y)+\|y-p\|_2+\|p-\widehat a\|_2\\
 &\le \frac{R\tau}{r}+\tau+\frac{2\tau(R+\tau)}{r+2\tau}+\frac{\tau}{4}
 \le 4\tau(1+R/r).
\end{align*}
The first inequality decomposes $x^\star-\widehat a=(x^\star-x')+(x'-y)+(y-p)+(p-\widehat a)$ and applies the Cauchy--Schwarz inequality with $\|v\|_2=1$ to the first, third and fourth terms. The second inequality bounds the four terms as follows:
\begin{itemize}%
\item $\|x^\star-x'\|_2=(\tau/r)\|x^\star-u\|_2\le R\tau/r$, since $x^\star\in\mathbb B(u,R)$;
\item $v^\top(x'-y)\le\tau$, since $\mathbb B(x',\tau)\subseteq\cS$;
\item $\|y-p\|_2=(1-\sigma)\|y-u\|_2\le\frac{2\tau}{r+2\tau}(R+\tau)$, since $\|y-u\|_2\le\|y-z\|_2+\|z-u\|_2\le\tau+R$;
\item $\|p-\widehat a\|_2\le\tau/4$ by the rounding.
\end{itemize}
The last inequality uses $\frac{2(R+\tau)}{r+2\tau}\le\frac{2R}{r}+1$. Choosing
\[
 \tau=\min\left\{\frac r2,\frac{\gamma}{4(1+R/r)\|c\|_2}\right\}
\]
and multiplying by $\|c\|_2$ gives $c^\top(x^\star-\widehat a)\le\gamma$. Finally, translating $\cS$ by $-u$ and rescaling by $1/R$ places the body inside the unit ball with inner radius $r/R$ and normalized tolerance $\tau/(2R)$. The optimizer of~\cite{vanapeldoorn2020convexopt} then uses $\widetilde O(d)$ membership queries, with polylogarithmic dependence on $2R/\tau$ and $1/\delta$, which gives the bound $M_{\mathrm{qco}}(d,R/r,\gamma,\delta)$.
\end{proof}

Because every membership query in our third model risks triggering a bomb explosion, we simulate the coherent membership queries of \Cref{lem:qco} using the bomb-query framework of Lin and Lin~\cite{lin2014bomb}. We first recall their bomb-query model and adapt their simulation theorem.

\begin{definition}[Bomb query to a Boolean predicate]
\label{def:linlin-bomb-query}
Let $h:\mathcal X\to\{0,1\}$ be a Boolean predicate whose coherent oracle acts as
$$
    O_h:\ket{x}\ket{b}\mapsto \ket{x}\ket{b\oplus h(x)} .
$$
In the circuit formulation of Lin and Lin~\cite{lin2014bomb}, a \emph{controlled bomb query} to $h$ with control qubit $\ket{e}\in\mathrm{span}\{\ket{0},\ket{1}\}$ computes $e\cdot h(x)$ into a fresh flag qubit initialized to $\ket{0}$ and immediately measures the flag qubit in the computational basis: outcome $1$ triggers a bomb explosion and terminates the algorithm, whereas outcome $0$ projects onto the subspace with $e\cdot h(x)=0$. A bomb-query algorithm is required to keep its cumulative explosion probability across all queries within a prescribed budget.
\end{definition}

\begin{theorem}[Bomb-query simulation, adapted from {\cite[Theorem~4.2]{lin2014bomb}}]
\label{thm:linlin-simulation}
Let $\mathcal Q$ be a quantum query algorithm that makes $Q$ coherent queries to a Boolean oracle $O_h$. For any explosion budget $\eta\in(0,1)$ and state-error tolerance $\varsigma\in(0,1)$, there is a bomb-query algorithm that simulates $\mathcal Q$ with total explosion probability at most $\eta$ and Euclidean distance at most $\varsigma$ between the surviving (subnormalized) final state $\psi_{\rm surv}$ and the ideal final state $\psi_{\rm ideal}$, using
$$
    O\!\left(\frac{Q^2}{\min\{\eta,\varsigma\}}\right)
$$
controlled bomb queries.
\end{theorem}

\begin{proof}
As in the proof of~\cite[Theorem~4.2]{lin2014bomb}, we replace every call to $O_h$ by a repeated-rotation gadget with $\Lambda\ge 1$ rounds. Let $\vartheta=\pi/(2\Lambda)$; a rotation of a qubit by the angle $\vartheta$ maps $\ket{0}$ to $\cos\vartheta\ket{0}+\sin\vartheta\ket{1}$. On a basis state $\ket{x}\ket{b}$ and a fresh control qubit $\ket{0}$, the gadget does the following:
\begin{enumerate}[label=(\arabic*)]
\item $\Lambda$ times: rotate the control qubit by $\vartheta$, then make a controlled bomb query to $h$ at $x$;
\item flip $b$ if the control qubit is $\ket{0}$;
\item $\Lambda$ times: rotate the control qubit by $-\vartheta$, then make a controlled bomb query to $h$ at $x$.
\end{enumerate}

\emph{One call.} If $h(x)=0$, no query explodes. The rotations of step~(1) add up to $\pi/2$, so the control qubit is $\ket{1}$ after step~(1) and $\ket{0}$ after step~(3), and $b$ is unchanged. If $h(x)=1$, every round rotates $\ket{0}$ to $\cos\vartheta\ket{0}\pm\sin\vartheta\ket{1}$, and on survival the query keeps the branch $\ket{0}$, with amplitude $\cos\vartheta$. Hence $b$ is flipped and the control qubit ends in $\ket{0}$. By linearity, on survival the gadget acts on the registers of $\mathcal Q$ as $G=D\,O_h$, where $D$ multiplies the basis states $\ket{x}\ket{b}$ with $h(x)=1$ by $\cos^{2\Lambda}\vartheta$ and fixes the others. Therefore
\begin{align*}
 \|G-O_h\|_{\mathrm{op}}&\le 1-\cos^{2\Lambda}\vartheta\le\Lambda\sin^2\vartheta\le\frac{\pi^2}{4\Lambda},\\
 \|G\phi\|_2^2&\ge\cos^{4\Lambda}\vartheta\,\|\phi\|_2^2\quad\text{for every vector }\phi,
\end{align*}
where the first line uses $1-x^{\Lambda}\le\Lambda(1-x)$ for $x=\cos^2\vartheta$ and $\sin\vartheta\le\vartheta$.

\emph{All calls.} The surviving final state $\psi_{\rm surv}$ is obtained from $\psi_{\rm ideal}$ by replacing each of the $Q$ calls to $O_h$ by $G$. Hence
\begin{align*}
 \|\psi_{\rm surv}-\psi_{\rm ideal}\|_2&\le Q\,\|G-O_h\|_{\mathrm{op}}\le\frac{\pi^2Q}{4\Lambda},\\
 \Pr[\text{explosion}]&=1-\|\psi_{\rm surv}\|_2^2\le 1-\cos^{4\Lambda Q}\vartheta\le\frac{\pi^2Q}{2\Lambda},
\end{align*}
where the first line is a telescoping sum over the $Q$ calls, in which the other operations of $\mathcal Q$ are unitary and $\|G\|_{\mathrm{op}}\le 1$, and the second line uses that each call keeps at least the fraction $\cos^{4\Lambda}\vartheta$ of the squared norm, together with $1-x^{2\Lambda Q}\le 2\Lambda Q(1-x)$ for $x=\cos^2\vartheta$. Taking $\Lambda=\lceil\pi^2Q/(2\min\{\eta,\varsigma\})\rceil$ gives explosion probability at most $\eta$ and state error at most $\varsigma$, using $2Q\Lambda=O(Q^2/\min\{\eta,\varsigma\})$ controlled bomb queries.
\end{proof}

\paragraph{Implementing controlled bomb queries from \Cref{def:safe-action-bomb-oracle}.}
On the optimization grid, define $g(x):=\one\{x\notin\cA\}$ and $h(x):=\one\{x\in\cA\setminus\cS\}$. Because $\cS\subseteq\cA$, the two predicates have disjoint supports and satisfy $g(x)\oplus h(x)=\one\{x\notin\cS\}$, so each ideal coherent membership query in \Cref{lem:qco} factors as
$$
    O_{\one\{\cdot\notin\cS\}}=O_g O_h.
$$
Implementing $O_g$ reversibly without oracle calls via \Cref{ass:safe-body} rewrites the convex optimizer of \Cref{lem:qco} as a quantum algorithm making the same number of coherent queries to $O_h$, interleaved with known unitaries. We apply \Cref{thm:linlin-simulation} directly to this rewritten algorithm, whose sole unknown predicate is $h$ (which vanishes on every label outside $\cA$).

The safe-action bomb oracle of \Cref{def:safe-action-bomb-oracle} has no external control wire and also acts on the reward register. Two calls to it implement one controlled bomb query to $h(x)=\one\{x\in\cA\setminus\cS\}$ on any basis branch $\ket{x}\ket{e}_{\mathrm{ctrl}}$, as follows:
\begin{enumerate}[label=(\arabic*)]
\item initialize an auxiliary action register and reversibly write $a_{\mathrm{aux}}=x$ if $e=1$ and $x\in\cA$, and $a_{\mathrm{aux}}=a_{\mathrm{safe}}$ otherwise, leaving $\ket{x}\ket{e}_{\mathrm{ctrl}}$ unchanged;
\item make one forward call to \Cref{def:safe-action-bomb-oracle} on $a_{\mathrm{aux}}$ with blank flag and reward registers $\ket{0}_f\ket{0}_X$, and measure the flag;
\item conditioned on survival (outcome $0$), make one inverse call to \Cref{def:safe-action-bomb-oracle} on $a_{\mathrm{aux}}$ and then uncompute $a_{\mathrm{aux}}$.
\end{enumerate}
In step~(2), all branches with $e=0$ or $x\notin\cA$ query $a_{\mathrm{aux}}=a_{\mathrm{safe}}\in\cS$ and survive with certainty. An explosion therefore occurs if and only if $e=1$ and $x\in\cA\setminus\cS$, as in \Cref{def:linlin-bomb-query}. On the surviving state after step~(2), every branch has $a_{\mathrm{aux}}\in\cS$. The inverse call in step~(3) therefore returns $\ket{\chi_{a_{\mathrm{aux}}}}_X$ to $\ket{0}_X$ and writes $f(a_{\mathrm{aux}})=0$ into a fresh flag qubit, which measures $0$ with certainty. All ancillas are disentangled, at a cost of two oracle calls per controlled bomb query.

\begin{lemma}[Bomb-certified safe optimization]
\label{lem:bomb-safe-opt}
Fix objective $c\in\RR^d$, accuracy $\gamma\in(0,1)$, failure probability $\delta\in(0,1)$, and explosion budget $\eta\in(0,1]$. There is a bomb-query implementation of the optimizer in \Cref{lem:qco} with total explosion probability at most $\eta$ such that every point $\widehat a$ it returns lies in $\cS$ and, conditional on returning without an explosion, satisfies
$$
    c^\top \widehat a
    \ge
    \max_{a\in\cS}c^\top a-\gamma
$$
with probability at least $1-\delta$, using at most
$$
    B_{\mathrm{opt}}(d,R/r,\gamma,\delta,\eta)
    =
    \widetilde O\!\left(
        \frac{M_{\mathrm{qco}}(d,R/r,\gamma,\delta)^2}{\eta}
    \right)
$$
queries to the oracle of \Cref{def:safe-action-bomb-oracle}.
\end{lemma}

\begin{proof}
If $c=0$, returning $a_{\mathrm{safe}}\in\cS$ requires zero queries and satisfies all claims. Assume $c\ne 0$, and let $\mathcal Q$ denote the algorithm of \Cref{lem:qco} instantiated with constant failure probability $1/10$, which uses $Q=M_{\mathrm{qco}}(d,R/r,\gamma,1/10)$ coherent membership queries. We write $\mathcal Q$ as a unitary query algorithm followed by a measurement of its output register: its intermediate measurements are deferred, its classical steps are computed reversibly, and by the factorization above its queries are calls to $O_h$. We boost its success probability and certify exact safety of the returned point in two phases across $J=\lceil\log_2(4/\delta)\rceil$ independent trials, halting immediately upon any explosion.

\emph{Phase 1: repeated Lin--Lin simulation.} In each trial $j\in[J]$, we simulate $\mathcal Q$ via \Cref{thm:linlin-simulation} with explosion budget and state-error tolerance both equal to $\eta'=\eta/(5J)$. Let $P$ be the projector onto the basis states of the output register that encode feasible, $\gamma$-optimal points in $\cS$. In a single trial,
\begin{gather*}
 \big|\|P\psi_{\rm surv}\|_2^2-\|P\psi_{\rm ideal}\|_2^2\big|\le 2\|\psi_{\rm surv}-\psi_{\rm ideal}\|_2\le 2\eta'\le\frac25,\\
 \Pr[\text{survive simulation and output a feasible, }\gamma\text{-optimal point}]= \|P\psi_{\rm surv}\|_2^2 \ge \frac{9}{10}-\frac25= \frac{1}{2},
\end{gather*}
where the first line uses $\|a\|_2^2-\|b\|_2^2=(\|a\|_2+\|b\|_2)(\|a\|_2-\|b\|_2)\le 2\|a-b\|_2$ for vectors of norm at most $1$, together with $\|P\|_{\mathrm{op}}\le 1$ and \Cref{thm:linlin-simulation}. The second line uses $\|P\psi_{\rm ideal}\|_2^2\ge 9/10$, which is \Cref{lem:qco} with failure probability $1/10$.

\emph{Phase 2: repeated-rotation safety verification.} The optimizer fails with probability up to $1/10$, and the simulation error of Phase~1 adds to this, so a trial can output a point outside $\cS$. We therefore verify each trial's output $a$ before accepting it. Invalid encodings and points outside $\cA$ are discarded locally without oracle queries. For each candidate $a\in\cA$, we run the repeated-rotation Elitzur--Vaidman bomb test~\cite{Elitzur:1993xh,lin2014bomb}:
\begin{enumerate}[label=(\arabic*)]
\item initialize a control qubit in $\ket{0}$;
\item in each of $N=\lceil\pi^2 J/(2\eta)\rceil$ rounds, rotate the control qubit by the angle $\vartheta=\pi/(2N)$ and make one controlled bomb query at $a$, controlled on the qubit being $\ket{1}$;
\item after $N$ surviving rounds, measure the control qubit and accept $a$ if and only if the outcome is $1$.
\end{enumerate}
If $a\in\cS$, every controlled bomb query acts as the identity and survives with certainty. The $N$ rotations then add up to $N\vartheta=\pi/2$, which maps $\ket{0}$ to $\ket{1}$, so the test accepts $a$ with probability $1$.

If $a\in\cA\setminus\cS$, each round starts from $\ket{0}$, and the rotation places amplitude $\sin\vartheta$ on $\ket{1}$. Measuring the bomb flag then explodes with probability $\sin^2\vartheta$; on survival, which has probability $\cos^2\vartheta$, the control qubit is projected back to $\ket{0}$. Thus an unsafe $a$ is accepted with probability $0$, and its explosion probability over the $N$ rounds is
\begin{align*}
 1-\cos^{2N}\!\vartheta
 \le N(1-\cos^2\!\vartheta)
 = N\sin^2\!\vartheta
 \le N\vartheta^2
 = \frac{\pi^2}{4N}\le\frac{\eta}{2J},
\end{align*}
where the first inequality uses $1-x^N\le N(1-x)$ for $x=\cos^2\vartheta\in[0,1]$ (a union bound over the $N$ rounds), the second inequality uses $|\sin\vartheta|\le\vartheta$, and the last inequality substitutes $\vartheta=\pi/(2N)$ and $N\ge\pi^2 J/(2\eta)$.

\emph{Output guarantee.} If all $J$ trials survive, the algorithm outputs the point $\widehat a$ that maximizes $c^\top a$ among $a_{\mathrm{safe}}$ and all accepted candidates. Since $a_{\mathrm{safe}}\in\cS$ and Phase~2 never accepts an unsafe candidate, every returned point lies in $\cS$. Conditioned on reaching trial $j$, Phase~1 survives and produces a feasible, $\gamma$-optimal point with probability at least $1/2$, and Phase~2 then accepts it with probability $1$ and no explosion risk. Hence
\begin{gather*}
 \Pr\Bigl[\text{survive all trials and } c^\top\widehat a < \max_{a\in\cS}c^\top a-\gamma\Bigr]
 \le 2^{-J}\le\delta/4,\\
 \Pr[\text{explode}]
 \le J\left(\eta'+\frac{\eta}{2J}\right)=\frac{7\eta}{10}\le \frac{3\eta}{4},
\end{gather*}
where the first line multiplies the conditional failure probabilities, each at most $1/2$, over the $J=\lceil\log_2(4/\delta)\rceil$ trials. The second line is a union bound over the $J$ simulations, each exploding with probability at most $\eta'=\eta/(5J)$, and the $J$ verification tests, each exploding with probability at most $\eta/(2J)$. Since $\eta\le 1$, the algorithm returns with probability $\Pr[\text{return}]\ge 1-3\eta/4\ge 1/4$. Dividing the first line by $\Pr[\text{return}]$ gives
\[
 \Pr\Bigl[c^\top\widehat a < \max_{a\in\cS}c^\top a-\gamma\ \Big|\ \text{return}\Bigr]\le\frac{\delta/4}{1/4}=\delta .
\]

\emph{Query count.} Each of the $J$ simulations uses $O(Q^2/\eta')=O(Q^2J/\eta)$ bomb queries, and each verification test uses $N=O(J/\eta)$ bomb queries, with $2$ oracle calls per controlled bomb query. The total is $O((Q^2+1)J^2/\eta)=\widetilde O(M_{\mathrm{qco}}(d,R/r,\gamma,\delta)^2/\eta)$, where $\widetilde O$ absorbs the factor $J^2=O(\log^2(4/\delta))$.
\end{proof}

\subsection{BCO-QLinTS algorithm}

Because a single Gaussian reward perturbation is optimistic at $a^\star$ only with constant probability $\pi_{\mathrm{opt}}$, stage $s$ of BCO-QLinTS samples $K_s$ independent perturbed reward vectors $\widetilde\theta_s^{(1)},\dots,\widetilde\theta_s^{(K_s)}$, optimizes each over $\cS$ via \Cref{lem:bomb-safe-opt} to obtain safe candidates $a_s^{(1)},\dots,a_s^{(K_s)}\in\cS$, and selects the pair maximizing $\langle\widetilde\theta_s^{(k)},a_s^{(k)}\rangle$ (the winning index $k^\star$ need not itself be an optimistic perturbation at $a^\star$).

Fix $c_{\rm len}>0$ and assume $T\ge 2$, so that the accuracy $\gamma_s=T^{-2}$ used below is less than one (when $T=1$, playing $a_{\mathrm{safe}}$ once incurs zero explosion risk and regret at most $1$). The reward is learned as in QRS-COLTS with $m=0$ and $\lambda=1$: the design matrix is $V_s=I+\sum_{j<s}n_j^2a_ja_j^\top$, the confidence radius is $\beta_s=1+\sqrt{s-1}$, and stage $s$ estimates $\ip{\theta^\star}{a_s}$ to accuracy $1/n_s$ with failure probability $\nu_s$, using $q_s=L_sn_s$ queries, where
\begin{align*}
 \nu_s = \frac{\delta}{\pi^2s^2}, \qquad 
 L_s = \bigl\lceil\max\{1,C_{\rm qmc}\}\log(2/\nu_s)\bigr\rceil,
\end{align*}
and $C_{\rm qmc}>0$ is the universal constant in \Cref{lem:qmc_scalar}. The algorithm uses the stage bound~\eqref{eq:qrs-stage-budget} of QRS-COLTS,
\begin{equation}\label{eq:bco-stage-count}
 S_{\max}=\min\!\left\{T,\;1+\left\lceil\frac{d\log(1+T^2/d)}{\log(1+c_{\rm len}^2)}\right\rceil\right\} .
\end{equation}
At most $S_{\max}$ stages begin on any sample path. Every stage that begins uses at least one query, so at most $T$ stages begin. Moreover, the completed stages satisfy $\sum_sn_s\le\sum_sq_s\le T$, so \Cref{lem:stage_count} (with $\lambda=1$ and $d_{\cA}\le d$) allows at most $\lceil d\log(1+T^2/d)/\log(1+c_{\rm len}^2)\rceil$ of them, and only the last stage can end without completing.

\begin{algorithm}[t]
\caption{BCO-QLinTS with resampled reward directions}\label{alg:bco-qlints}
\begin{algorithmic}[1]
\Require Query budget $T$, failure probability $\delta$, explosion budget $\varepsilon$, known safe point $a_{\rm safe}$, and radii $r,R$.
\State Initialize $V_1=I$ and $s=1$. Use $S_{\max}$ from~\eqref{eq:bco-stage-count}. Terminate immediately upon any bomb explosion or query-budget exhaustion, including inside any subroutine.
\While{queries remain}
\State Set $\widehat\theta_s=V_s^{-1}\sum_{j<s}n_j^2a_j\widehat r_j$, $\eta_s=\varepsilon/S_{\max}$, $\delta'_s=\delta/(2S_{\max})$, $\gamma_s=T^{-2}$, and $K_s$ as in~\eqref{eq:Ks_def_main}.
\State Draw $K_s$ independent standard normal vectors $\zeta_s^{(k)}\in\mathbb R^d$ and form $\widetilde\theta_s^{(k)}=\widehat\theta_s+\beta_s V_s^{-1/2}\zeta_s^{(k)}$.
\State For each $k=1,\dots,K_s$, invoke \Cref{lem:bomb-safe-opt} with objective $\widetilde\theta_s^{(k)}$, accuracy $\gamma_s$, failure probability $\delta'_s/K_s$, and explosion budget $\eta_s/K_s$ to obtain $a_s^{(k)}\in\cS$.
\State Choose $k^\star\in\argmax_{k\in[K_s]}\ip{\widetilde\theta_s^{(k)}}{a_s^{(k)}}$ (breaking ties by smallest index), and set $a_s=a_s^{(k^\star)}$, $\widetilde\theta_s=\widetilde\theta_s^{(k^\star)}$, and $w_s=\|a_s\|_{V_s^{-1}}$.
\State If $w_s=0$, spend all remaining queries at $a_s$ and stop. Otherwise set $n_s\gets\lceil c_{\rm len}/w_s\rceil$ and $q_s\gets L_sn_s$, and estimate the mean reward at $a_s$ to accuracy $1/n_s$ using $q_s$ queries (as specified below).
\State If all $q_s$ reward queries complete before horizon $T$, set $V_{s+1}=V_s+n_s^2a_sa_s^\top$, record $\widehat r_s$, and increment $s\gets s+1$.
\EndWhile
\end{algorithmic}
\end{algorithm}

\paragraph{Stage reward estimator.}
Because $a_s\in\cS$ is a single classical safe action, querying \Cref{def:safe-action-bomb-oracle} at $\ket{a_s}$ always yields flag outcome $0$ and implements a scalar reward qsample oracle for $R(a_s,\omega)$. Line~7 runs scalar QME (\Cref{lem:qmc_scalar}) with accuracy $1/n_s$ and failure probability $\nu_s$, which uses at most $C_{\rm qmc}n_s\log(2/\nu_s)\le q_s$ queries, and spends any remaining calls at $a_s$. Since $\nu_s=\delta_s/6$ for the schedule $\delta_s$ of~\eqref{eq:delta_s_def_main}, the reward estimator has the two properties of \Cref{lem:stage_estimator} with $m=0$ and $\delta/6$ in place of $\delta$. A reward phase truncated by the horizon is discarded without updating $V_s$.

\subsection{Regret analysis}

\begin{proposition}[Reward confidence and optimism under resampling]\label{prop:bco-resampled-optimism}
Assume $\|\theta^\star\|_2\le 1$ and $\|a\|_2\le 1$ for all $a\in\cA$. Let $N_{\mathrm{Gauss}}$ be as in~\eqref{eq:NGauss_def_main}, so that $\sum_{s\le S_{\max}}K_s\le N_{\mathrm{Gauss}}$, and in this appendix let $B:=\sqrt d+\sqrt{2\log(6N_{\mathrm{Gauss}}/\delta)}$, which is~\eqref{eq:B_def_main} with $\delta/3$ in place of $\delta$ under the logarithm. With probability at least $1-\delta/2$, every perturbed reward vector sampled during the run satisfies
\begin{align*}
 \|\widetilde\theta_s^{(k)}-\theta^\star\|_{V_s}\le (B+1)\beta_s, \qquad 
 \|\widetilde\theta_s^{(k)}\|_2\le 1+(B+1)\beta_s,
\end{align*}
and at every reached stage $s$ in which all $K_s$ calls in Line~5 return $\gamma_s$-optimal actions in $\cS$, the selected pair $(\widetilde\theta_s,a_s)$ satisfies $\ip{\widetilde\theta_s}{a_s}\ge\mu^\star-\gamma_s$.
\end{proposition}

\begin{proof}
At most $S_{\max}$ stages begin, so the run draws at most $\sum_{s\le S_{\max}}K_s\le N_{\mathrm{Gauss}}$ Gaussian vectors. Consider three events:
\begin{enumerate}[label=(\roman*)]
\item $\|\widehat\theta_s-\theta^\star\|_{V_s}\le\beta_s$ at every stage, which is the reward part of~\eqref{eq:con_norm}; by \Cref{lem:Con_explicit} with $m=0$ and $\delta/6$ in place of $\delta$ (the stage estimates fail with probabilities $\nu_s=\delta_s/6$), it fails with probability at most $\delta/6$;
\item $\|\zeta_s^{(k)}\|_2\le B$ for every draw; by \Cref{lem:gauss_norm} with $N=N_{\mathrm{Gauss}}$ and $\delta'=\delta/3$, it fails with probability at most $\delta/6$;
\item at every stage $s$ at which the bound of~(i) holds, some draw $k_0$ satisfies $\ip{\widetilde\theta_s^{(k_0)}}{a^\star}\ge\mu^\star$; by \Cref{lem:local_opt,lem:global_opt}, whose proofs use only the reward coordinate when $m=0$, it fails with probability at most $\delta/6$.
\end{enumerate}
Their intersection has probability at least $1-\delta/2$. On it,
\begin{align*}
 \|\widetilde\theta_s^{(k)}-\theta^\star\|_{V_s}
 &\le\|\widehat\theta_s-\theta^\star\|_{V_s}+\beta_s\|\zeta_s^{(k)}\|_2\le(B+1)\beta_s,\\
 \|\widetilde\theta_s^{(k)}\|_2
 &\le\|\theta^\star\|_2+\|\widetilde\theta_s^{(k)}-\theta^\star\|_{V_s}\le 1+(B+1)\beta_s,
\end{align*}
where the first line uses $\widetilde\theta_s^{(k)}=\widehat\theta_s+\beta_sV_s^{-1/2}\zeta_s^{(k)}$ with (i) and (ii), and the second uses $V_s\succeq I$. If every call in Line~5 returns a $\gamma_s$-optimal point in $\cS$, then
\[
 \ip{\widetilde\theta_s}{a_s}
 \ge \ip{\widetilde\theta_s^{(k_0)}}{a_s^{(k_0)}}
 \ge \ip{\widetilde\theta_s^{(k_0)}}{a^\star}-\gamma_s
 \ge \mu^\star-\gamma_s,
\]
where the first inequality is the selection rule in Line~6, the second uses $\gamma_s$-optimality of $a_s^{(k_0)}$ and $a^\star\in\cS$, and the third is (iii).
\end{proof}

\begin{lemma}[Stage gap bound]\label{lem:bco-stage-gap}
On the event of \Cref{prop:bco-resampled-optimism}, if every optimizer call in Line~5 of stage $s$ returns a $\gamma_s$-optimal action in $\cS$, then $\Delta_+(a_s)\le (B+1)\beta_s w_s+\gamma_s$.
\end{lemma}

\begin{proof}
We upper-bound the unclipped gap $\mu^\star-\ip{\theta^\star}{a_s}$ at the selected pair $(\widetilde\theta_s,a_s)$:
\begin{align*}
 \mu^\star-\ip{\theta^\star}{a_s}
 &\le \gamma_s+\ip{\widetilde\theta_s-\theta^\star}{a_s}\\
 &\le \gamma_s+\|\widetilde\theta_s-\theta^\star\|_{V_s}\|a_s\|_{V_s^{-1}}
 \le \gamma_s+(B+1)\beta_s w_s,
\end{align*}
where the first inequality rearranges $\ip{\widetilde\theta_s}{a_s}\ge\mu^\star-\gamma_s$ from \Cref{prop:bco-resampled-optimism}, the second inequality applies Cauchy--Schwarz in the $V_s$ and $V_s^{-1}$ norms, and the third inequality substitutes $\|\widetilde\theta_s-\theta^\star\|_{V_s}\le(B+1)\beta_s$ and $w_s=\|a_s\|_{V_s^{-1}}$. Because the right-hand side is nonnegative, the same upper bound holds for $\Delta_+(a_s)=(\mu^\star-\ip{\theta^\star}{a_s})_+$.
\end{proof}

\begin{theorem}[End-to-end bomb-certified safe optimization bound]\label{thm:bco-qlints}
Suppose \Cref{ass:safe-body} holds, $\|a\|_2\le 1$ on $\cA$, and $\|\theta^\star\|_2\le 1$. Fix $\varepsilon\in(0,1]$, $\delta\in(0,1)$, and $c_{\rm len}>0$, and run \Cref{alg:bco-qlints} with $S_{\max}$ as in~\eqref{eq:bco-stage-count}. Then the cumulative explosion probability is at most $\varepsilon$, and with probability at least $1-\delta$, the query-weighted regret up to termination satisfies
\begin{align*}
 R_{\mathrm w}(T)&\le(c_{\rm len}+1)(B+1)\sum_{s=1}^{S_{\max}}L_s\beta_s +T^{-1}\\
 &\quad+\sum_{s=1}^{S_{\max}}K_s\,B_{\mathrm{opt}}(d,R/r,\gamma_s,\delta'_s/K_s,\eta_s/K_s)\\
 &\le\widetilde O\!\left(d^2\log^{3/2}(2T/\delta)\right)
 +\widetilde O\!\left(\frac{d^4\log^2(2T)}{\varepsilon}\operatorname{polylog}(2+R/r,2/\delta)\right),
\end{align*}
where $B$ is the value defined in \Cref{prop:bco-resampled-optimism} and $B_{\mathrm{opt}}$ is evaluated at the worst-case objective norm bound $1+(B+1)\beta_{S_{\max}}$ from that proposition. In particular, the constraints enter the bound only through $R/r$.
\end{theorem}

\begin{proof}
We bound the explosion and optimizer failure probabilities, decompose the query-weighted regret across optimization and reward-estimation queries, and evaluate the resulting stage sums.

\emph{Explosion and optimizer failure probabilities.} Condition on the history before a reached optimizer call $(s,k)$ in Line~5. By \Cref{lem:bomb-safe-opt}, the call explodes with probability at most $\eta_s/K_s$, and it returns without exploding but with objective gap greater than $\gamma_s$ (a ``bad return'') with probability at most $\delta'_s/K_s$. Truncating an in-progress call when the budget $T$ runs out can only prevent later explosions or bad returns. Every returned candidate $a_s$ lies in $\cS$, so the reward-estimation queries in Line~7 cannot explode. Summing over all $s\le S_{\max}$ and $k\in[K_s]$ gives
\begin{align*}
 \Pr[\text{an explosion}] &\le \sum_{s=1}^{S_{\max}}K_s(\eta_s/K_s)=\varepsilon,\\
 \Pr[\text{any bad optimizer return}] &\le \sum_{s=1}^{S_{\max}} K_s(\delta'_s/K_s)=\delta/2,
\end{align*}
where we substituted $\eta_s=\varepsilon/S_{\max}$ and $\delta'_s=\delta/(2S_{\max})$.

\emph{Query-weighted regret decomposition.} Let $\mathcal E$ be the intersection of the event of no bad optimizer returns with the event of \Cref{prop:bco-resampled-optimism}. By a union bound, $\Pr[\mathcal E]\ge 1-\delta/2-\delta/2=1-\delta$; note that $\mathcal E$ includes sample paths that terminate early because of an explosion or the horizon. Rewards lie in $[0,1]$, so $\Delta_+(a)\le 1$ for all $a\in\cS$, and every query $\pi_t$ has loss
\[
 L(\pi_t)=\int_{\cS}\Delta_+(a)\,\pi_t(da)+\pi_t(\cA\setminus\cS)\le 1 .
\]
Hence each bomb-optimization query in Line~5, including a terminating one, contributes at most $1$ to $R_{\mathrm w}(T)$. On $\mathcal E$, each call $(s,k)$ executes at most $B_{\mathrm{opt}}(d,R/r,\gamma_s,\delta'_s/K_s,\eta_s/K_s)$ queries.

If stage $s$ reaches the reward-estimation phase in Line~7, then on $\mathcal E$ all $K_s$ optimizer calls of the stage returned $\gamma_s$-optimal points. Let $\ell_s$ be the number of reward queries executed in stage $s$ at the safe point $a_s\in\cS$. The action marginal of each is the point mass at $a_s$, so each has loss $\Delta_+(a_s)$. When $w_s>0$,
\begin{align*}
 \ell_s\Delta_+(a_s)
 \le\ell_s\bigl((B+1)\beta_sw_s+\gamma_s\bigr)
 \le(c_{\rm len}+1)(B+1)L_s\beta_s+\ell_s\gamma_s ,
\end{align*}
where the first inequality is \Cref{lem:bco-stage-gap}, and the second uses $\ell_sw_s\le q_sw_s=L_sn_sw_s\le L_s(c_{\rm len}+w_s)\le L_s(c_{\rm len}+1)$, which follows from $n_s\le c_{\rm len}/w_s+1$ and $w_s\le\|a_s\|_2\le 1$. The same bound holds when $w_s=0$, where \Cref{lem:bco-stage-gap} gives $\Delta_+(a_s)\le\gamma_s$. Summing over the stages and using $\sum_s\ell_s\gamma_s\le T\cdot T^{-2}=T^{-1}$ gives the first regret inequality.

\emph{Asymptotic evaluation of the stage sums.} Abbreviate $S=S_{\max}$ and $K=K_S$. From~\eqref{eq:bco-stage-count} and~\eqref{eq:Ks_def_main}, $S=O(d\log(2T))$, $K=O(\log(2S/\delta))$, and $N_{\mathrm{Gauss}}=O(S\log(2S/\delta))$ by~\eqref{eq:NGauss_def_main}. Since $L_s\le L_S=O(\log(2S/\delta))$ and $\sum_{s=1}^S\beta_s\le S+\tfrac23S^{3/2}$, the reward-learning term satisfies
\[
 (B+1)\sum_{s=1}^SL_s\beta_s
 \le O\!\left((1+\sqrt d+\sqrt{\log(2S/\delta)})\,L_S\,S^{3/2}\right)
 \le\widetilde O\!\left(d^2\log^{3/2}(2T/\delta)\right),
\]
suppressing logarithmic factors in $2d$, $2/\delta$ and $\log(2T)$.

For the bomb-optimization sum, note that $\|a_{\rm safe}\|_2\le 1$ and $\|a\|_2\le 1$ for all $a\in\cS$, so $\cS\subseteq\mathbb B(a_{\rm safe},\min\{R,2\})$, and we may replace $R$ by $\min\{R,2\}$ in the outer radius of \Cref{lem:qco}. On $\mathcal E$, \Cref{prop:bco-resampled-optimism} gives $\min\{R,2\}\,\|\widetilde\theta_s^{(k)}\|_2\le 2[1+(B+1)\beta_S]$. Hence, at accuracy $\gamma_s=T^{-2}$, \Cref{lem:qco} uses $M_{\mathrm{qco}}=\widetilde O(d\operatorname{polylog}(2+R/r,2/\delta))$ coherent membership queries, suppressing logarithms in $d$ and $T$. With $H=\operatorname{polylog}(2+R/r,2/\delta)$, the bomb-query complexities sum to
\begin{align*}
 \sum_{s=1}^S K_s B_{\rm opt}(d,R/r,\gamma_s,\delta'_s/K_s,\eta_s/K_s)
 &\le \widetilde O\!\left(d^2H\sum_{s=1}^S K_s^2/\eta_s\right)\\
 &\le \widetilde O\!\left(d^2H K^2 S^2/\varepsilon\right)
 \le \widetilde O\!\left(d^4H\log^2(2T)/\varepsilon\right),
\end{align*}
where the first inequality substitutes \Cref{lem:bomb-safe-opt} at per-call explosion budget $\eta_s/K_s$ (one factor of $K_s$ counts the calls and another comes from $(\eta_s/K_s)^{-1}$), the second substitutes $\eta_s=\varepsilon/S$ and $K_s\le K$, and the last substitutes $S=O(d\log(2T))$ and absorbs $K^2=\operatorname{polylog}(d,\log(2T),1/\delta)$ into $\widetilde O(\cdot)$ and $H$. Intersecting $\mathcal E$ with the non-explosion event gives the full-horizon bound, without termination, with probability at least $1-\delta-\varepsilon$.
\end{proof}

\end{document}